\documentclass{article}
\usepackage[utf8]{inputenc}
\usepackage[english]{babel}
\usepackage{graphicx}
\usepackage{amsthm}
\usepackage{thmtools}
\usepackage{caption}
\usepackage{subcaption}
\usepackage{color}

\usepackage{algorithm}
\usepackage{algpseudocode}

\usepackage{bbm}
\usepackage{csquotes}

\usepackage[a4paper, total={6in, 9in}]{geometry}

\declaretheorem[name = Theorem]{Theorem}
\declaretheorem[name = Lemma]{Lemma}
\declaretheorem[name= Assumption]{Assumption}
\declaretheoremstyle[headfont=\normalfont]{normalhead}

\title{Ordering for Non-Replacement SGD}
\author{Yuetong Xu and Baharan Mirzasoleiman}

\usepackage[
backend=biber,
style=authoryear-icomp,
]{biblatex}
\begin{document}
\maketitle

\begin{abstract}
    One approach for reducing run time and improving efficiency of machine learning is to reduce the convergence rate of the optimization algorithm used. Shuffling is an algorithm technique that is widely used in machine learning, but it only started to gain attention theoretically in recent years. With different convergence rates developed for random shuffling and incremental gradient descent, we seek to find an ordering that can improve the convergence rates for the non-replacement form of the algorithm. Based on existing bounds of the distance between the optimal and current iterate, we derive an upper bound that is dependent on the gradients at the beginning of the epoch. Through analysis of the bound, we are able to develop optimal orderings for constant and decreasing step sizes for strongly convex and convex functions. We further test and verify our results through experiments on synthesis and real data sets. In addition, we are able to combine the ordering with mini-batch and further apply it to more complex neural networks, which show promising results.
\end{abstract}

\section{Introduction}\label{sec:intro}

Throughout this paper we consider the minimization of the finite sum additive cost function with the update function of the form:\\
\begin{equation}
F(x) = \frac{1}{n}\cdot\sum_{i = 1}^n f_i(x) \quad , \quad x_{i} = x_{i-1} - \alpha \nabla f_{i}(x_{i-1}).
\end{equation}
\\
Where the component functions $f_i$ are all convex if not strongly convex. This model is standard for learning from $n$ data examples, where $f_i$ represent the loss with respect to data sample $i$. The data examples can be ordered by random or by specific permutation. In standard SGD and Random Shuffling (SGD without resampling), the algorithms randomize the order the data examples are processed and hence how the iterate is updated. $\alpha$ represent the learning rate or step size. In practice, algorithms apply decreasing learning rate for better convergence.\\
\\
Only in recent years Shuffling, algorithms without replacement, started to attract attention of scholars in the field though in practice such algorithms are largely applied instead of SGD for optimization problems. It is already shown that Random Shuffling converges faster than SGD after finite epochs for strongly convex cases (\cite{haochen2019random}). \cite{nedic2001incremental} presented an upper bound for the distance between algorithm iterate and the optimal for incremental gradient descent (gradient descent with specific ordering) which establishes a basis for our analysis. The distance between the iterate and optimal is the error of the algorithm which we try to minimize. Hence by improving this distance, we can further optimize the algorithm.\\
\\
For large scale machine learning, the size of the data can result in long run time and convergence time. One way to improve the efficiency of learning is to improve the convergence rate of the algorithms used. Stochastic gradient descent and Random Shuffling are one of the widely applied algorithms for large scale machine learning. In this paper, we seek to find an optimal ordering that can improve the convergence of the non-replacement algorithm.\\
\\
In our analysis, we utilize the inherent properties of convex and strongly convex functions to develop a tighter upper bound whose value is dependent on the gradient values. From the bound we reach the following conclusions:
For strongly convex function $f_i$s, in which the functions are
and Lipschitz continuous $f_i$s with Lipschitz constant $L$ and bounded gradients with $M$, if our initial $x_0$ is sufficiently close to the optimal and with initial step size $\alpha < \frac{1}{LGM}$, then 
\begin{itemize}
    \item \textbf{for decreasing step size, ordering by decreasing initial gradient would result in the fastest convergence.}
    \item for constant step size, random ordering, ordering by decreasing gradient norm, ordering by increasing gradient norm all result in the same convergence rate.
\end{itemize}
From convex functions, with the same conditions, the results of regarding the ordering is the same.\\
\\
We confirm our theoretical results through experiments on synthetic and real data sets including Iris data set and Boston Housing data set. By comparing the loss values for random ordering, decreasing gradient ordering, and increasing gradient ordering, we observed that the relationship stated above is indeed valid. We test the influence of the initial value of $x$ on the convergence in which if $x_0$ is close enough to the optimal $x^*$, decreasing ordering is best. We further expand our results to more complex neural networks and mini-batch algorithms through experiments on MNIST data set, Fashion MNIST data set, and CIFAR data sets.\\

\section{Related Works}
Gradient Descent methods have been long studied and applied in research. Various studies have attempted to improve the convergence of SGD upon stochastic gradient descent by ordering based on average gradient error (\cite{lu2022sgd}), ordering based on gradient bounds (\cite{Sebastian2017}), ordering based on gradient error (\cite{Mohtashami2022CharacterizingF}), and weighting samples by gradients (\cite{Alain2015}). Others have approached the issue by decreasing batch size and prioritizing important data, e.g. by training on top $q$ samples from the batch with greatest loss (\cite{kawaguchi2020ordered}), resampling a smaller batch for training based on pre-sampled data distribution (\cite{katharopoulos2018not}), and training on data with smallest loss in the set (\cite{Shah2020sample}).\\\\
In contrast to the existing volume of studies on SGD, Shuffling, the non-replacement version of SGD only started to gain attention recently. Large portion of results mostly focus on comparing Shuffling methods with SGD. \cite{shamir2016} provided theoretical support that in worst case scenario, shuffling will not be significantly worse than SGD. \cite{safran2020good} proved the optimization error decays with $O(\frac{1}{k})$ for SGD with replacement, while it is $O(\frac{1}{k^2})$ for SGD without replacement for sufficiently large k. \cite{haochen2019random} proved random shuffling could achieve better worst case convergence rates for all scenarios. In addition, many papers focus purely on theoretical guarantees on convergence rates of various shuffling methods. \cite{nedic2001incremental} proved the convergence rates of the Incremental Subgradient Methods which contain an upper bound for the distance between algorithm iterate and the optimal. \cite{gurbuzbalaban2015random} proposed a de-biased shuffling method which takes on averages of iterates with convergence rate of $O( \frac{1}{k^2} )$ . \cite{gurbuzbalaban2019convergence} proved shuffling has  convergence rate of $O(\frac{1}{k^2})$. \cite{Nguyen2021unified} proposed a unified convergence analysis of various shuffling methods in which randomized reshuffling is $O(\frac{1}{nT^2})$ with general case being $O(\frac{1}{T^2})$. \cite{Rajput2020close} provided a convergence rate of $O (\frac{1}{T^2} + \frac{n^2}{T^3})$ for quadratic functions and a lower bound of $\omega (\frac{n}{T^2})$ for strongly convex functions composed of summation of smooth functions. \cite{MishchenkoBR20} further investigated on convergence behavior of random shuffling, shuffle once, and incremental gradient descent. \cite{needell2014stochastic} focused on methods with reweighting the sampling distribution.\\\\
Our paper is unique from existing work as it seeks to find a \textit{deterministic ordering} of examples at every epoch that can improve the convergence of non-replacement SGD through providing theoretical and empirical evidence. Our permutation utilizes the gradient properties of the training data. It differs from related work in SGD not only from the non-replacement setting, but also that we derive a concrete permutation based on gradient value. It distinguishes from \cite{kawaguchi2020ordered} in that our permutation results in a ordering of that can applied to full data while their result only samples the top $q$ data points of each batch with no specified order.

\section{Optimal Ordering}

The loss function we minimize is $ F(x) = \frac{1}{n}\cdot\sum_{i = 1}^n f_i(x)$, where each $f_i$ is smooth and convex if not strongly convex. For this paper we will focus on the non-replacement version of SGD in which we would update each step as $
x_{i,k} = x_{i-1, k} - \alpha_{i,k} \nabla f_{\sigma_k(i)}(x_{i-1, k})$, where $x_{i,k}$ denote the iterated value at $i^{th}$ iteration of $k^{th}$ epoch, $\alpha_{i,k}$ denotes the step size at the corresponding iterate,$\nabla f$ denotes the sub-gradient at the corresponding iterate, and $\sigma_k(i)$ denote the permutation of the data values at epoch k. For simplicity, in the following content we would not explicitly write out $\sigma_k(i)$ in our equation and $x_k$ will denote the first point at epoch $k$ i.e. $x_{0,k}$.\\
\begin{Assumption}
    In this paper, we further assume there is a bound on the norm of the gradient of $f_i$. (Denote as $C_i$). 
    \begin{equation}
    ||\nabla f_i(x)|| \leq C_i \quad \forall x
    \end{equation}
\end{Assumption}
\noindent
Next, we provide near-optimal orderings for minimizing strongly convex and convex functions.\\
\subsection{Strongly Convex Functions}

\begin{Lemma}
For different $f_i$s which are strongly convex or convex at the same iterate x, the larger $||\nabla f_i(x)^T||$, the larger the upper bound for $f_i(x) - f_i(x^*)$, where $x^*$ denote the optimal point we seek to obtain. (The proof can be found in Appendix \ref{pf:lemma_1}.)
\end{Lemma}
\noindent
Furthermore, by property of strongly convex function, for $\forall x$
\begin{equation} \label{strong_convex_bound}
    f(x - A) \geq f(x)+\nabla f(x)^T(- A)+\mu^2||x - A||^2
    = f(x)-\nabla f(x)^T(A)+\mu^2||x - A||^2
\end{equation}
where $\mu >0$ is the strongly convex constant.\\
\\
For strongly convex functions, we obtain the permutation by observing the bound on distance between initial point of the epoch and the optimal solution, i.e. $||x_k - x^*||^2$. By using existing bound from \cite{nedic2001incremental} and the existence of the lower bound of the strongly convex functions (Eq\ref{strong_convex_bound}), we are able to further tighten the upper bound on the distance by ordering the data points in a specific order. For strongly convex case, we would consider 2 scenarios: decreasing step size per iteration and constant step size.\\\\
Intuitively, for decreasing step size, if we wish to converge faster, we would need to have the algorithm approximate Gradient Descent as close as possible, which in turn suggests having larger gradients in the beginning of the epoch to pair with the relative larger values of the step sizes. On the other hand, if we assume the sub-gradient we obtain in the algorithm is very close to the actual value in gradient descent (since we are sufficiently close to the optimal), then no matter how we order the data points, with the same step size we would update the iterate with same amount after an epoch.\\
\begin{Theorem}[Strongly convex with decreasing step size per iteration]
For strongly convex functions $f_i$ that are differentiable, with bounded gradients, and with strongly convex constant $m_i$, ordering by \textbf{decreasing value of norm of gradient $M_i$ at the beginning of every epoch} provides a tighter upper bound for $||x_{k+1} - x^*||$, which is
    \begin{equation}
    \begin{array}{l@{}l}
    ||x_{k+1}-x^*||^2\leq ||x_{k} - x^*||^2 - 2||x_k-x^*||\sum_{i = 1}^n\alpha_{i,k}||M_i|| + \alpha_{k}^2\left(2\sum_{i = 1}^{n-1}[(n-i)C_i^2] + \sum_{i = 1}^n C_i^2 \right) \\
    \qquad\qquad\qquad\qquad\qquad\qquad-\alpha_{k}^2\alpha_{k+1}\sum_{i = 1}^{n-1}(n-i)m_i\cdot max{(||M_i||,||M_i^{'}||)}^2 + n\alpha_k\cdot\epsilon_k
    \end{array}
    \end{equation}
\end{Theorem}
In the equation, n denotes the total number of data samples, $\alpha$ representing the learning rate, and $\epsilon$ being a small constant. Since $\alpha$ is small, to minimize the bound on the difference with the optimal, we have to maximize the term of with $\alpha$ in first power, i.e. $\sum_{i = 1}^n\alpha_{i,k}||M_i||$. With decreasing step size, this suggest ordering by decreasing value of norm of gradients.\\

\begin{Theorem}[Strongly convex with constant step size]
For strongly convex functions $f_i$ that are differentiable, with bounded gradients, and with strongly convex constant $m_i$, \textbf{random ordering, ordering by decreasing gradient norm, ordering by increasing gradient norm generate the same upper bound} for $||x_{k+1} - x^*||$ which is
    \begin{equation}
    \begin{array}{l@{}l}
    ||x_{k+1}-x^*||^2\leq ||x_{k} - x^*||^2 - 2\alpha ||x_k-x^*|| \sum_{i = 1}^n ||M_i|| + \alpha^2\left(2\sum_{i = 1}^{n-1}[(n-i)C_i^2] + \sum_{i = 1}^n C_i^2 \right) \\ \hspace{4.5cm}
    -\alpha^3\sum_{i = 1}^{n-1}(n-i)m_i\cdot max{(||M_i||,||M_i^{'}||)}^2 + n\alpha\cdot\epsilon_k
    \end{array}
    \end{equation}
\end{Theorem}

Similar as previous theorem, n denotes total samples, $\alpha$ denotes learning rate, and $\epsilon$ is a small constant. Since $\alpha$ is small and constant, the only value that can influence the upper bound is $\sum_{i = 1}^n ||M_i||$. However, the sum of initial gradients would be the same if we start training from the same state, and hence the three orderings results the same upper bound.\\
\\
The proof of the theorems can be found in the Appendix \ref{pf:Theorem_1_2}.
\subsection{Convex Functions}
Following the result of Lemma 1, for convex function, for $\forall x$
\begin{equation}
f(x - A) \geq f(x) + \nabla f(x)^T(-A)
= f(x) - \nabla f(x)^T(A)
\end{equation}
\noindent
With this property, we can derive the following:
\begin{equation}
\begin{array}{l@{}l}
    f_n(x_{0,k} - \sum_{i = 1}^{n-1} \alpha_{i,k}M_i^{'}) - f_n(x^*) &\geq f_n(x_{0,k}) - \nabla f_n(x_{0,k})^T(\sum_{i = 1}^{n-1} \alpha_{i,k}M_i^{'})- f_n(x^*)\\
    &{}\geq [f_n(x_{0,k})- f_n(x^*)] - \sum_{i = 1}^{n-1}\alpha_{i,k}\max{(||M_i||,||M_i^{'}||)}^2\\
\end{array}
\end{equation}
Following the logic for strongly convex loss function, the result about ordering the examples by the sub-gradient value at the beginning of the epoch also holds for convex loss functions.\\
\begin{Theorem}[Convex with decreasing step size per iteration]
For convex functions $f_i$ that are differentiable with bounded gradients, ordering by \textbf{decreasing value of norm of gradient $M_i$ at beginning of every epoch} provides a tighter upper bound for $||x_{k+1} - x^*||$ which is
    \begin{equation}
    \begin{array}{l@{}l}
    ||x_{k+1}-x^*||^2\leq ||x_{k} - x^*||^2 - 2||x_k-x^*||\sum_{i = 1}^n\alpha_{i,k}||M_i||\\  \hspace{4.5cm} +\alpha_{k}^2\left(2\sum_{i = 1}^{n-1}[(n-i)C_i^2] + \sum_{i = 1}^n C_i^2 \right) + n\alpha_k\cdot\epsilon_k
    \end{array}
    \end{equation}
\end{Theorem}
\noindent
Similar as the strongly convex case, with small learning rate $\alpha$, to minimize the upper bound we need to maximize the term $\sum_{i = 1}^n\alpha_{i,k}||M_i||$. With decreasing step size, this implies ordering by decreasing gradient norm so the larger $||M_i||$ are paired with the larger $\alpha_{i,k}$.

\begin{Theorem}[Convex with constant step size]
For convex functions $f_i$ that are differentiable with bounded gradients, \textbf{random ordering, ordering by decreasing gradient norm, ordering by increasing gradient norm generate the same upper bound} for $||x_{k+1} - x^*||$ which is
    \begin{equation}
    \begin{array}{l@{}l}
    ||x_{k+1}-x^*||^2\leq ||x_{k} - x^*||^2 - 2\alpha ||x_k-x^*|| \sum_{i = 1}^n ||M_i||\\ \hspace{4.5cm}
    + \alpha^2\left(2\sum_{i = 1}^{n-1}[(n-i)C_i^2] + \sum_{i = 1}^n C_i^2 \right) + n\alpha\cdot\epsilon_k
    \end{array}
    \end{equation}
\end{Theorem}
\noindent
With constant step size and small step size $\alpha$, the only term that can significantly change the upper bound is $\sum_{i = 1}^n ||M_i||$. However, starting at the same state, the summation of gradient norms at beginning of the epoch is the same for the three orderings being compared. Therefore, the upper bound would be the same for the three orderings.\\
\\
The proof of the theorems could be found in the Appendix \ref{pf:Theorem_3_4}. The direct result of these theorems is the following algorithm \ref{alg:cap_1}.\\

\begin{algorithm}
\caption{Full Data Ordering}\label{alg:cap_1}
\begin{algorithmic}
\Require step size $\alpha$, number of iterations T, initial weights $w_0$
\State Construct gradient array G of size n
\For{t = 0, 1, ..., T}
\For{i = 0, 1, ..., n}
\State $G[i] = ||\nabla f(w_{T \cdot n +i}; D[i])||$
\Comment{Calculate gradient norm}
\EndFor
\State Sort data points within  sorted based on decreasing gradient norm, denote $\sigma_{T}$. 
\For {i = 0, 1, ..., n} 
\State Update the model parameters: $w_{T \cdot n +i + 1} \gets w_{T \cdot n +i} - \alpha \nabla f(w_{T \cdot n +i}; x(\sigma_{T}(i)) )$
\EndFor
\EndFor
\end{algorithmic}
\end{algorithm}

\subsection{Ordering with Data Selection}
In this algorithm, we split the data points into small batches of size S and iterate over q samples within each batch. Similar to decreasing ordering for full data, we can maximize $\sum_{i = 1}^n\alpha_{i,k}||M_i||$ within each batch. Within batches, we select examples with largest gradients. If we assume $||M_{N\cdot S+{q+1}}|| < ||M_{(N+1)\cdot S+{1}}||$ (N: number of batches ran, S: size of each batch, q: the amount of sample we train on in each batch), then comparing continuing training on $q+1$ sample in $(N+1)$th mini-batch and directing start training on $(N+2)$th mini-batch results $\alpha_{N\cdot S+{q+1},k}||M_{N\cdot S+{q+1}}|| < \alpha_{N\cdot S+{q+1},k}||M_{(N+1)\cdot S+{1}}||$. Similar results follows for all points in the batch after the top q iterates with largest gradient norm.\\
\\
For decreasing step size, with our algorithm, the iterates with large $||M_i||$ in the following batch will have a larger corresponding $\alpha_{i,k}$, maximizing $\sum_{i = 1}^{\frac{n}{S}\cdot q}\alpha_{i,k}||M_i||$. 
For constant step size, since we assume the neglected data have a small gradient, the difference between neglecting and not neglecting the data will be small.\\

\begin{algorithm}
\caption{Ordering With Data Selection}\label{alg:cap_2}
\begin{algorithmic}
\Require step size $\alpha$, number of iterations T, initial weights $w_0$
\State Construct data set into batches (B[i]s) of size S
\State Construct gradient array G of size S
\For{t = 0, 1, ..., T}
\For{i = 0, 1, ..., n}
\For{j = 0, 1, ..., S} 
\Comment{Calculate gradient norm}
\State $G[i][j] = ||\nabla f(w_{T \cdot n \cdot S +i \cdot S}; B[i][j])||$
\Comment{B[i][j] denote the $j^{th}$ point in batch i.}
\EndFor
\State Sort data points within B[i] sorted based on decreasing gradient norm, denote $\sigma_{T,i}$. \\
\For {j = 0, 1, ..., S} 
\Comment{Update model parameters}
\State $w_{T \cdot n \cdot S +i \cdot S + j + 1} \gets w_{T \cdot n \cdot S +i \cdot S + j} - \alpha \nabla f(w_{T \cdot n \cdot S +i \cdot S + j}; x(\sigma_{T,i}(j)) )$
\EndFor
\EndFor
\EndFor
\end{algorithmic}
\end{algorithm}

\subsection{Ordering with Mini-Batch SGD}
In addition to the ordering proposed above, we combine our ordering with mini-batch algorithm, which is a more efficient way of training (\cite{kawaguchi2020ordered}), to apply to large data sets which SGD cannot be utilized. Two algorithms result from this combination: sorting before mini-batch and sorting within mini-batch. \\
\\
For sorting before mini-batch, we directly combine our ordering to the mini-batch. In every epoch, we first order the data based on decreasing gradient ordering. Then we construct a balanced data set from the ordered data, in which each mini-batch contain equal number of data from each class if data is categorical. Following the construction, we run mini-batch algorithm on the resultant data set. We can control the size of data iterated by setting the mini-batch size to q instead of S if needed. Since we have sorted the data, the skipped data points will be the ones with small gradient norm. Hence with decreasing step size, the skipped data will have minimal effect on the model.\\
\begin{algorithm}
\caption{Sorting Before Mini-Batch}\label{alg:cap_3}
\begin{algorithmic}
\Require step size $\alpha$, number of iterations T, initial weights $w_0$
\For{t = 0, 1, ..., T}
\State Construct gradient array G of size nS
\For{k = 0, 1, ..., nS} \Comment{Denote the data set as D, size nS.}
\State $G[k]= ||\nabla f(w_{T*n}; D[k])||$
\Comment{Calculate gradient norm}
\EndFor
\State Sort data points within D based on decreasing gradient norm, denote $\sigma_{T}$.
\State Construct balanced mini-batches from the ordered data.
\For{i = 0, 1, ..., n}
\State Update the model parameters: $w_{T \cdot n + i + 1} \gets w_{T \cdot n +i} - \alpha \frac{1}{S} \sum_{j = 0}^{S} \nabla f(w_{T \cdot n +i}; x(\sigma_{T,i}(j)) )$
\EndFor
\EndFor
\end{algorithmic}
\end{algorithm}
\\
Regarding sorting within mini-batch, instead of evaluating over the entire mini-batch (size $S$), within each mini-batch the top $q$ examples with largest gradients sorted in decreasing order are evaluated. The algorithm will calculate the gradients at beginning of each mini-batch, resulting in a finer gradient estimate for the entire batch.\\
\begin{algorithm}
\caption{Sorting Within Mini-Batch}\label{alg:cap_4}
\begin{algorithmic}
\Require step size $\alpha$, number of iterations T, initial weights $w_0$
\State Construct data set into batches (B[i]s) of size S
\State Construct gradient array G of size S
\Comment{B[i][j] denote the $j^{th}$ point in batch i.}
\For{t = 0, 1, ..., T}
\For{i = 0, 1, ..., n}
\For{j = 0, 1, ..., S}
\State $G[i][j] = ||\nabla f(w_{T \cdot n +i \cdot S}; B[i][j])||$
\Comment{Calculate gradient norm}
\EndFor
\State Sort data points within B[i] sorted based on decreasing gradient norm, denote $\sigma_{T,i}$.
\State Update the model parameters: $w_{T \cdot n + i + 1} \gets w_{T \cdot n +i} - \alpha \frac{1}{S} \sum_{j = 0}^{S} \nabla f(w_{T \cdot n +i}; x(\sigma_{T,i}(j)) )$
\EndFor
\EndFor
\end{algorithmic}
\end{algorithm}
\\
Sorting within mini-batch is a more direct application of our result, but for large data sets, the calculation can be costly if the model needs to update and sort every mini-batch. Thus, for large data sets, sorting before mini-batch is a more practical and efficient way of training.\\

\section{Data from Experiments}
We further test our algorithm on synthetic data sets and real data sets to compare the loss function value for three ordering: random ordering, decreasing gradient norm, increasing gradient. For each ordering norm, we generate the corresponding permutation at the beginning of each epoch and visit the data points based on the ordering.\\

\subsection{Synthetic Data}
For the synthetic data, we randomly generate 32 vectors in $\mathbbm{R}^2$ with value in the range of [-10, 10]. The initial step size used for this case is $10^{-4}$. The loss function is defined as $F = \sum_{i = 1}^m ||x - x^*||^4_4 +||x - x^*||^2_2$. The following graphs plot loss vs iteration for constant step size, decreasing step size per iteration, and decreasing step size per epoch correspondingly.\\

\begin{figure}[htbp]\centering
     \begin{subfigure}[b]{0.30\textwidth}
         \includegraphics[width=\textwidth]{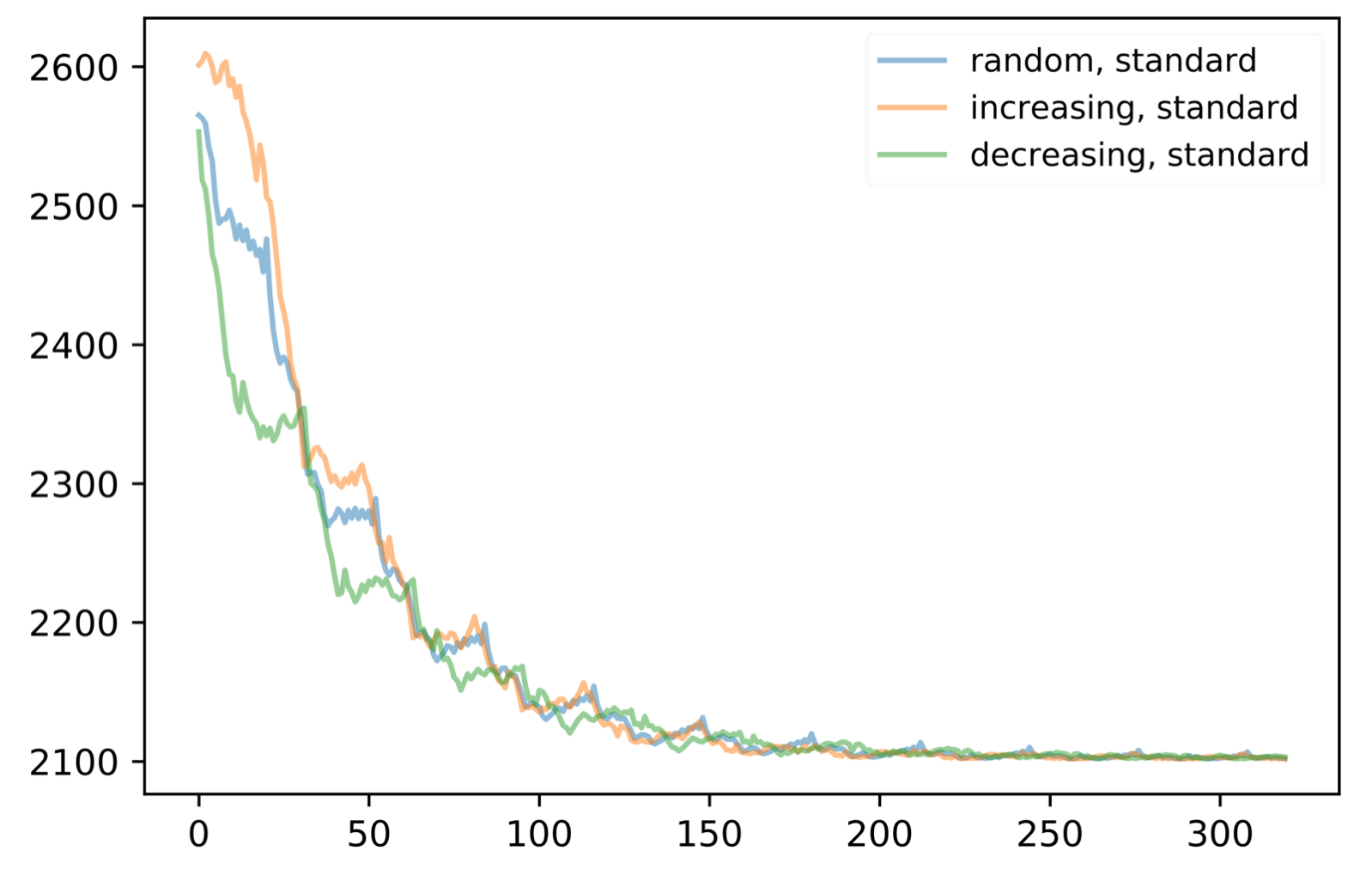}
         \caption{constant step size}
         \label{fig:syn_constant}
     \end{subfigure}
     \begin{subfigure}[b]{0.30\textwidth}
         \includegraphics[width=\textwidth]{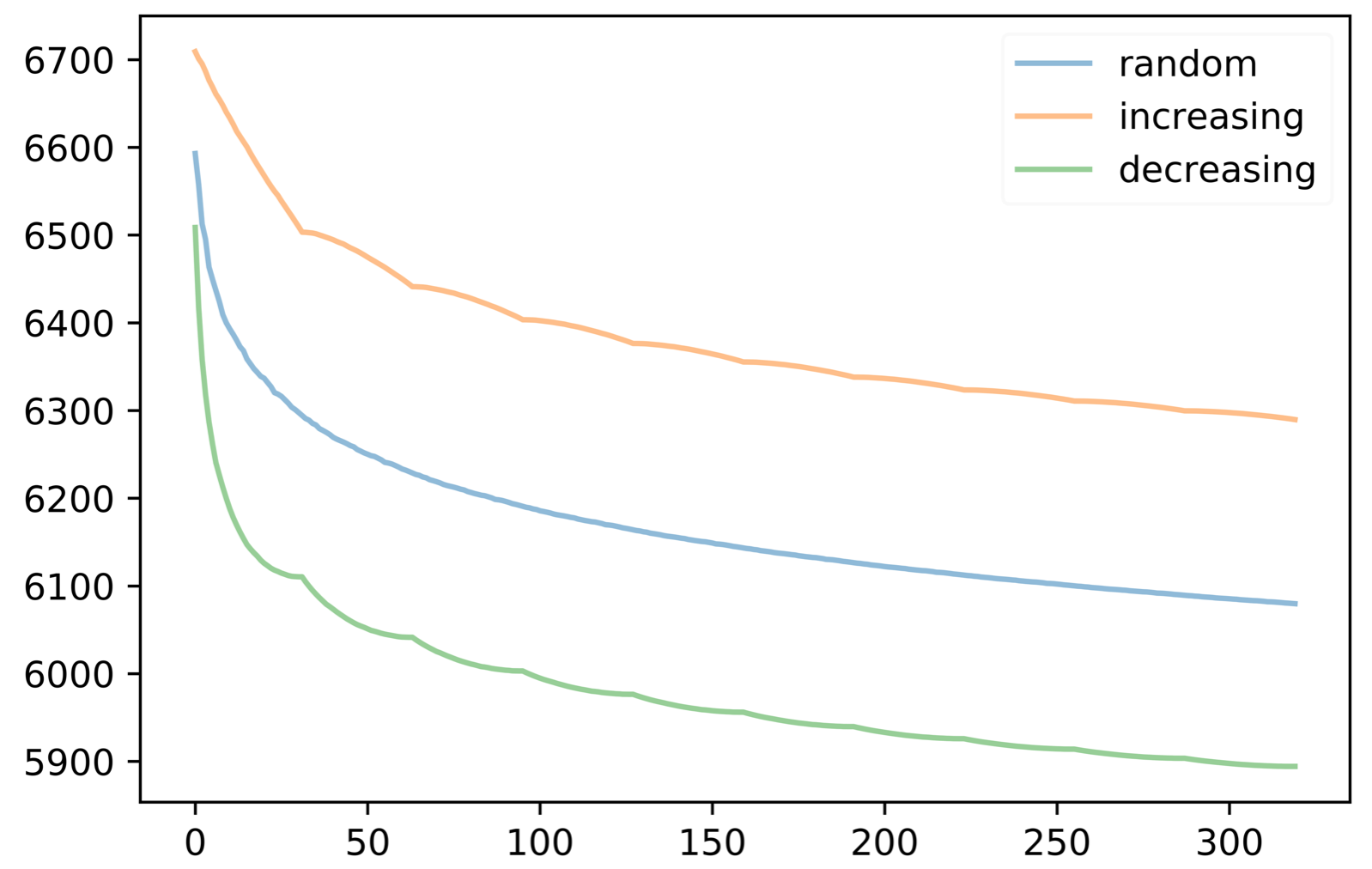}
         \caption{decreasing per iteration}
         \label{fig:syn_decrease_it}
     \end{subfigure}
     \begin{subfigure}[b]{0.30\textwidth}
         \includegraphics[width=\textwidth]{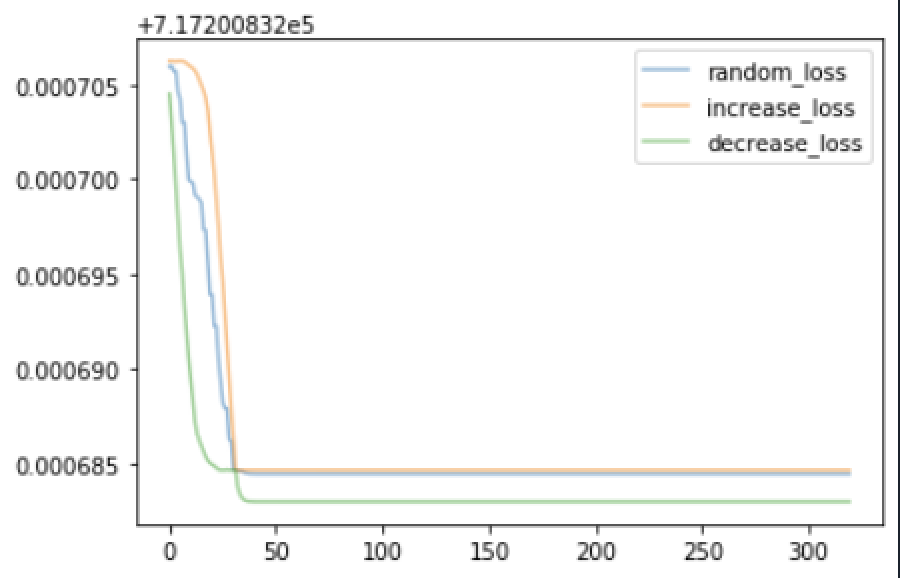}
         \caption{decreasing per epoch}
         \label{fig:syn_decrase_epo}
     \end{subfigure}
        \caption{Synthetic Data}
        \label{fig:synthesis three graphs}
\end{figure}
\hfill \break
From the graph we could observe that for constant step size the loss value at the beginning and end of an epoch is approximately the same for the three orderings, which supports that convergence rate is same for all orderings when step size is constant. From the graph for decreasing step size per epoch, we could clearly observe a difference between the loss of the three orderings, which further validate our theoretical results.\\
\\
To further explore the relationship between distance from the optimal and convergence for the different orderings, we conducted two different experiments.\\ \\
First, we applied a different loss function: $F = \sum_{i = 1}^m \mathbbm{1}_{x<0}||x - x^*||^4_4 + (1-\mathbbm{1}_{x<0})||x - x^*||^2_2$, where $\mathbbm{1}_{x<0}$ is the indicator function.\\
\\
With the case when the initial value is [-9,-9], the result for decreasing step size per iteration is shown below in figure \ref{fig:syn_decrase_distance},
where we can observe random ordering actually achieves a better result than decreasing gradient. This is because the initial value is too far from the optimal. \\

\subsection{Iris Data Set}
\begin{figure}[htbp]\centering
     \begin{subfigure}[b]{0.30\textwidth}
         \includegraphics[width=\textwidth]{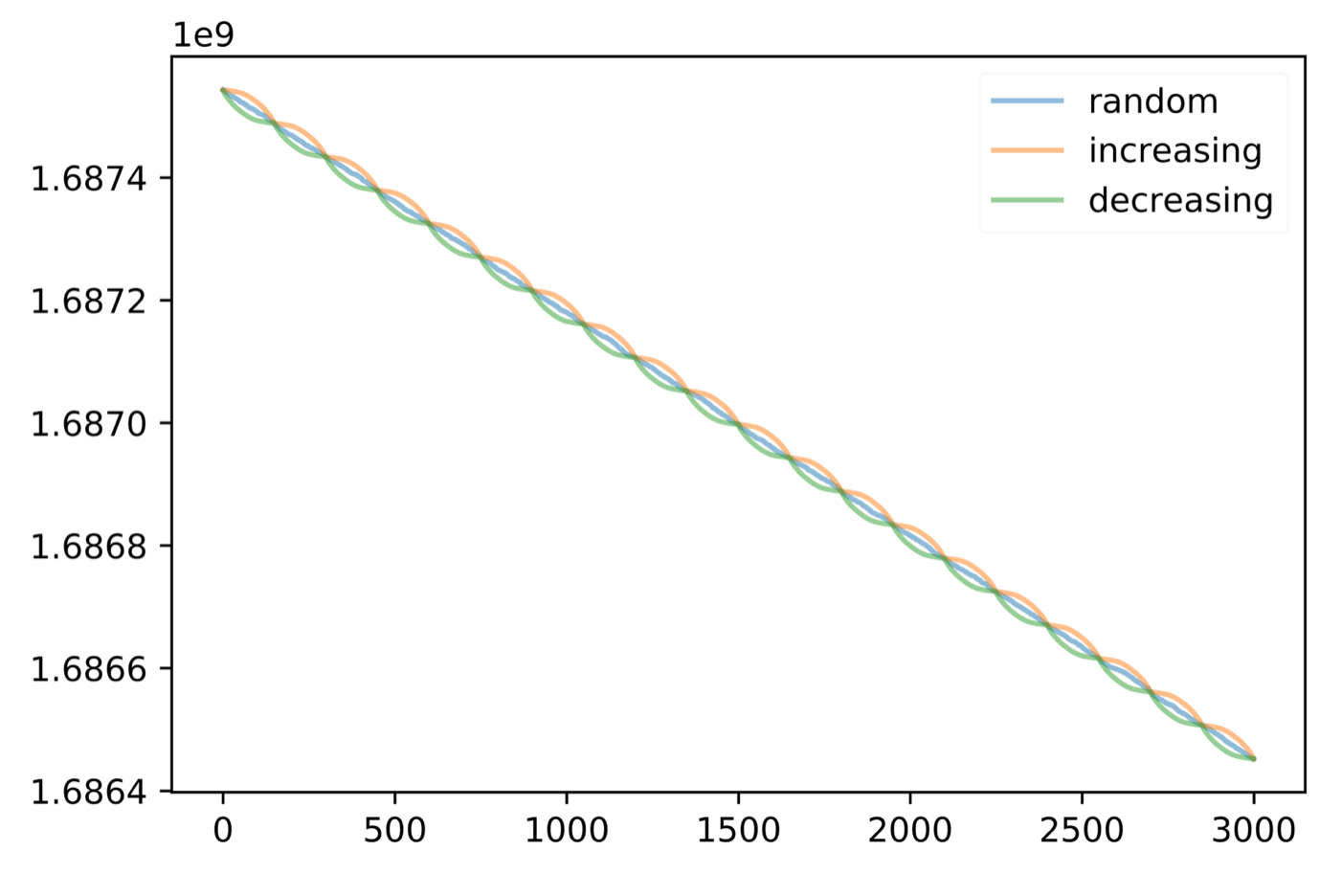}
         \caption{constant step size}
         \label{fig:iris_constant}
     \end{subfigure}
     \begin{subfigure}[b]{0.30\textwidth}
         \includegraphics[width=\textwidth]{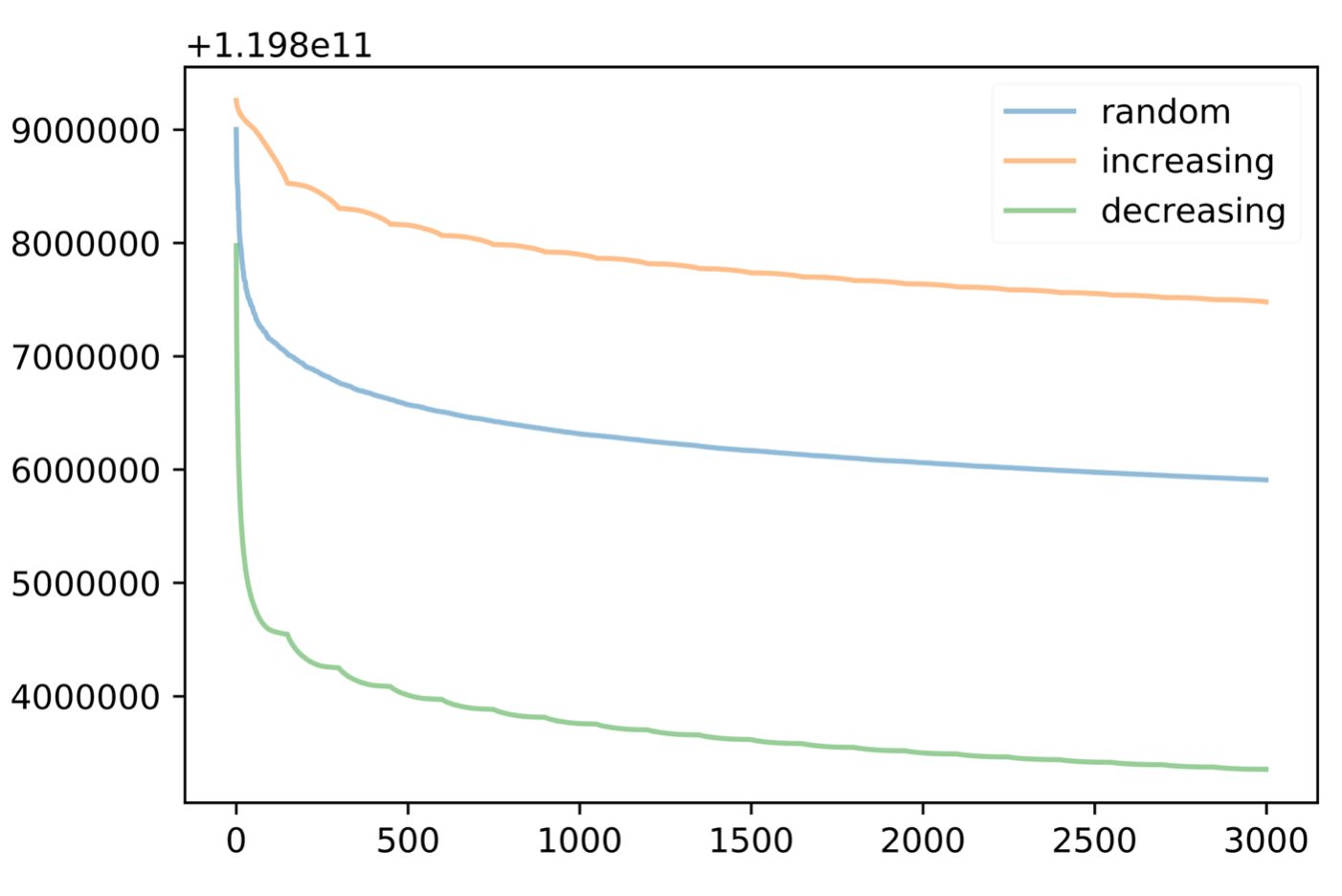}
         \caption{decreasing per iteration}
         \label{fig:iris_decrease_it}
     \end{subfigure}
     \begin{subfigure}[b]{0.30\textwidth}
         \includegraphics[width=\textwidth]{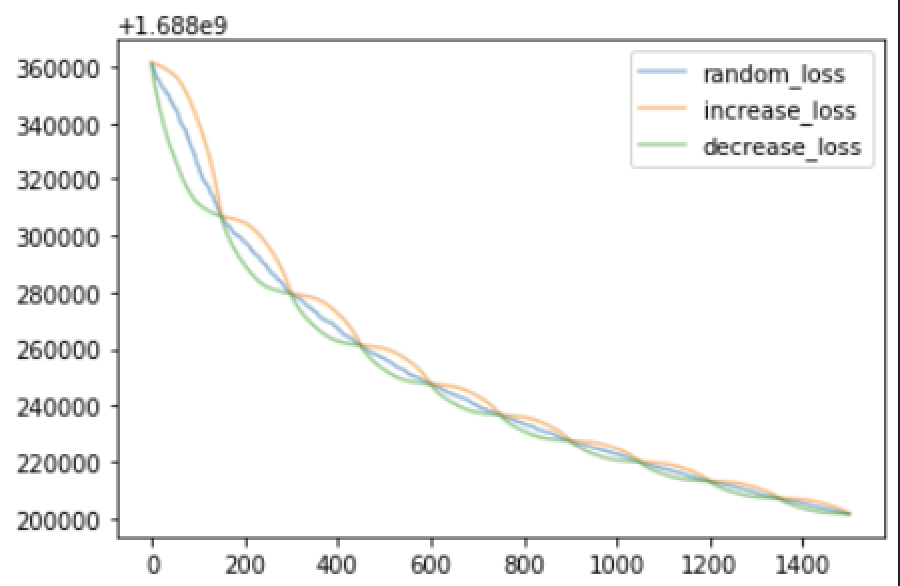}
         \caption{decreasing per epoch}
         \label{fig:iris_decrase_epo}
     \end{subfigure}
        \caption{Iris Data set}
        \label{fig:iris three graphs}
\end{figure}

We also test the algorithm on the Iris data set which contain 150 vectors with 4 features (sepal length, sepal width, petal length, and petal width; denote as $x$) and 1 target (species; denote as y). Here we apply a simple linear regression model. The initial step size for this case is $6\cdot10^{-4}$. (To avoid overflow due to high power, we further multiply $10^{-10}$ in front of the gradients.) The loss function in this case is $F = \sum_{i = 1}^m ||(w^Tx+b) - y||^4_4 +||(w^Tx+b) - y||^2_2$. \\
\\
Figure \ref{fig:iris three graphs} plot loss vs iteration for constant step size, decreasing step size per iteration, and decreasing step size per epoch correspondingly.
Similar to the result from Synthetic data, for constant step size, the loss value is approximately the same for all orderings after an epoch; for decreasing step size, ordering by decreasing initial gradients results in the smallest loss value. These results further enhances our theoretical conclusion.\\
\subsection{Boston Housing Data}
We further test the algorithm on the Boston Housing data set which contain 560 vectors with 13 features (denote as $x$) and 1 target (price; denote as $y$). Here we also apply a simple linear regression model. The initial step size for this case is $6\cdot10^{-4}$. (To avoid overflow due to high power, we further multiply $10^{-10}$ in front of the gradients.) Similarly, $F = \sum_{i = 1}^m ||(w^Tx+b) - y||^4_4 +||(w^Tx+b) - y||^2_2$.\\
\\
The following graph plot loss vs iteration for constant step size, decreasing step size per iteration, and decreasing step size per epoch correspondingly. The same relation persists and can be clearly observed.\\
\begin{figure}[htbp] \centering
     \begin{subfigure}[b]{0.30\textwidth}
         \includegraphics[width=\textwidth]{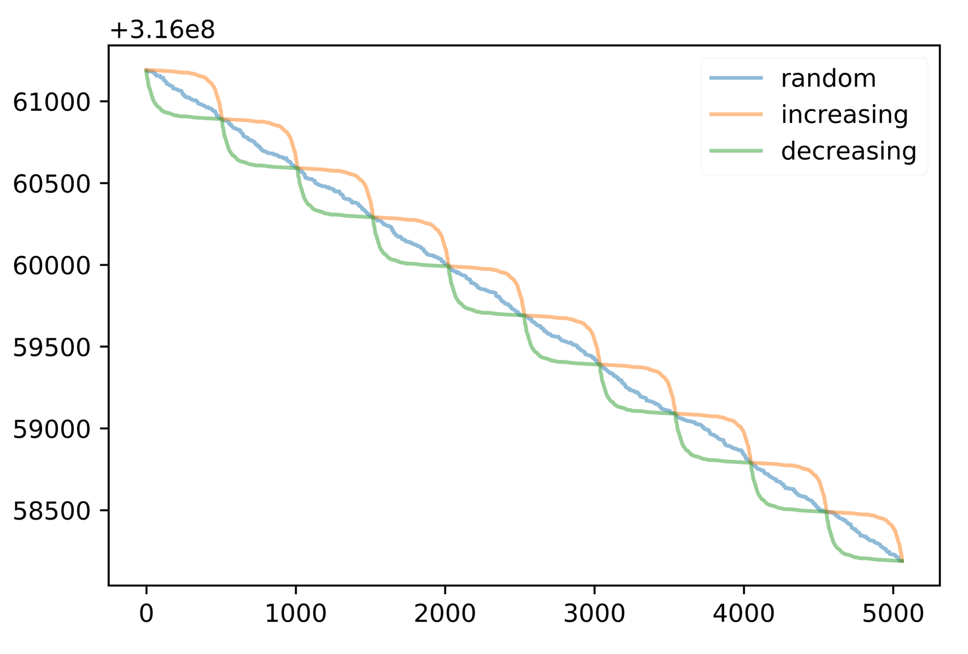}
         \caption{constant step size}
         \label{fig:boston_constant}
     \end{subfigure}
     \begin{subfigure}[b]{0.30\textwidth}
         \includegraphics[width=\textwidth]{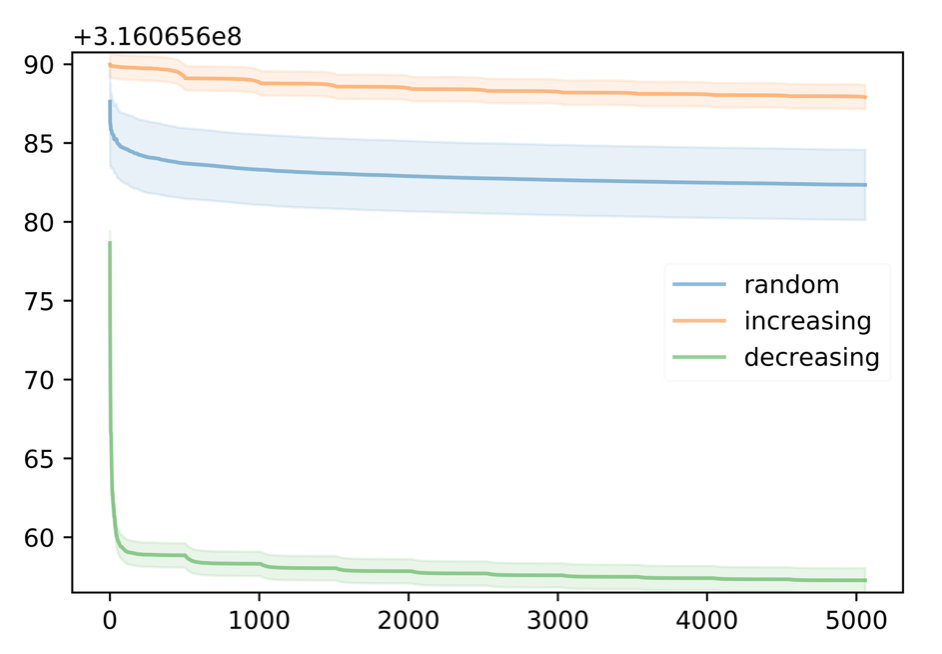}
         \caption{decreasing per iteration}
         \label{fig:boston_decrease_it}
     \end{subfigure}
     \begin{subfigure}[b]{0.30\textwidth}
         \includegraphics[width=\textwidth]{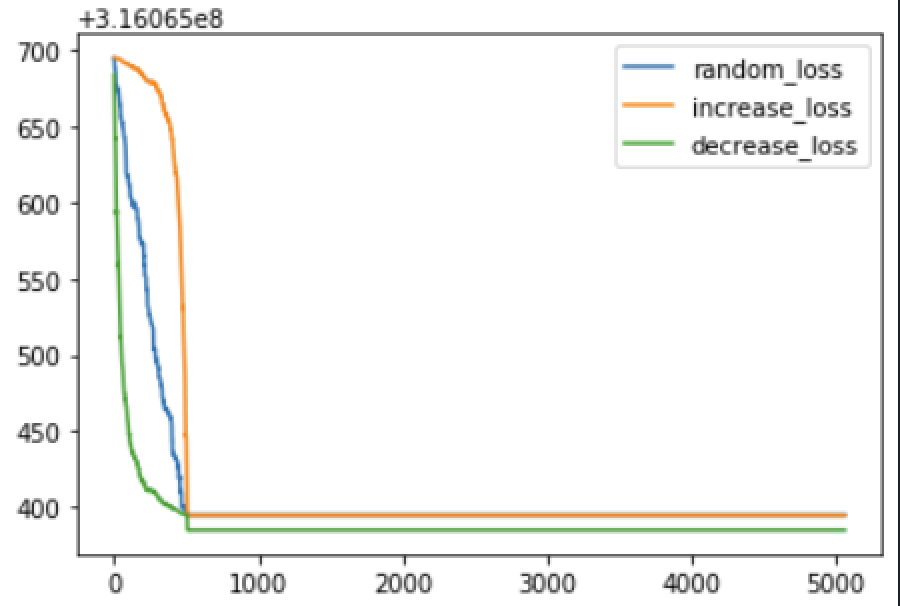}
         \caption{decreasing per epoch}
         \label{fig:boston_decrase_epo}
     \end{subfigure}
        \caption{Boston Data set}
        \label{fig:boston three graphs}
\end{figure}
\subsection{Comparison with Data Selection Algorithm}
In this section, we apply the proposed permutation combined with data selection with batches. For the algorithm, each batch is of size $s$, and within each batch only $q$ data points with largest gradient is iterated. For all the algorithms, we first run random shuffling using the full data (as above) till the iterate is close enough to the optimal and then switch to the permutations being compared.\\
\\
We apply the algorithm to the same three data sets above. Within each data set, we set $S$ as 0.3, 0.6, 0.8 of the original data size and $q$ as 0.18, 0.5, 0.8 of $S$, with full results in Appendix \ref{exp:data_selec}. \\
\begin{figure}[htbp] \centering
     \begin{subfigure}[b]{0.30\textwidth}
         \includegraphics[width=\textwidth]{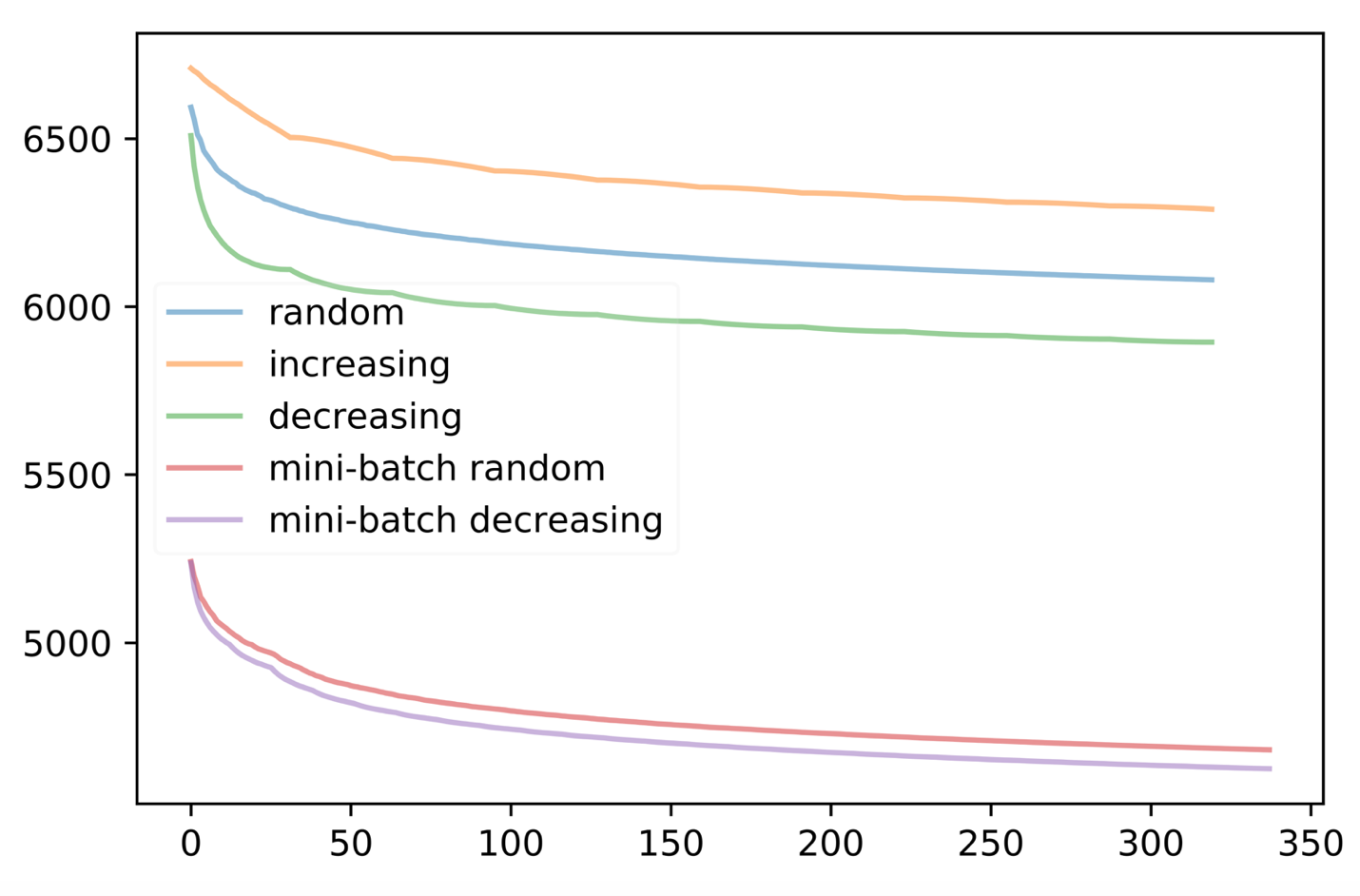}
         \caption{Synthesis}
         \label{fig:syn_mini}
     \end{subfigure}
     \begin{subfigure}[b]{0.30\textwidth}
         \includegraphics[width=\textwidth]{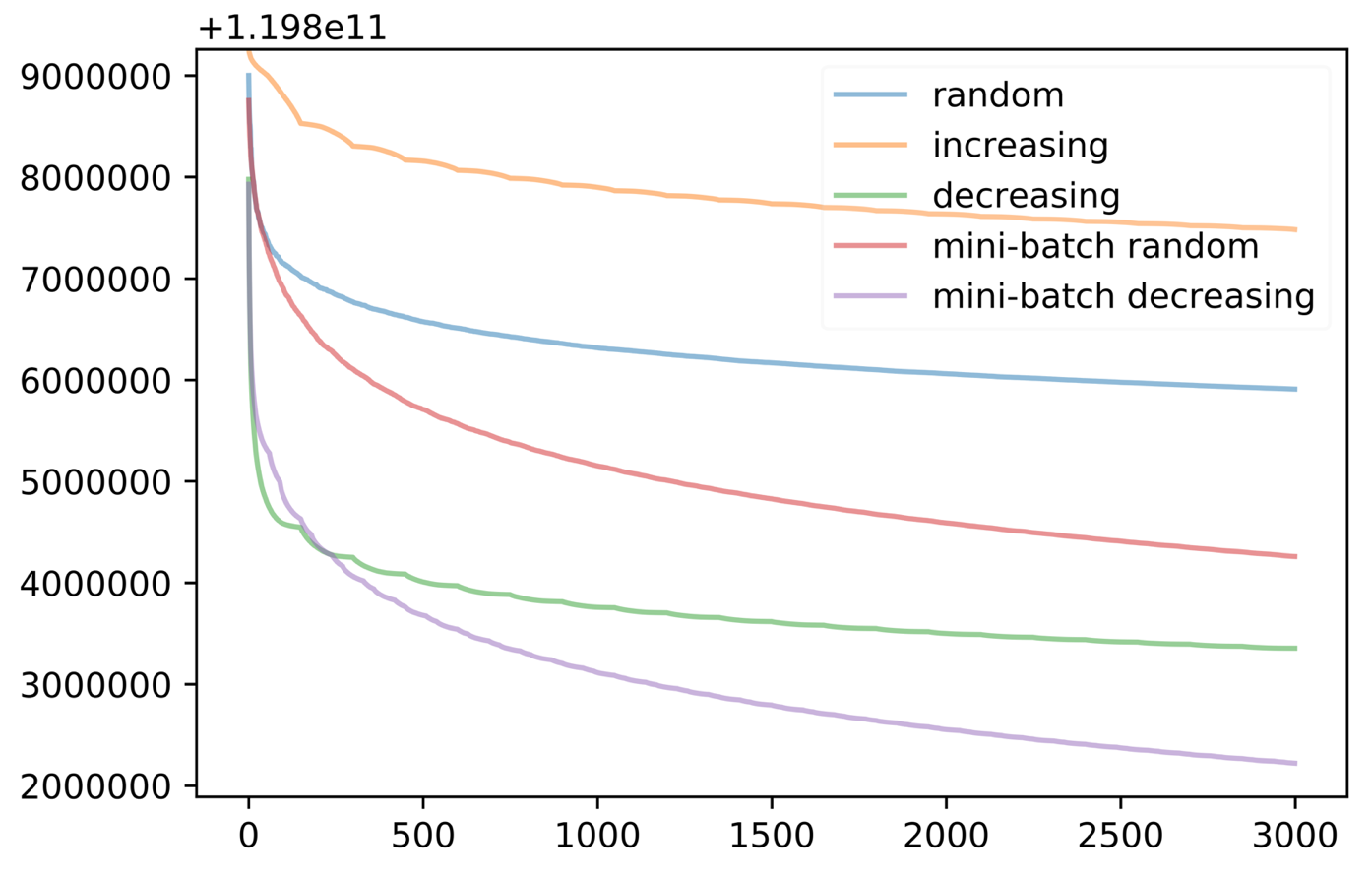}
         \caption{Iris}
         \label{fig:iris_mini}
     \end{subfigure}
     \begin{subfigure}[b]{0.30\textwidth}
         \includegraphics[width=\textwidth]{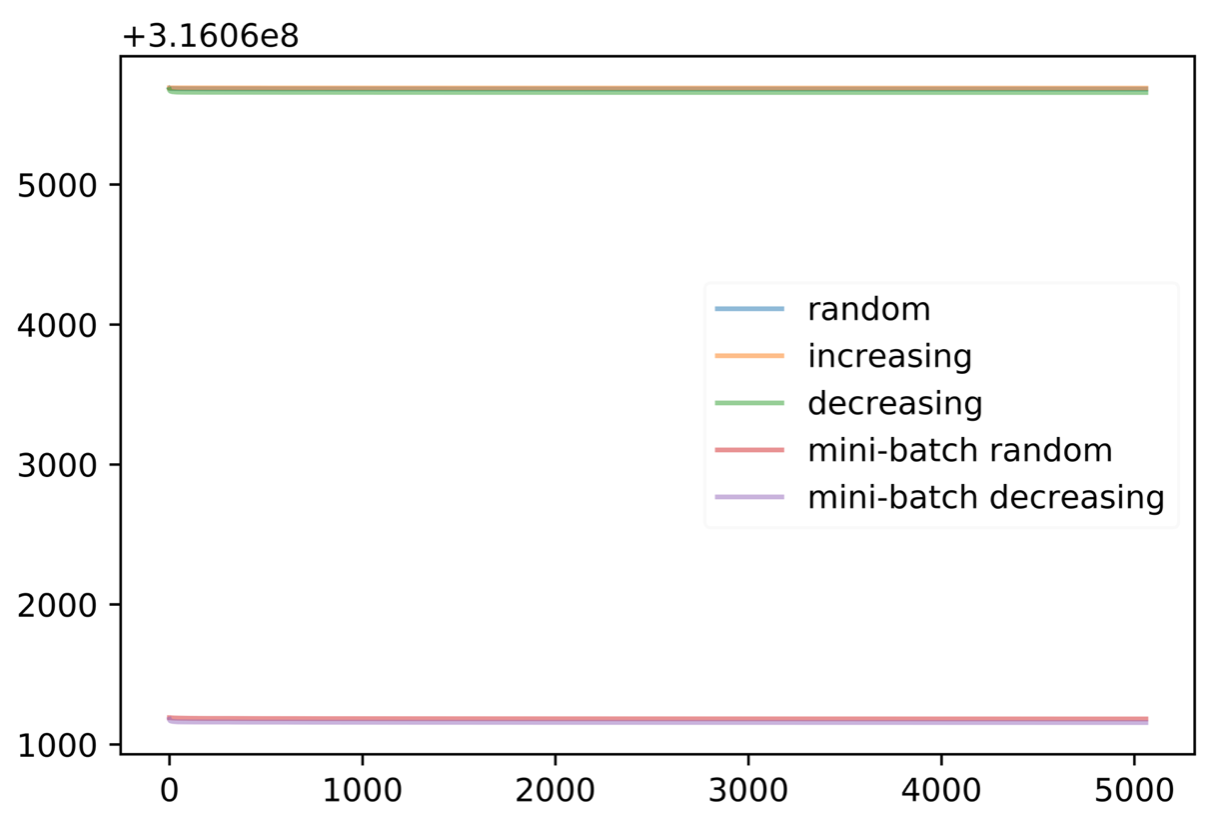}
         \caption{Boston Housing}
         \label{fig:mini_boston}
     \end{subfigure}
        \caption{Data Selection $S$ = 0.8, $q$ = 0.5}
        \label{fig:mini_three_graphs}
\end{figure} \\
\noindent
For these algorithms, we order the $q$ iterates within each batch both randomly and by decreasing gradient norm. From the figures, we could observe that for this algorithm, ordering by decreasing gradient norm for the $q$ iterates results in smaller loss in synthetic and real data sets. Comparing with the none data-selection algorithms, data-selection algorithms with decreasing gradient ordering results in the best result.\\
\subsection{Extending to Mini-Batches on Neural Networks}
Non-convex functions can be split into finite small bounded segments such that each segment is convex. Intuitively, performing our proposed algorithm on these convex segments, we can at least reach a local minimal for the lost functions. Hence in the following experiments, we test our algorithms on larger data sets with both simple and more complex neural networks. (Full results in Appendix \ref{exp:nn_mnist} and \ref{exp:nn_cifar})
\subsubsection{MNIST and Fashion MNIST Data set}
We test our algorithm on the MNIST data set with two different neural network models: 2 layer and 7 layer. Both models are applied to each data set. The MNIST data set consists of images of handwritten digits, with 60,000 training images and 10,000 testing images. For the experiments, we iterate over 50 percent of samples within each mini-batch ($q = 0.5$), with batch sizes of 128 and 256 correspondingly. From the results, we could identify that in both training and testing, the mini-batch algorithm with less data and decreasing gradient ordering results in the highest accuracy. This supports that our gradient ordering can be combined with mini-batch on more complex models like neural networks to further improve convergence and decrease training time. (Full graphs can be found in Appendix \ref{exp:nn_mnist}).\\
\begin{figure}[htbp] \centering
     \begin{subfigure}[b]{0.30\textwidth}
         \includegraphics[width=\textwidth]{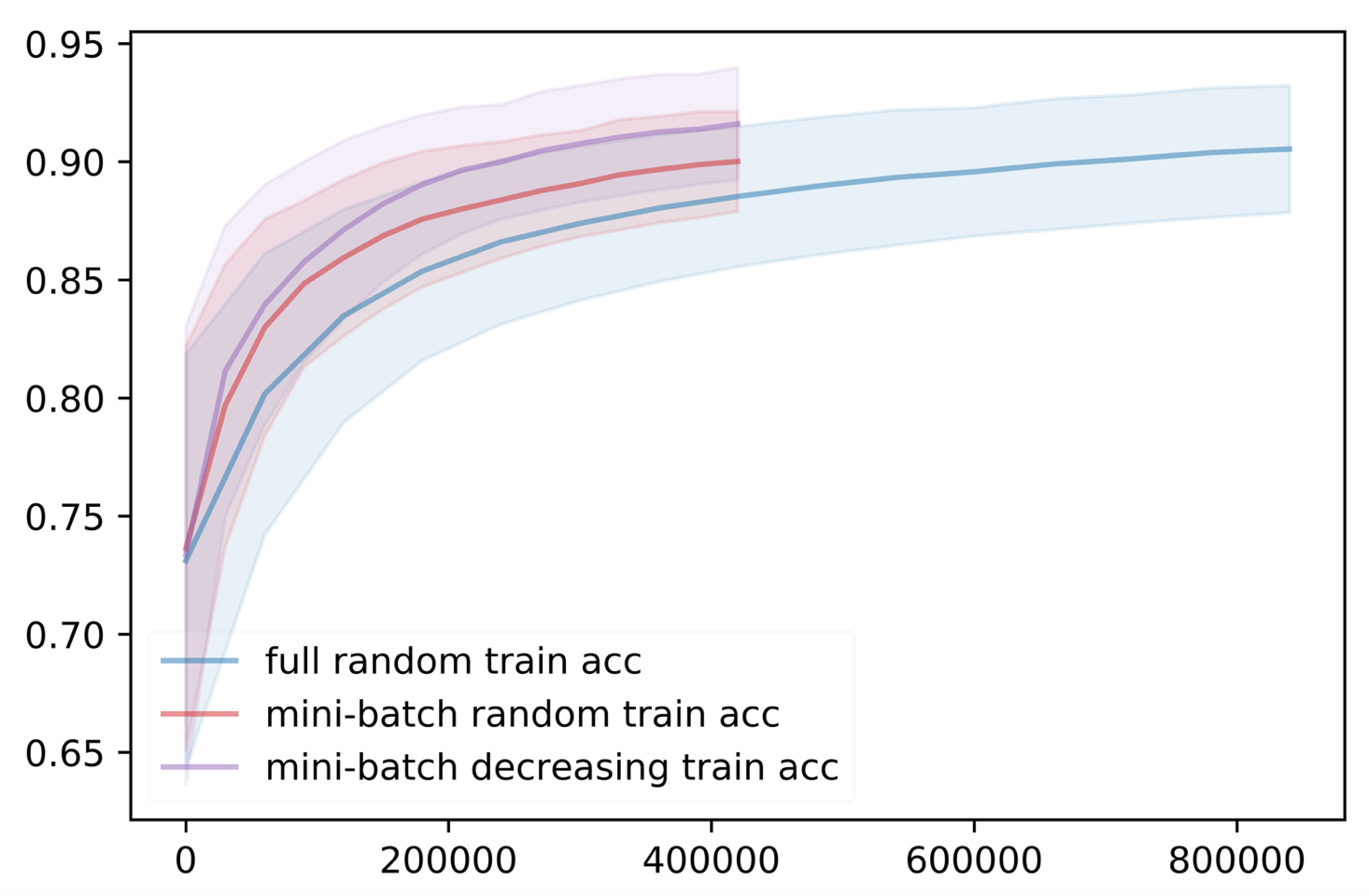}
         \caption{Training Accuracy}
         \label{fig:mini_mnist_2_train}
     \end{subfigure}
      \begin{subfigure}[b]{0.30\textwidth}
         \includegraphics[width=\textwidth]{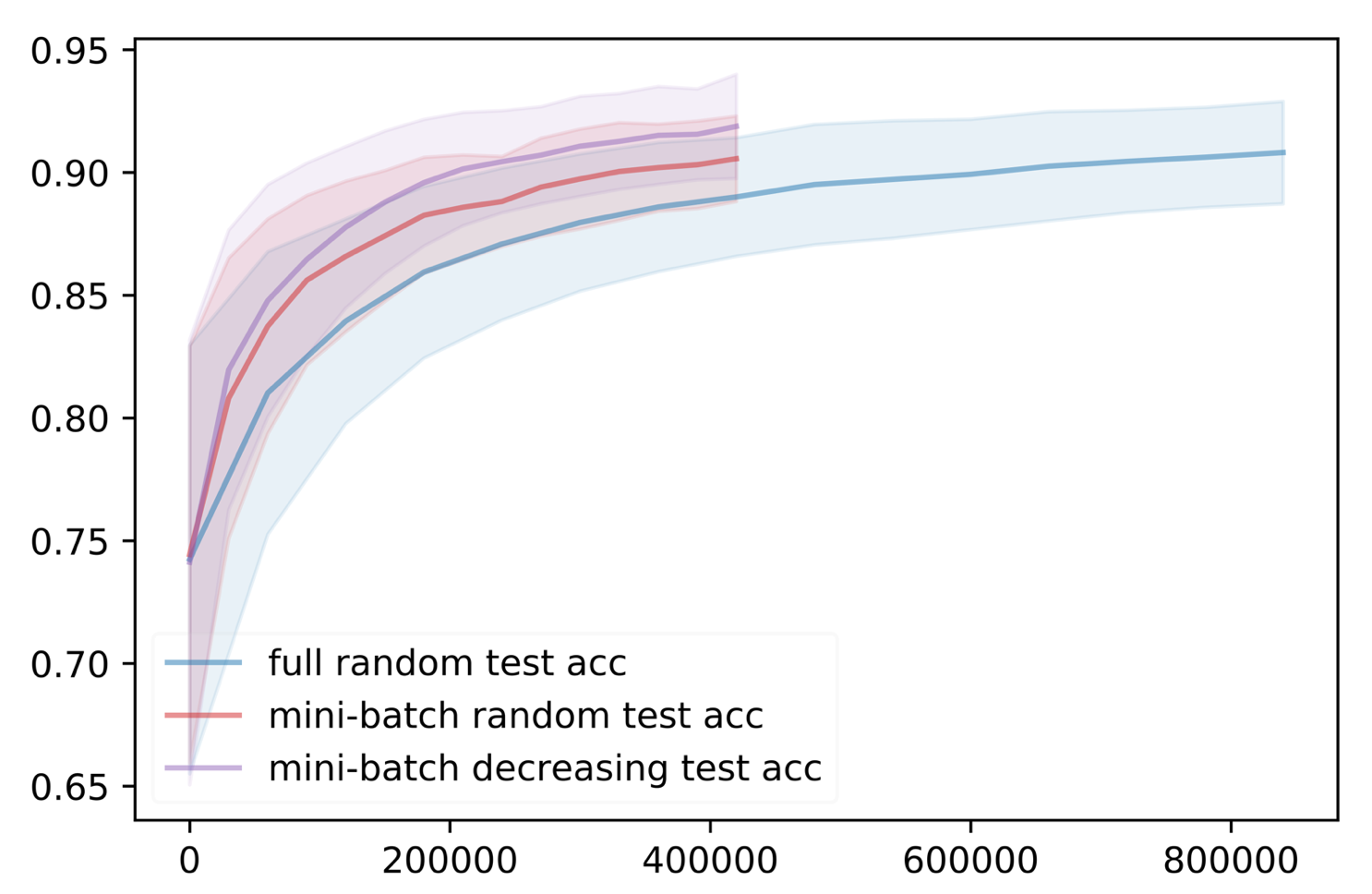}
         \caption{Test Accuracy}
         \label{fig:mini_mnist_2_test}
     \end{subfigure}
        \caption{MNIST 2-layer Batch = 128}
        \label{fig:mini_mnist_128}
\end{figure}
\\
To further investigate the performance of our algorithm on neural networks, we apply it to the Fashion MNIST Dataset, containing fashion images of clothing, with 60,000 training images and 10,000 testing images. Similarly, we perform the training with $q = 0.5$, and utilize the 2-layer and 7-layer networks as above. The mini-batch algorithms still outperform the algorithm training on full data on both models. In particular, the mini-batch algorithm with decreasing sorting results in the best performance. This result justifies that for adequate models, our ordering can be combined with mini-batch to further improve convergence (Full graphs can be found in Appendix \ref{exp:nn_mnist}).\\
\begin{figure}[htbp] \centering
     \begin{subfigure}[b]{0.30\textwidth}
         \includegraphics[width=\textwidth]{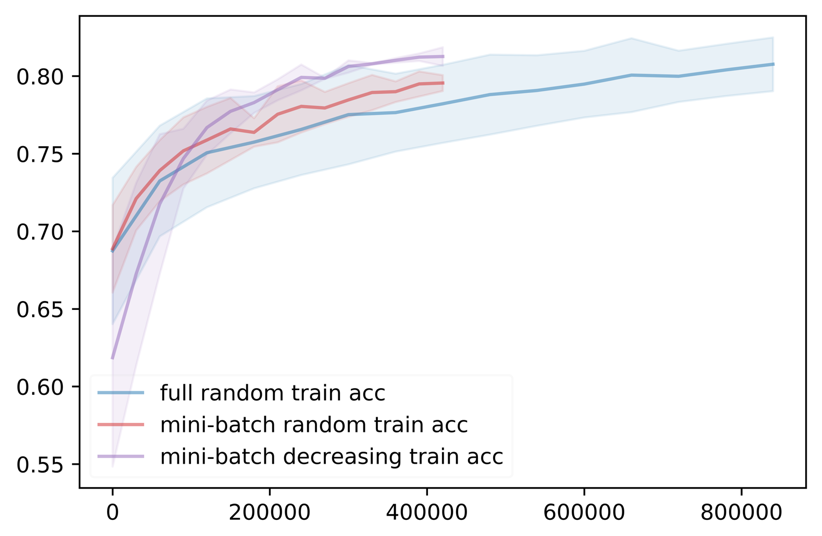}
         \caption{Training Accuracy}
         \label{fig:mini_fmnist_2_train}
     \end{subfigure}
      \begin{subfigure}[b]{0.30\textwidth}
         \includegraphics[width=\textwidth]{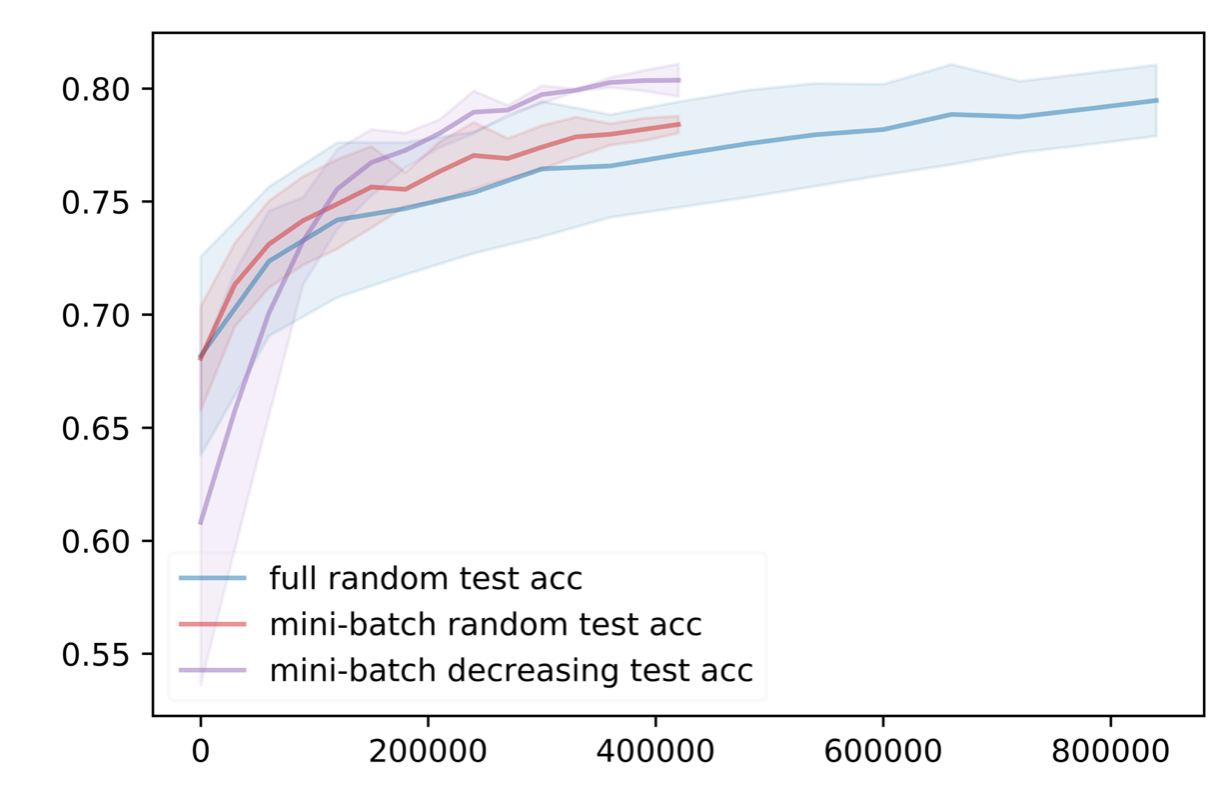}
         \caption{Test Accuracy}
         \label{fig:mini_fmnist_2_test}
     \end{subfigure}
        \caption{Fashion MNIST 2-layer Batch = 128}
        \label{fig:fmini_mnist_128}
\end{figure}

\subsubsection{CIFAR-10 and CIFAR-100 Data set}
Approaching more complex data sets and networks, we validate our ordering and algorithm on CIFAR-10 and CIFAR-100 data sets with ResNet20 and ResNet18, respectively. The baseline result we are comparing with is training the network 200 epochs that achieves test accuracy about 92\% for CIFAR-10 and about 79\% for CIFAR-100. Due the complexity of the model, instead of sorting by decreasing gradient, we respectively sort by norm of logits, the performance of the model on the sample. Also, instead of calculating the permutation within each mini-batch, we calculate the total ordering at the beginning of the epoch to reduce calculation run-time. As a result, samples that are predicted worst (least learned) will be be first iterated in the epoch.\\\\
The highest test accuracy for mini-batch algorithms ($q = 0.5$, using 50\% of data from each mini-batch) is starting the corresponding ordering (random or increasing accuracy) at epoch 140. Comparing with training on full data, the mini-batch algorithm with less data and increasing accuracy ordering reaches the lowest training error with fewer amount of iterations. Though there is a slight difference in test accuracy, the difference is not too significant. (Full graphs can be found in Appendix \ref{exp:nn_cifar}).\\
\begin{figure}[htbp] \centering
     \begin{subfigure}[b]{0.30\textwidth}
         \includegraphics[width=\textwidth]{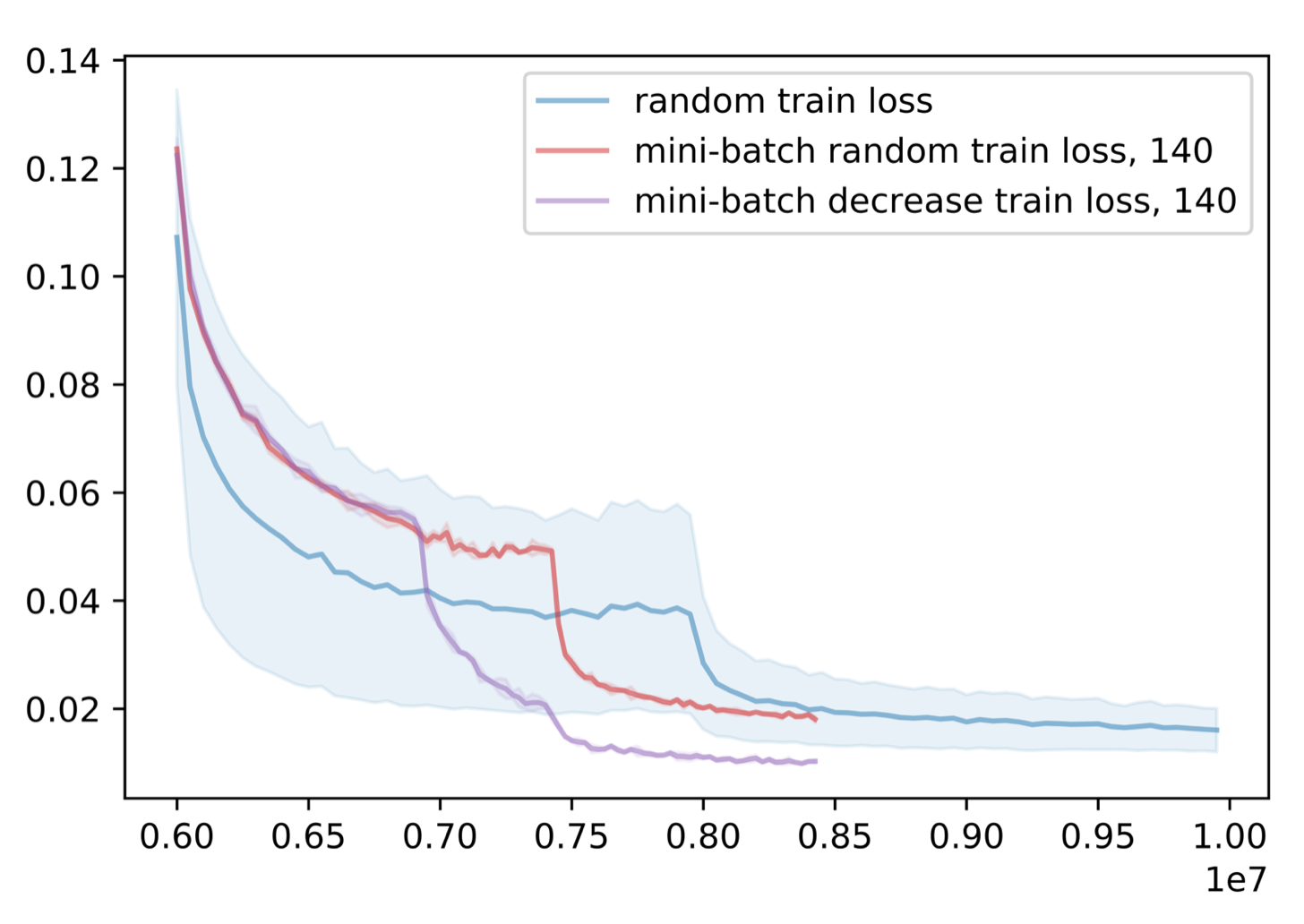}
         \caption{CIFAR-10 Training Loss}
         \label{fig:cifar10_train}
     \end{subfigure}
     \begin{subfigure}[b]{0.30\textwidth}
         \includegraphics[width=\textwidth]{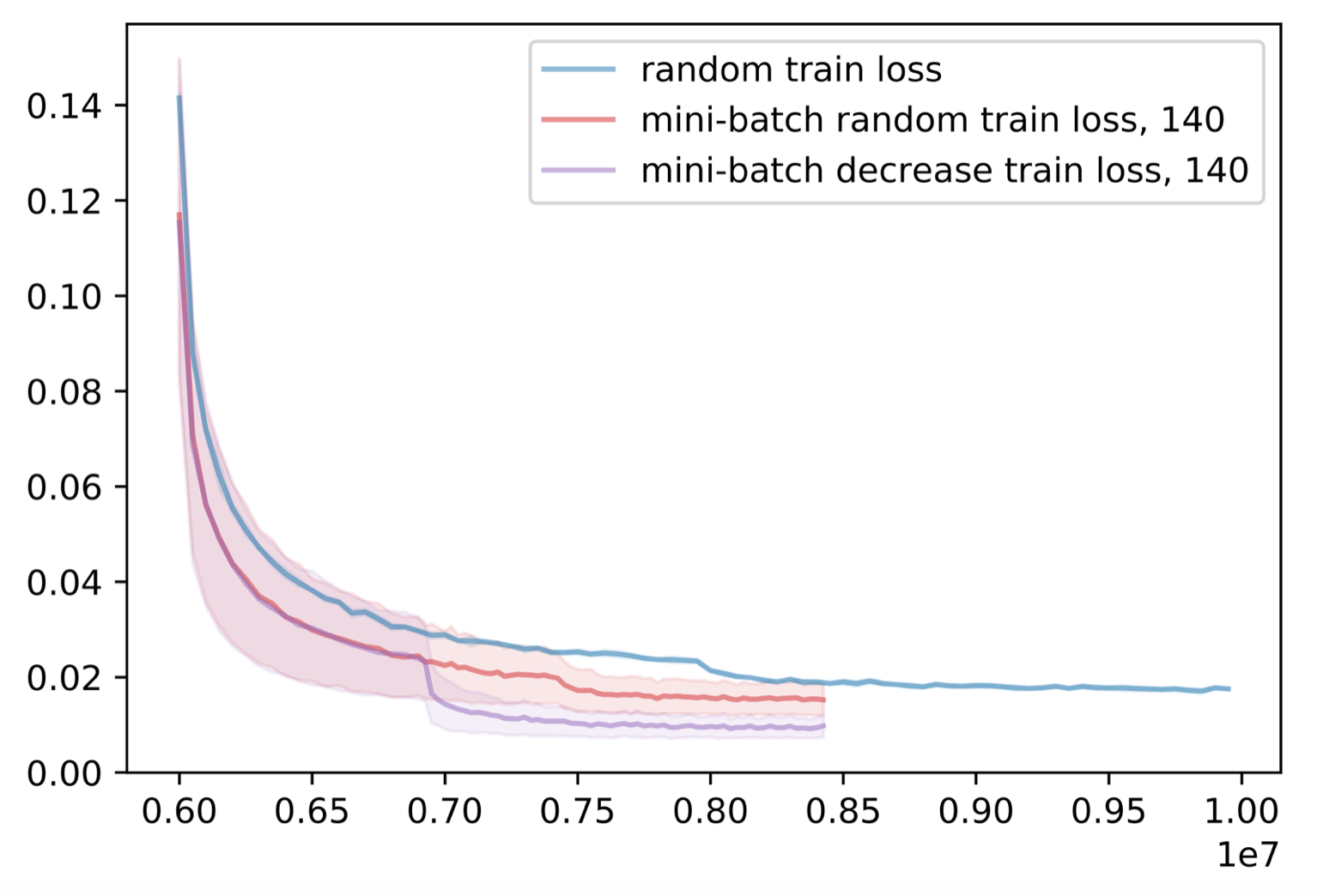}
         \caption{CIFAR-100 Training Loss}
         \label{fig:cifar100_train}
     \end{subfigure}\\
     \begin{subfigure}[b]{0.30\textwidth}
         \includegraphics[width=\textwidth]{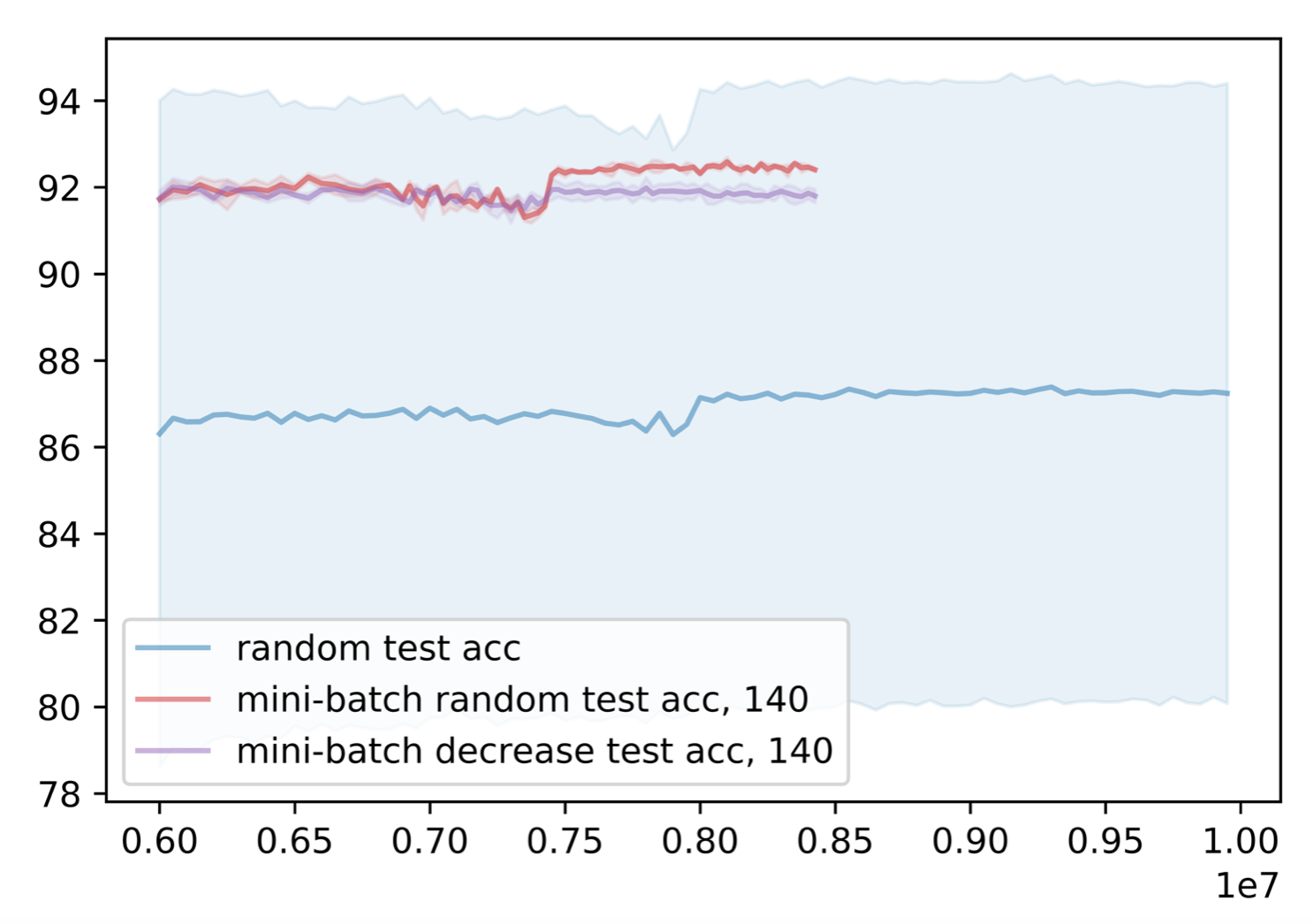}
         \caption{CIFAR-10 Test Acc}
         \label{fig:cifar10_test}
     \end{subfigure}
     \begin{subfigure}[b]{0.30\textwidth}
         \includegraphics[width=\textwidth]{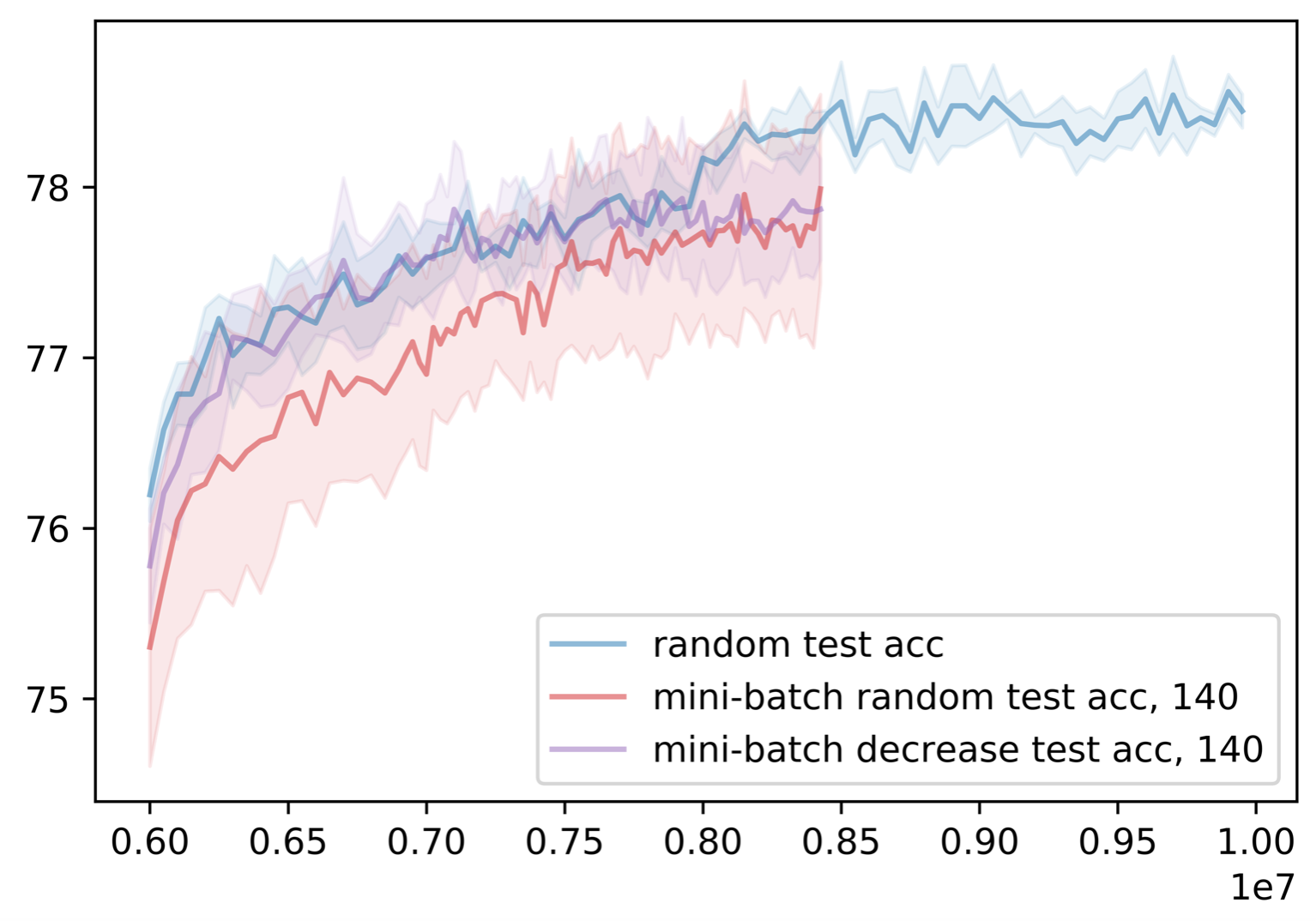}
         \caption{CIFAR-100 Test Acc}
         \label{fig:cifar100_test}
     \end{subfigure}
        \caption{Ordering Starting at Epoch 140}
        \label{fig:CIFAR_140_baseline}
\end{figure}

\begin{center}
\begin{tabular}{c c c c c} \hline
  & Train Loss & Train Acc & Test Loss & Test Acc \\
 \hline
 CIFAR-10 full random & 0.019 & 99.72\% & 0.247 & 92.56 \% \\ 
 \hline
 full ordered & 0.014 & 99.81\% & 0.261 & 92.09 \% \\ 
 \hline
 mini random & 0.017 & 99.70\% & 0.253 & 92.72 \% \\ 
 \hline
 mini ordered & 0.010 & 99.94\% & 0.243 & 92.31 \% \\ 
 \hline
 CIFAR-100 full random & 0.017 & 99.93\% & 0.90 & 78.86 \% \\ 
 \hline
 full ordered & 0.013 & 99.98\% & 0.89 & 78.80 \% \\ 
 \hline
 mini random & 0.017 & 99.94\% & 0.91 & 78.71 \% \\ 
 \hline
 mini ordered & 0.010 & 99.99\% & 0.92 & 78.46 \% \\ 
 \hline
\end{tabular}
\end{center}
\hfill \break
In addition, if we adjust the mini-batch ordering to begin at smaller epochs (epoch 80 for CIFAR-10 and epoch 100 for CIFAR-100), we can arrive at approximately similar low training error and high testing accuracy with significantly fewer epochs. Though with this approach the test accuracy is not the highest, the difference is negligible compared to the amount of time and calculation saved in training, at least $10^6$ iterations. (Full graphs can be found in Appendix \ref{exp:nn_cifar}).\\
\begin{figure}[htbp] \centering
     \begin{subfigure}[b]{0.30\textwidth}
         \includegraphics[width=\textwidth]{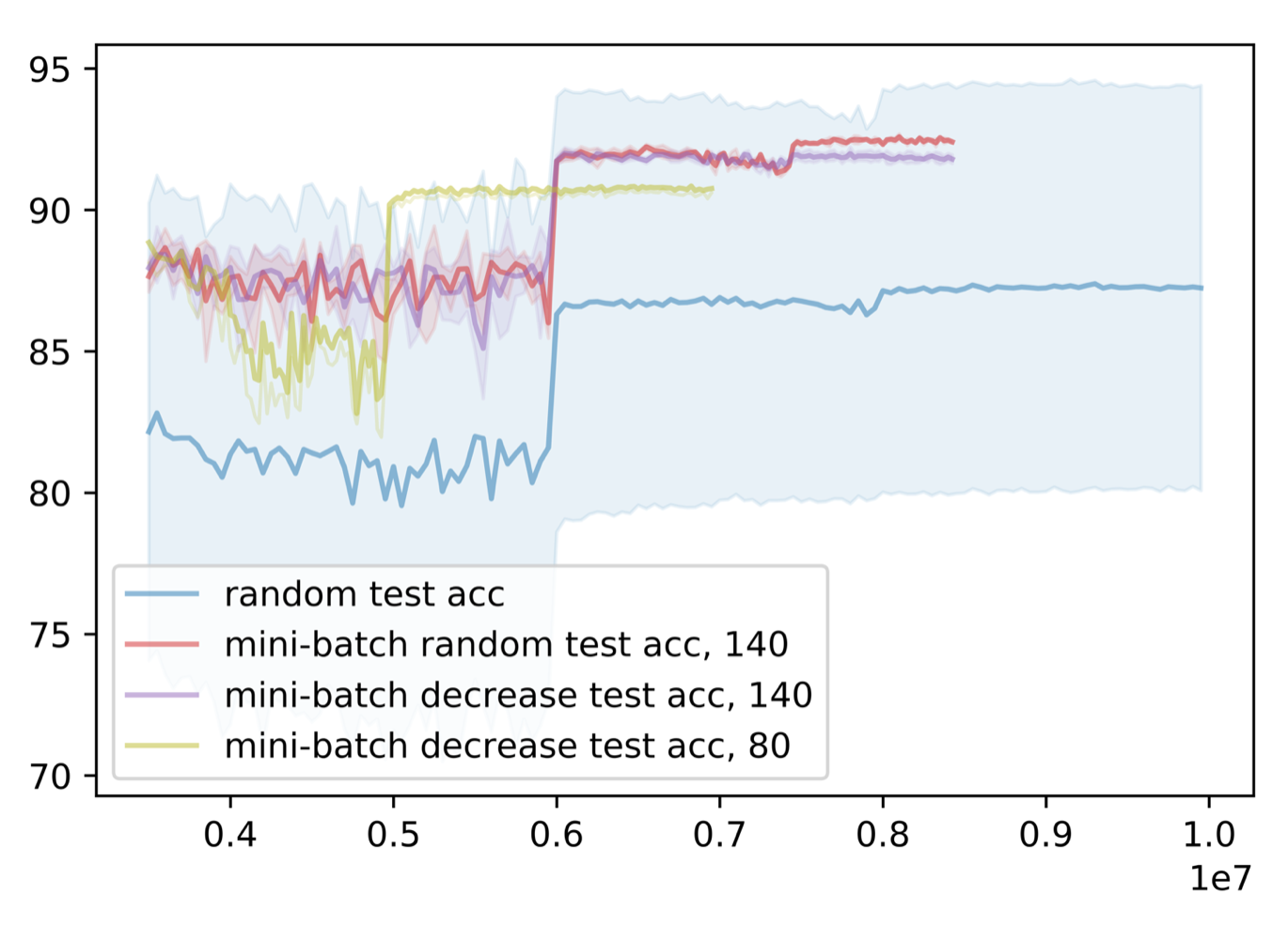}
         \caption{CIFAR-10 Test Acc, Start Epoch 80}
         \label{fig:cifar10_80_test}
     \end{subfigure}
      \begin{subfigure}[b]{0.30\textwidth}
         \includegraphics[width=\textwidth]{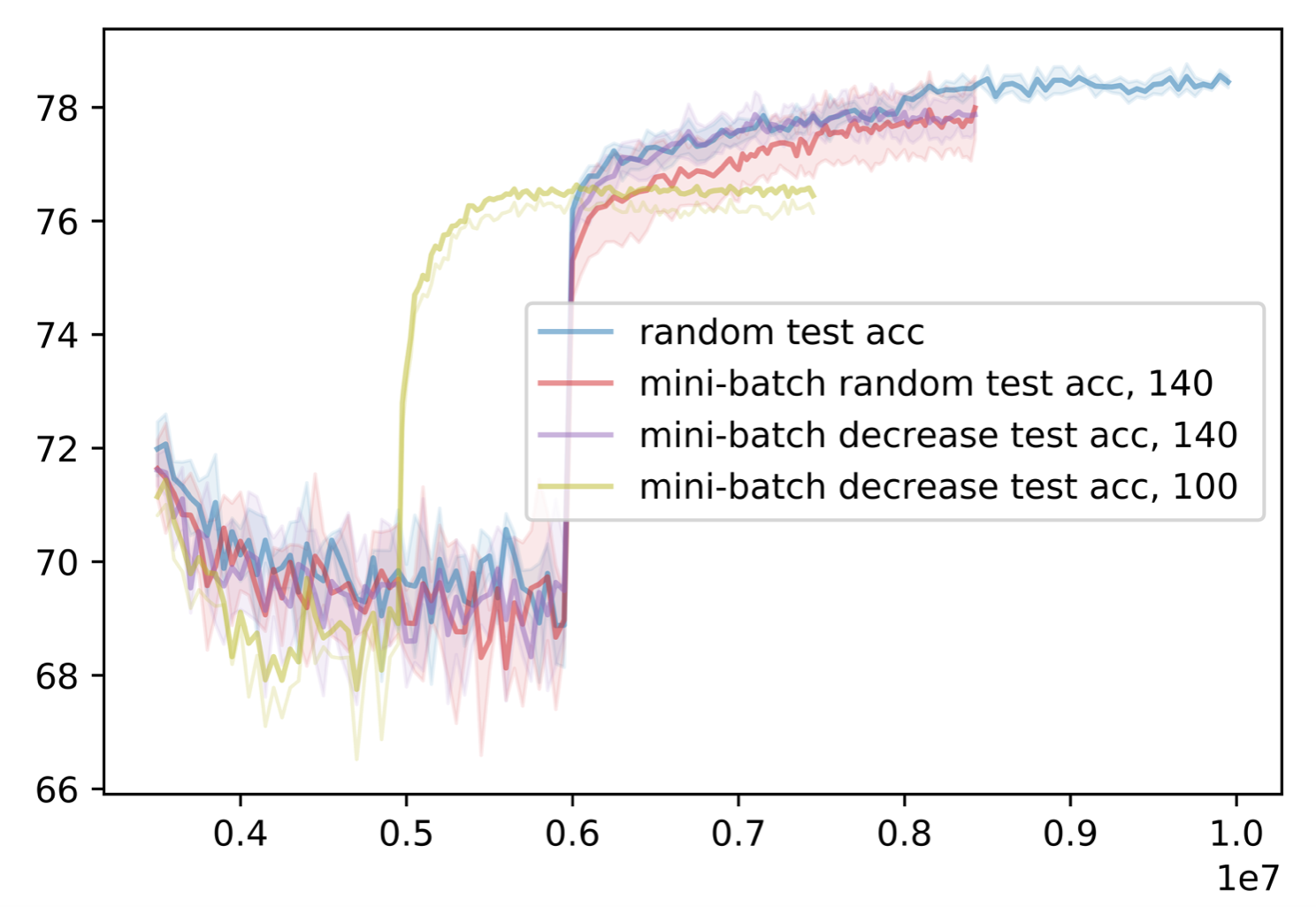}
         \caption{CIFAR-100 Test Acc, Start Epoch 100}
         \label{fig:cifar100_100_test}
     \end{subfigure}
        \caption{Ordering Starting at Earlier Epochs}
        \label{fig:CIFAR_early_start}
\end{figure}
\section{Conclusion}
In this paper, we explored the optimal ordering for shuffling algorithms which seek to improve the convergence of random shuffling: For decreasing step size per iteration, ordering by decreasing gradient at the beginning of the epoch results in a faster convergence rate; for constant step size, all orderings results in the same convergence rate. From experimental results we indeed verified the orderings and confirmed that the initial point should be sufficiently close to the optimal. Through comparison of different step size conditions we could observe for decreasing step size per epoch, the behavior varies for different data sets which suggests further analysis in the future. We also extended our results to mini-batch algorithms and non-convex settings with neural networks. Through empirical result, our ordering can be extended to these more complex settings. Further work is needed to theoretically explain the phenomenon.

\nocite{*}
\printbibliography
\pagebreak
\section{Appendix}
\subsection{Proof of Lemma 1} \label{pf:lemma_1}
\begin{proof}
By property of convex functions, we have for $\forall x$
\begin{equation}
\begin{array}{l@{}l}
    f_i(x^*) &\geq f_i(x)+\nabla f_i(x)^T(x^* - x)\\
    f_i(x^*)-f_i(x) &\geq \nabla f_i(x)^T(x^* - x)\\
    f_i(x) - f_i(x^*) &\leq \nabla f_i(x)^T(x - x^*)\\
    f_i(x) - f_i(x^*)&\leq ||\nabla f_i(x)^T(x - x^*)|| \leq ||\nabla f_i(x)^T||||(x - x^*)||
\end{array}
\end{equation}
Here we used the property that $f_i(x) \geq f_i(x^*)$ for all i, and $||AB|| \leq||A||||B||$.\\
Since the $||x-x^*||$ is constant at each iteration, then the larger the $||\nabla f_i(x)^T||$, the larger the upper bound for $f_i(x) - f_i(x^*)$.\\
\end{proof}
\subsection*{Proof of Theorems 1 and 2} \label{pf:Theorem_1_2}
\begin{proof}
Following the proof from \textit{Convergence Rate of Incremental Subgradient Algorithms}(\cite{nedic2001convergence}), we have the following:
\begin{equation}
\begin{array}{l@{}l}
    ||x_{i,k}-x^*||^2 &= ||P_x[x_{i-1,k}-\alpha_{i,k}g_{i,k}] - x^*||^2\\
    &{}\leq ||x_{i-1,k}-\alpha_{i,k}g_{i,k} - x^*||^2\\
    &{}\leq ||x_{i-1,k} - x^*||^2 - 2\alpha_{i,k}g_{i,k}\cdot(x_{i-1,k}-x^*) + \alpha_{i,k}^2g_{i,k}^2\\
    &{}\leq ||x_{i-1,k} - x^*||^2 - 2\alpha_{i,k}g_{i,k}\cdot(x_{i-1,k}-x^*) + \alpha_{k}^2C_i^2\\
    &{}\leq ||x_{i-1,k} - x^*||^2 - 2\alpha_{i,k}\cdot(f_i(x_{i-1,k})-f_i(x^*)) + \alpha_{k}^2C_i^2\\
\end{array}
\end{equation}
\\
If we add the above relation for all points within a epoch we get:
\begin{equation}
    ||x_{k+1}-x^*||^2\leq ||x_{k} - x^*||^2 - 2\sum_{i = 1}^n[\alpha_{i,k}\cdot(f_i(x_{i-1,k})-f_i(x^*))] + \alpha_{k}^2\sum_{i = 1}^n C^2
\end{equation}
\\
Since $\alpha_{i,k}$ is decreasing, to minimize the upper bound for $||x_{k+1}-x^*||^2$, we have to maximize $f_i(x_{i-1,k})-f_i(x^*)$ for each iteration and have the larger values be in the beginning with the larger $\alpha_{i,k}$. Hence it's reasonable to have the $f_i(x_{i-1,k})-f_i(x^*)$ with larger upper bounds in the beginning.\\

To maximize $\sum_{i = 1}^n[\alpha_{i,k}\cdot(f_i(x_{i-1,k})-f_i(x^*))]$, we will order the $f_i$ such that the $f_i$'s with larger values of $f_i(x_{i-1}) - f_i(x^*)$ are iterated first.\\
Let $|M_1| > |M_2| > |M_3| > ... > |M_n|$ be the magnitude of the sub-gradients in the beginning of the epoch, and denote $f_1, f_2, f_3, ..., f_n$ the corresponding function for those sub-gradients.\\

For the first iteration, it's obvious that choosing $f_1$ with the largest $|M_i|$ would maximize upper bound for $f_i(x_0)-f_i(x^*)$.\\
For the following iterations, instead of directly finding the maximum value for $f_i(x)-f_i(x^*)$, we compare the lower bound $f_i(x) - f_i(x^*)$. Through ordering by that value, we can bound the minimum value of $f_i(x_{i-1,k})-f_i(x^*)$.\\
\\
For the second iteration:\\
we first compare $f_2$ and $f_3$. Note $x_{1,k} = x_{0,k}-\alpha_{1,k}M_1$.
\begin{equation}
\begin{array}{l@{}l}
    f_2(x_{0,k}-\alpha_{1,k}M_1) - f_2(x^*) &\geq f_2(x_{0,k}) - \nabla f_2(x_{0,k})^T(\alpha_{1,k}M_1) + \frac{m_2}{2}||\alpha_{1,k}M_1||^2 - f_2(x^*)\\
    &{}\geq [f_2(x_{0,k})- f_2(x^*)] - \alpha_{1,k}||M_1||^2 + \frac{m_2\alpha_{1,k}^2}{2}||M_1||^2\\
    f_3(x_{0,k}-\alpha_{1,k}M_1)- f_3(x^*) &\geq f_3(x_{0,k}) - \nabla f_3(x_{0,k})^T(\alpha_{1,k}M_1) + \frac{m_3}{2}||\alpha_{1,k}M_1||^2- f_3(x^*)\\
    &{}\geq [f_3(x_{0,k})- f_3(x^*)] - \alpha_{1,k}||M_1||^2 + \frac{m_3\alpha_{1,k}^2}{2}||M_1||^2\\
\end{array}
\end{equation}
Since $\alpha$ is small, $\alpha^2$ is insignificant compared with the rest of the terms, and can be treated as a error.\\
Since $||M_2||\geq||M_3||$,\\ $upperbound(f_2(x_{0,k})- f_2(x^*)) \geq upperbound(f_3(x_{0,k})- f_3(x^*))$\\
Hence, at the second iteration, we know $f_2(x_{1,k})- f_2(x^*)$ more likely has a relative larger value for the lower bound than $f_3(x_{1,k})- f_3(x^*)$. Following the same logic, we can show relatively $lowerbound(f_2(x_{1,k})- f_2(x^*))$ more likely have a larger value than $lowerbound(f_k(x_{1,k})- f_k(x^*))$ for $k >2$.
Thus it is optimal to have $f_2$ at the second iteration.\\\\
Assume the above relation hold for iterations 1, 2, ..., N-1.\\
For the $N^{th}$ iteration:\\
$x_{N,k} = x_{0,k} - \sum_{i = 1}^{N-1} \alpha_{i,k}M_i^{'}$ ($M_i^{'}$ is the sub-gradient value in iteration i).\\
\begin{equation}
\begin{array}{l@{}}
    f_N(x_{0,k} - \sum_{i = 1}^{N-1} \alpha_{i,k}M_i^{'}) - f_N(x^*) \\
    \geq f_N(x_{0,k}) - \nabla f_N(x_{0,k})^T(\sum_{i = 1}^{N-1} \alpha_{i,k}M_i^{'}) + \frac{m_N}{2}\sum_{i = 1}^{N-1} ||\alpha_{i,k}M_i^{'}||^2 - f_N(x^*)\\
    \geq [f_N(x_{0,k}) - f_N(x^*)] - \sum_{i = 1}^{N-1}\alpha_{i,k}\max{(||M_i||,||M_i^{'}||)}^2 + \frac{m_N}{2}\sum_{i = 1}^{N-1} ||\alpha_{i,k}M_i^{'}||^2\\\\
    f_{N+1}(x_{0,k} - \sum_{i = 1}^{N-1} \alpha_{i,k}M_i^{'}) - f_{N+1}(x^*) \\
    \geq f_{N+1}(x_{0,k}) - \nabla f_{N+1}(x_{0,k})^T(\sum_{i = 1}^{N-1} \alpha_{i,k}M_i^{'}) + \frac{m_{N+1}}{2}\sum_{i = 1}^{N-1} ||\alpha_{i,k}M_i^{'}||^2 - f_{N+1}(x^*)\\
    \geq [f_{N+1}(x_{0,k}) - f_{N+1}(x^*)] - \sum_{i = 1}^{N-1}\alpha_{i,k}\max{(||M_i||,||M_i^{'}||)}^2 + \frac{m_{N+1}}{2}\sum_{i = 1}^{N-1} ||\alpha_{i,k}M_i^{'}||^2\\
\end{array}
\end{equation}
Following the same argument as the second iteration we can see $lowerbound(f_N(x_{N-1,k}) - f_N(x^*))$ has a larger value than $lowerbound(f_{N+1}(x_{N-1,k}) - f_{N+1}(x^*))$, and if we compare $f_N$ with $f_k$ where $k > N$, we can get $lowerbound(f_N(x_{N-1,k}) - f_N(x^*))$ has a larger amount than $lowerbound(f_k(x_{N-1,k}) - f_k(x^*))$.\\
Thus it is optimal to have $f_N$ at the $N^{th}$ iteration.\\\\
Thus by induction, through ordering the iteration by the sub-gradient value at the beginning of the epoch we maximize the lower bound of $f_i(x_{i-1,k})$. This in turn help us derive a stricter upper bound for $||x_{k+1}-x^*||^2$.\\
Plugging the above bound in formula (9), we get:
\begin{equation}
\begin{array}{l@{}l}
||x_{k+1}-x^*||^2\\
\leq ||x_{k} - x^*||^2 - 2\sum_{i = 1}^n[\alpha_{i,k}\cdot(f_i(x_{i-1,k})-f_i(x^*))] + \alpha_{k}^2\sum_{i  = 1}^n C_i^2\\
\leq ||x_{k} - x^*||^2 - 2\sum_{i = 1}^n \alpha_{i,k}\left((f_i(x_{0,k}) - f_i(x^*)) - \sum_{j = 1}^{i-1}\alpha_{j,k}\max{(||M_j||,||M_j^{'}||)}^2 + \frac{m_i}{2}\sum_{j = 1}^{i-1} ||\alpha_{j,k}M_j^{'}||^2\right)\\ 
+ \alpha_{k}^2\sum_{i  = 1}^n C_i^2\\
\leq ||x_{k} - x^*||^2 - 2\sum_{i = 1}^n\alpha_{i,k}(f_i(x_k) - f_i(x^*)) +\sum_{i=1}^n \left(\alpha_{i,k}\sum_{j = 1}^{i-1} \max{(||M_j||,||M_j^{'}||)}^2(2\alpha_{j,k}-m_i\alpha_{j,k}^2)\right) \\+\alpha_{k}^2\sum_{i  = 1}^n C_i^2\\
\leq ||x_{k} - x^*||^2 - 2\sum_{i = 1}^n\alpha_{i,k}(||M_i||||x_k-x^*||) + \alpha_{k}^2\left(2\sum_{i = 1}^{n-1}[(n-i)C_i^2] + \sum_{i = 1}^n C_i^2 \right) \\
-\alpha_{k}^2\alpha_{k+1}\sum_{i = 1}^{n-1}(n-i)m_i\cdot max{(||M_i||,||M_i^{'}||)}^2 + n\alpha_k\cdot\epsilon_k\\
\end{array}
\end{equation}
\\
where $\epsilon_k$ is the max error between $f_i(x_k) - f_i(x^*)$ and $||M_i||||x_k-x^*||$ in epoch k. We will assume that $\epsilon_k$ is small so that $m\alpha_k\cdot\epsilon_k$ is negligible.\\ 
For the case with decreasing step size per iteration:
\begin{equation}
\begin{array}{l@{}l}
||x_{k+1}-x^*||^2\leq ||x_{k} - x^*||^2 - 2||x_k-x^*||\sum_{i = 1}^n\alpha_{i,k}||M_i|| + \alpha_{k}^2\left(2\sum_{i = 1}^{n-1}[(n-i)C_i^2] + \sum_{i = 1}^n C_i^2 \right) \\
- \alpha_{k}^2\alpha_{k+1}\sum_{i = 1}^{n-1}(n-i)m_i\cdot max{(||M_i||,||M_i^{'}||)}^2 + n\alpha_k\cdot\epsilon_k
\end{array}
\end{equation}
In this case, since $\alpha_{i,k}$ decreases with each iteration, to minimize the upper bound we need to maximize $\sum_{i = 1}^n\alpha_{i,k}||M_i||$, which is attained when we order by decreasing order of the gradient at the beginning of the epoch. Also with the decreasing order, in expectation $\sum_{i = 1}^n C_i^2$ is the smallest. The $\alpha_{k}^2$ and $\alpha_{k}^2\alpha_{k+1}$ are so small in magnitude compared to the prior terms that its value is negligible. Hence such ordering minimizes the upper bound, suggesting better convergence rate compared to other ordering.\\
\\
For the case with constant step size, the expression can be simplified as:
\begin{equation}
\begin{array}{l@{}l}
||x_{k+1}-x^*||^2\leq ||x_{k} - x^*||^2 - 2\alpha ||x_k-x^*|| \sum_{i = 1}^n ||M_i|| + \alpha^2\left(2\sum_{i = 1}^{n-1}[(n-i)C_i^2] + \sum_{i = 1}^n C_i^2 \right) \\
-\alpha^3\sum_{i = 1}^{n-1}(n-i)m_i\cdot max{(||M_i||,||M_i^{'}||)}^2+ n\alpha\cdot\epsilon_k
\end{array}
\end{equation}
In this case, $||x_k-x^*||$ and $\alpha$ are constant within an epoch. Though in expectation $\sum_{i = 1}^n C_i^2$ is the smallest when we order by decreasing gradient order, the $\alpha^2$ and $\alpha^3$ are so small in magnitude compared to the prior terms that the different is negligible. Hence, no matter how we order the data points, the upper bound after an epoch is approximately the same, suggesting similar convergence rate.\\
\end{proof}
\subsection{Proof of Theorems 3 and 4} \label{pf:Theorem_3_4}
\begin{proof}
Following the same logic as the proof form Theorems 1 and 2, we result:
\begin{equation}
\begin{array}{l@{}l}
||x_{k+1}-x^*||^2\\
\leq ||x_{k} - x^*||^2 - 2\sum_{i = 1}^n[\alpha_{i,k}\cdot(f_i(x_{i-1,k})-f_i(x^*))] + \alpha_{k}^2\sum_{i  = 1}^n C_i^2\\
\leq ||x_{k} - x^*||^2 - 2\sum_{i = 1}^n \alpha_{i,k}\left((f_i(x_{0,k}) - f_i(x^*)) - \sum_{j = 1}^{i-1}\alpha_{j,k}\max{(||M_j||,||M_j^{'}||)}^2 \right)+ \alpha_{k}^2\sum_{i  = 1}^n C_i^2\\
\leq ||x_{k} - x^*||^2 - 2\sum_{i = 1}^n\alpha_{i,k}(f_i(x_k) - f_i(x^*)) +\sum_{i=1}^n \left(2\alpha_{i,k}\sum_{j = 1}^{i-1} \alpha_{j,k} \max{(||M_j||,||M_j^{'}||)}^2 \right)+\alpha_{k}^2\sum_{i  = 1}^n C_i^2\\
\leq ||x_{k} - x^*||^2 - 2\sum_{i = 1}^n\alpha_{i,k}(||M_i||||x_k-x^*||) + \alpha_{k}^2\left(2\sum_{i = 1}^{n-1}[(n-i)C_i^2] + \sum_{i = 1}^n C_i^2 \right) + n\alpha_k\cdot\epsilon_k\\
\end{array}
\end{equation}
\\
For the case with decreasing step size per iteration:
\begin{equation}
\begin{array}{l@{}l}
||x_{k+1}-x^*||^2\leq ||x_{k} - x^*||^2 - 2||x_k-x^*||\sum_{i = 1}^n\alpha_{i,k}||M_i||\\ \quad \quad \quad \quad +\alpha_{k}^2\left(2\sum_{i = 1}^{n-1}[(n-i)C_i^2] + \sum_{i = 1}^n C_i^2 \right) + n\alpha_k\cdot\epsilon_k
\end{array}
\end{equation}
As in the strongly convex case, the best ordering that minimizes the upper bound is decreasing order of the gradients at the beginning of the epoch.
\hfill \break
For the case with constant step size, the expression can be simplified as:
\begin{equation}
\begin{array}{l@{}l}
||x_{k+1}-x^*||^2\leq ||x_{k} - x^*||^2 - 2\alpha ||x_k-x^*|| \sum_{i = 1}^n ||M_i|| \\ \quad \quad \quad \quad + \alpha^2\left(2\sum_{i = 1}^{n-1}[(n-i)C_i^2] + \sum_{i = 1}^n C_i^2 \right) + n\alpha\cdot\epsilon_k
\end{array}
\end{equation}
Following similar logic as the strongly convex case, all orderings for constant step size results in similar value of the upper bound.
\end{proof}

\subsection{Distance from the Optimal} \label{exp:dist_optimal}
\begin{figure}[htbp] \centering
     \begin{subfigure}[b]{0.30\textwidth}
         \includegraphics[width=\textwidth]{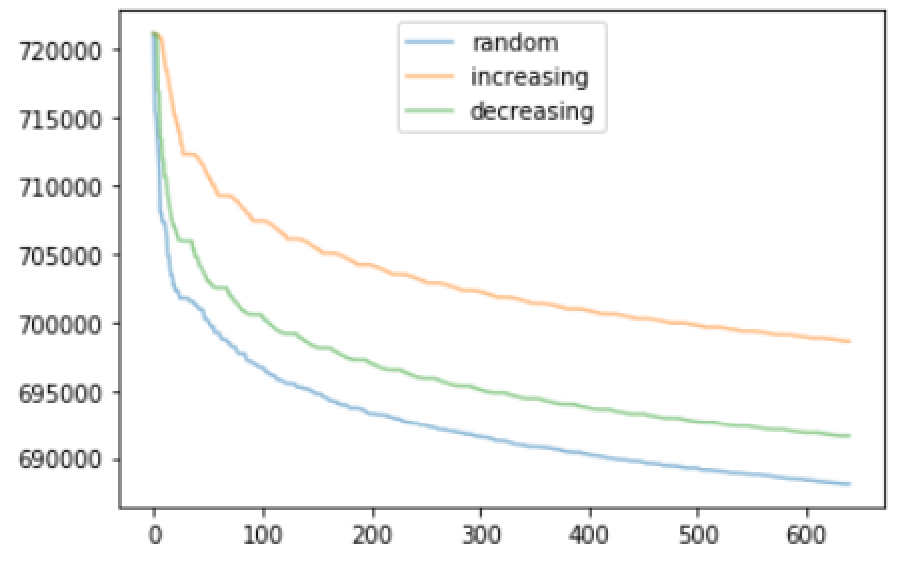}
         \caption{decreasing per iteration; far from optimal}
         \label{fig:syn_decrase_distance}
     \end{subfigure}
     \begin{subfigure}[b]{0.30\textwidth}
         \includegraphics[width=\textwidth]{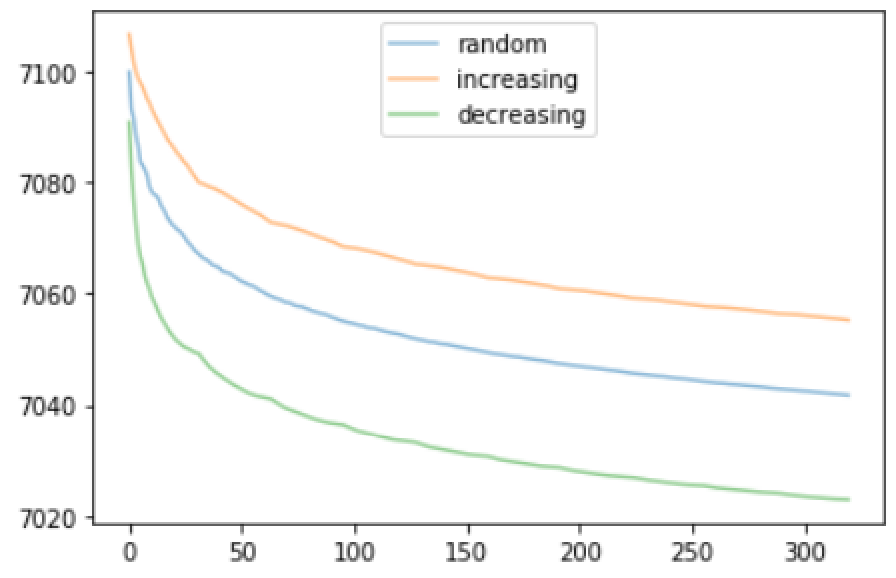}
         \caption{decreasing step size; closer to optimal}
         \label{fig:syn_diff_random}
     \end{subfigure}
        \caption{Synthetic Data (continued)}
        \label{fig:synthesis two graphs}
\end{figure}

Here, we resumed to the original loss function for the synthesis data set and tried a new algorithm: first run random shuffling till the iterate is relative closer to the optimal and then switch to the proposed ordering. For the synthesis data, we ran 15 runs of random shuffling and then ran 10 runs of the assigned ordering. Observing the results shown in figure \ref{fig:syn_diff_random}, ordering by decreasing initial gradient results in the smallest loss value, which further confirms our proposed ordering optimizes the convergence rate.
\subsection{Full Results for Data Selection Algorithm Experiments} \label{exp:data_selec}
The results below are mini-batch algorithms applied to synthesis data set, Iris data set, and Boston Housing data set correspondingly.\\\\
For the $9x9$ ordered graphs, the rows correspond to the sizes of mini-match with respect to total data ($S = 0.3, 0.6, 0.8$), and the columns correspond to the training size within each mini-batch with respect to the batch size ($q = 0.18, 0.5, 0.8$).
\begin{figure}[htbp]\centering
    \includegraphics[width=0.8\textwidth]{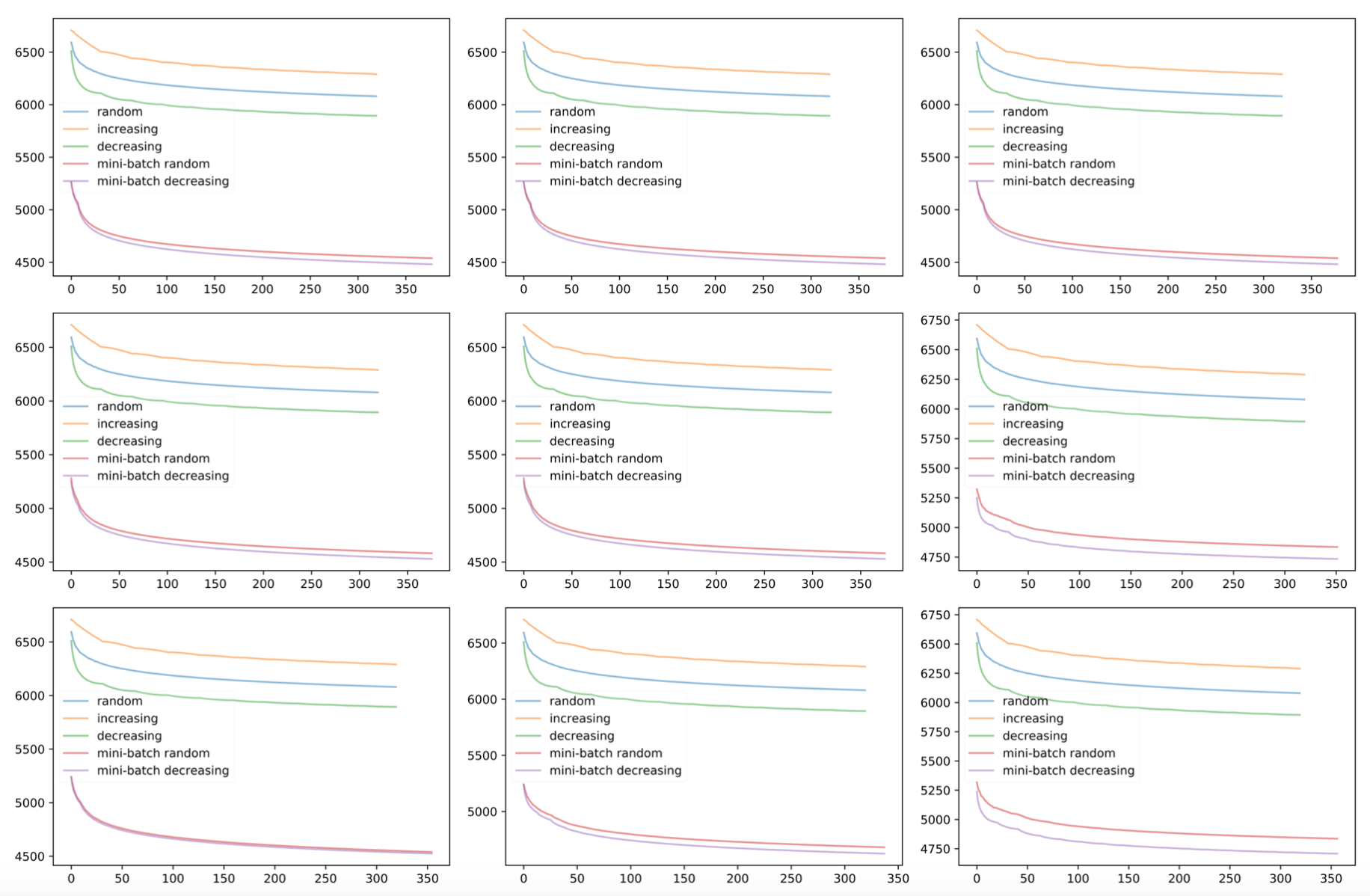}
    \caption{Synthesis, decreasing step size}
    \label{fig:syn_mini_decrease}
\end{figure}
\begin{figure}[htbp]\centering
    \includegraphics[width=0.8\textwidth]{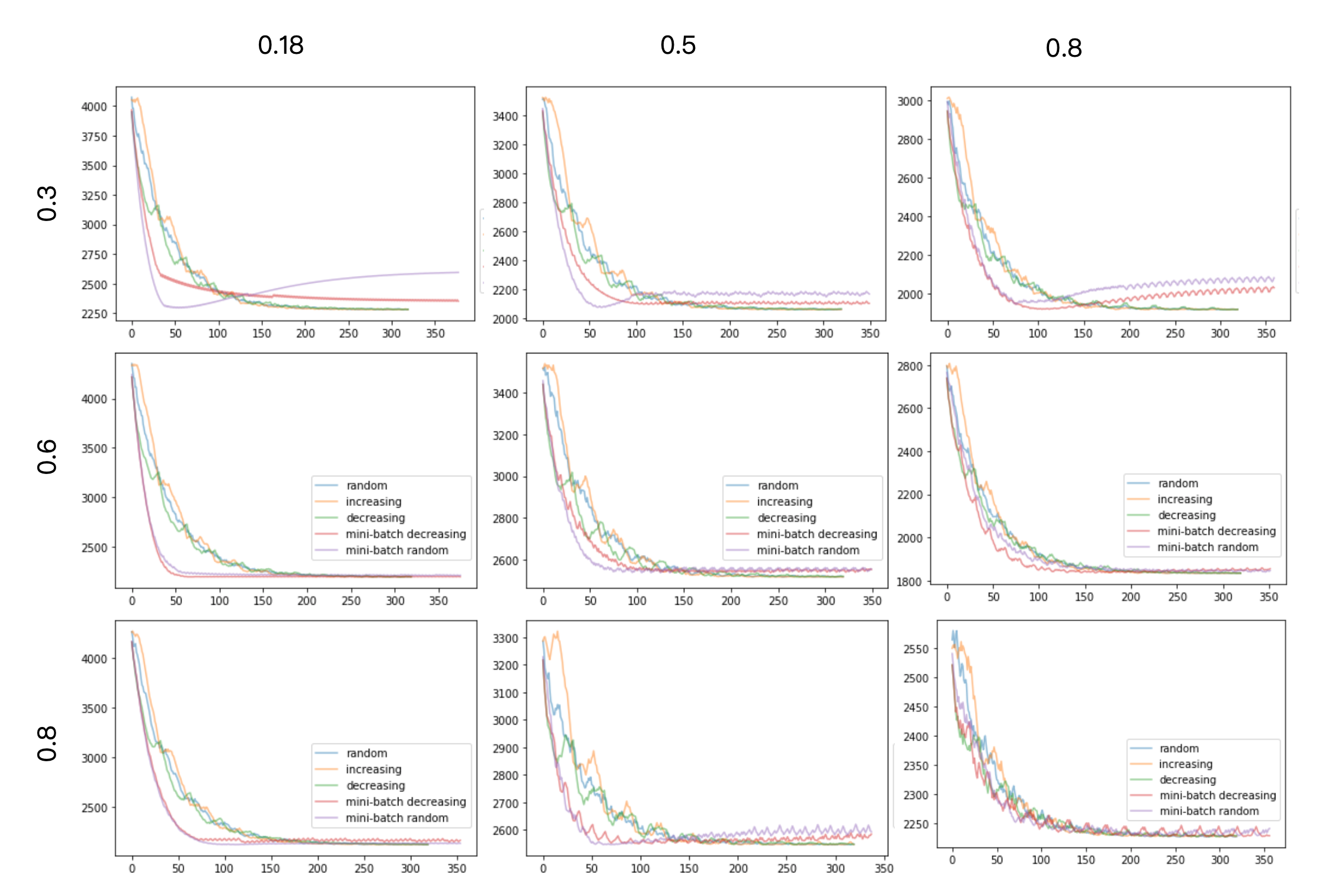}
    \caption{Synthesis, constant step size}
    \label{fig:syn_mini_const}
\end{figure}

\begin{figure}[htbp]\centering
         \includegraphics[width=0.8\textwidth]{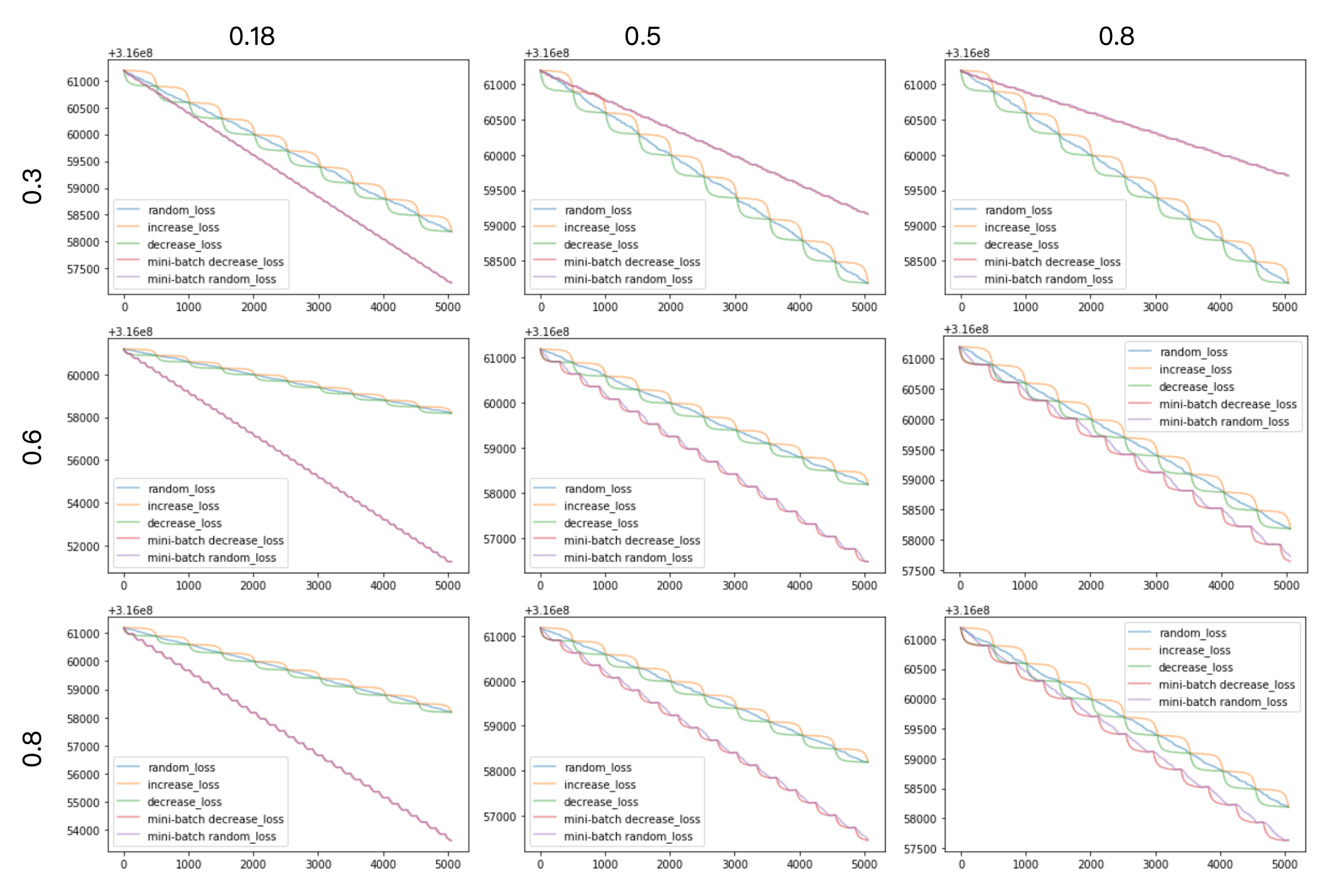}
         \caption{Boston Housing, constant step size}
         \label{fig:boston_mini_const}
\end{figure}

\begin{figure}[htbp]\centering
    \begin{subfigure}[b]{0.8\textwidth}
    \includegraphics[width=\textwidth]{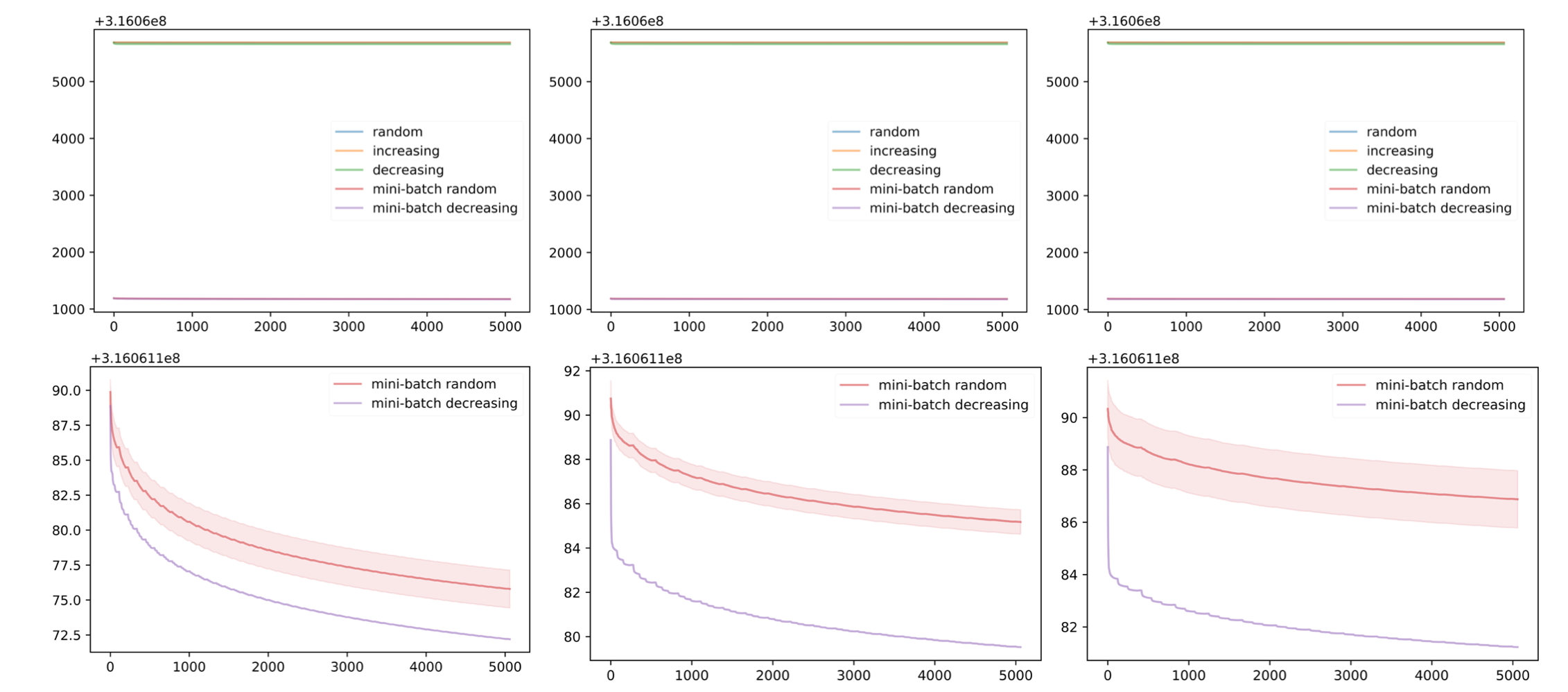}
         \caption{S = 0.3}
         \label{fig:boston_mini_decrease_1}
     \end{subfigure}
    \begin{subfigure}[b]{0.8\textwidth}
    \includegraphics[width=\textwidth]{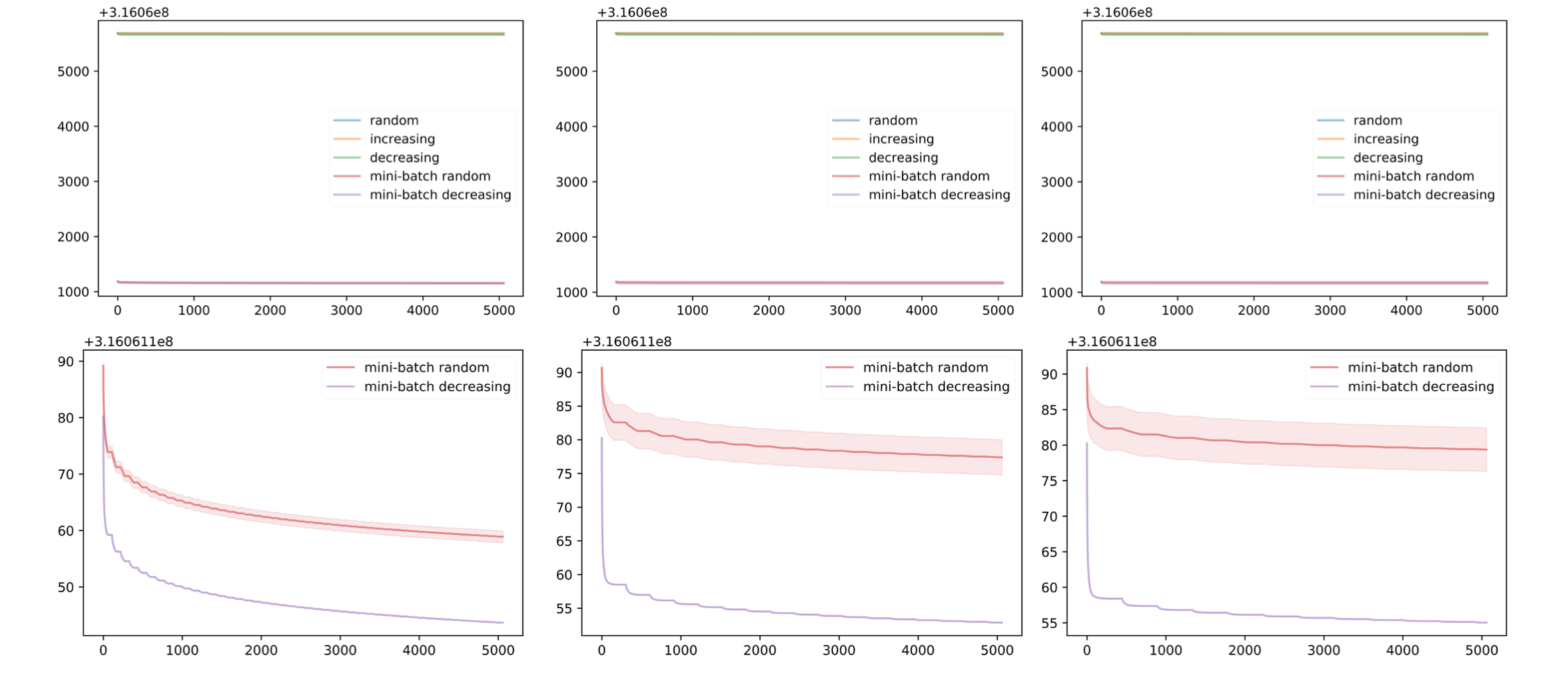}
         \caption{S = 0.6}
         \label{fig:boston_mini_decrease_2}
     \end{subfigure}
     \begin{subfigure}[b]{0.8\textwidth}
    \includegraphics[width=\textwidth]{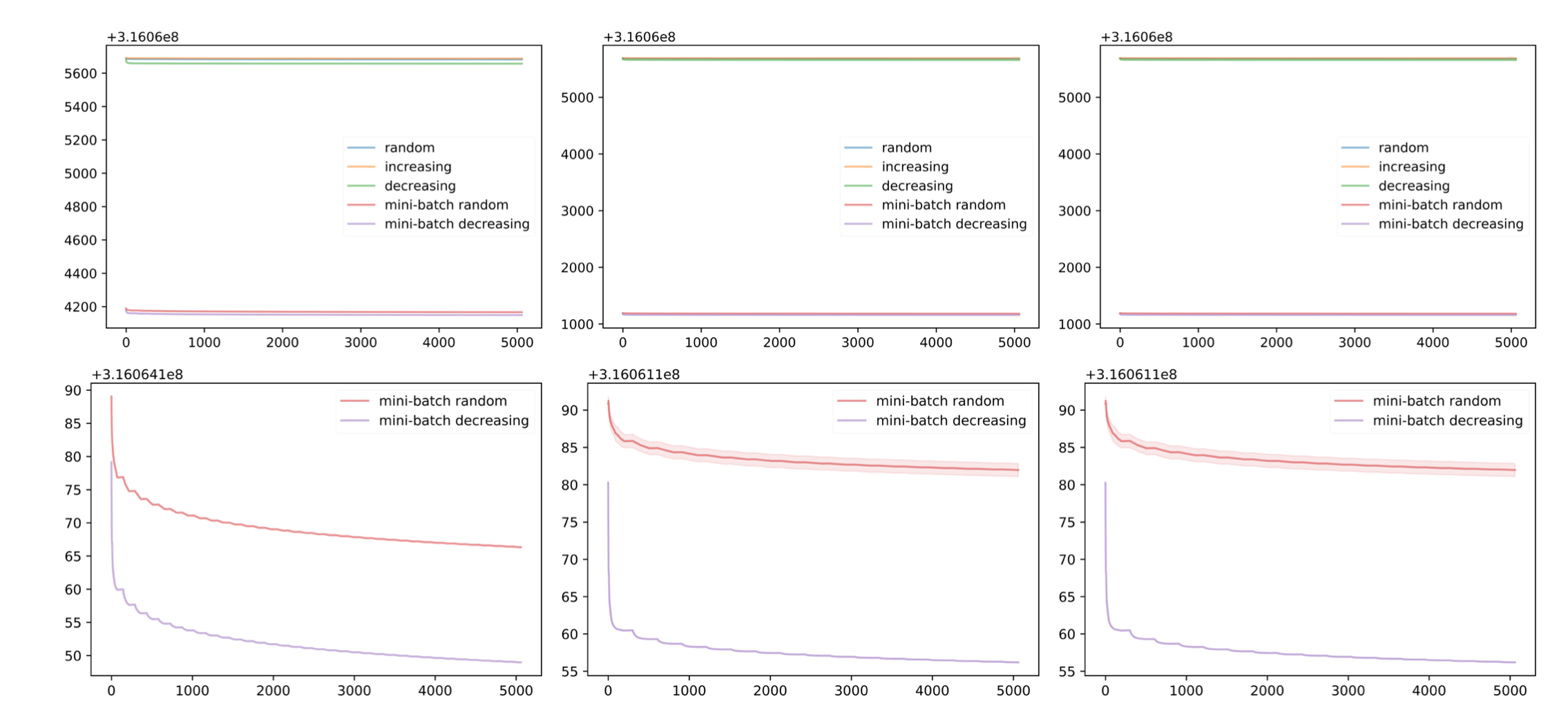}
         \caption{S = 0.8}
         \label{fig:boston_mini_decrease_3}
     \end{subfigure}
    \caption{Boston Housing, decreasing step size}
    \label{fig:boston_mini_decrease}
\end{figure}

\begin{figure}[htbp]\centering
    \includegraphics[width=0.8\textwidth]{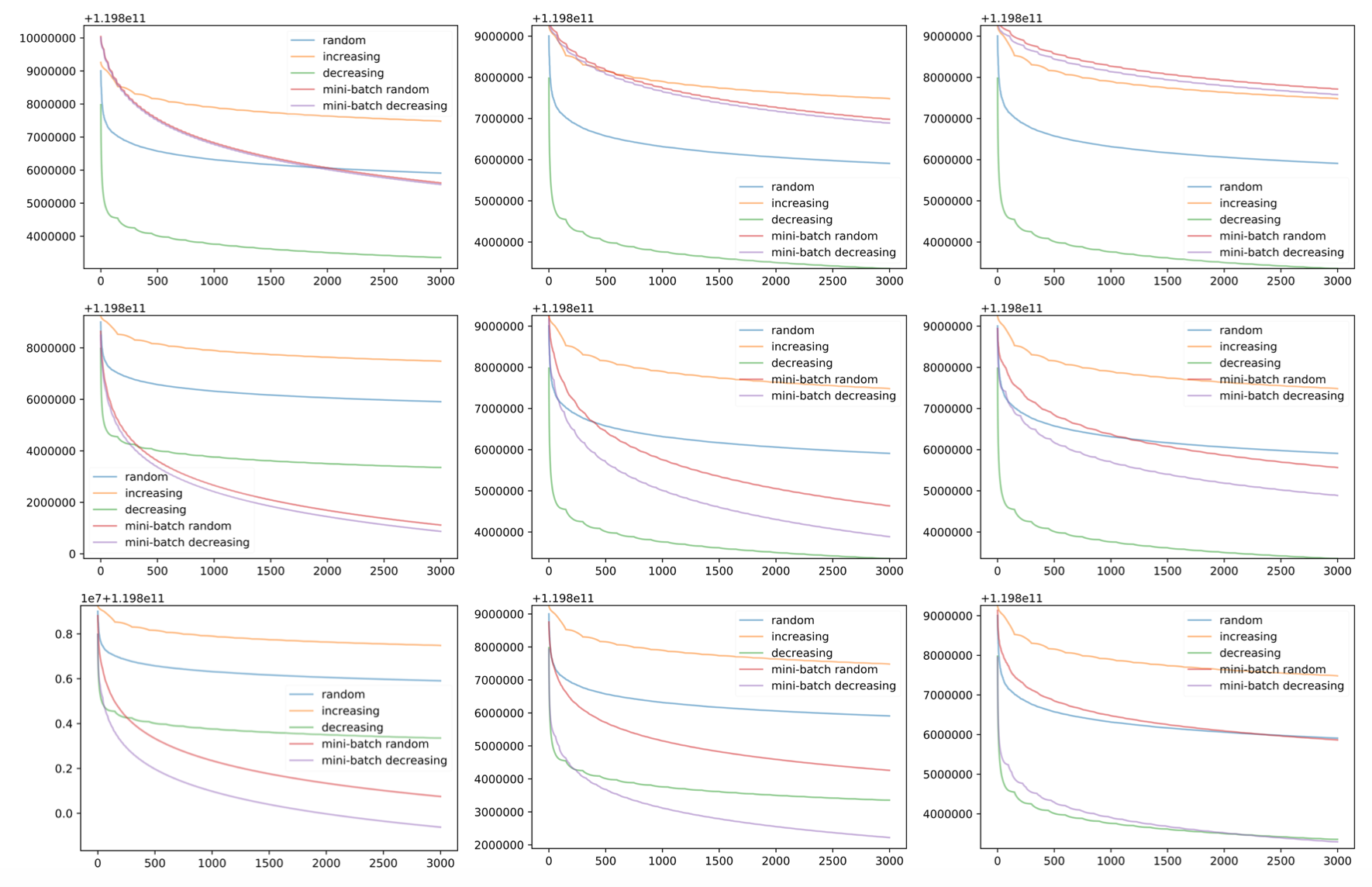}
    \caption{Iris, decreasing step size}
    \label{fig:iris_mini_decrease}
\end{figure}
\begin{figure}[htbp]\centering
    \includegraphics[width=0.8\textwidth]{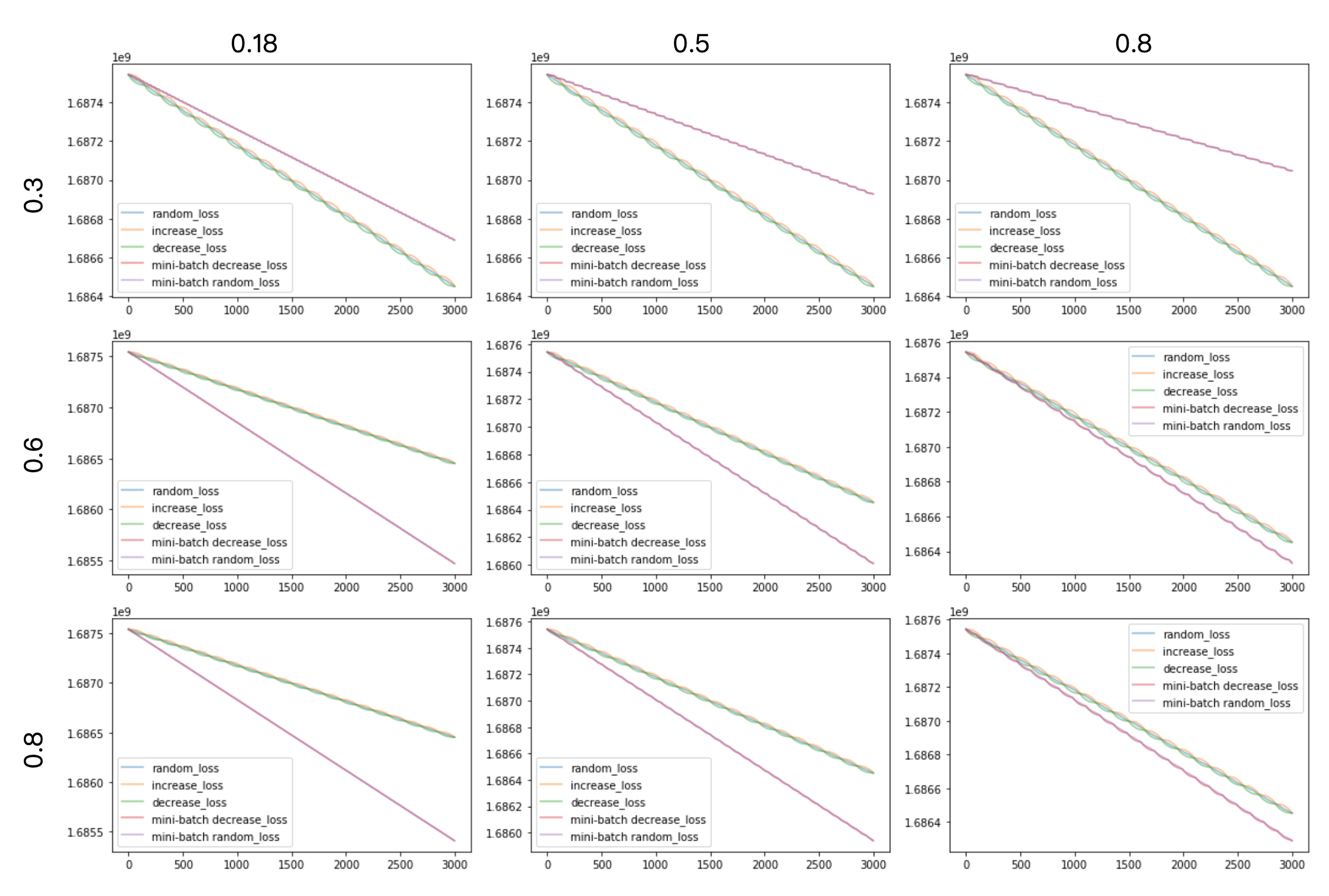}
    \caption{Iris, constant step size}
    \label{fig:iris_mini_const}
\end{figure}

\pagebreak
\subsection{Full Results for Experiments on Neural Networks with MNIST Data Sets}
\label{exp:nn_mnist}

For the $2x2$ ordered graphs, the rows correspond to complexity of the model (2-layer or 7-layer), and the columns correspond to the mini-batch size ($S = 128, 256$). For the experiments labeled mini-batch, $q = 0.5$.\\
\\
The following results are from experiments for MNIST Data Set:\\

\begin{figure}[!htb] \centering
    \includegraphics[width=0.8\textwidth]{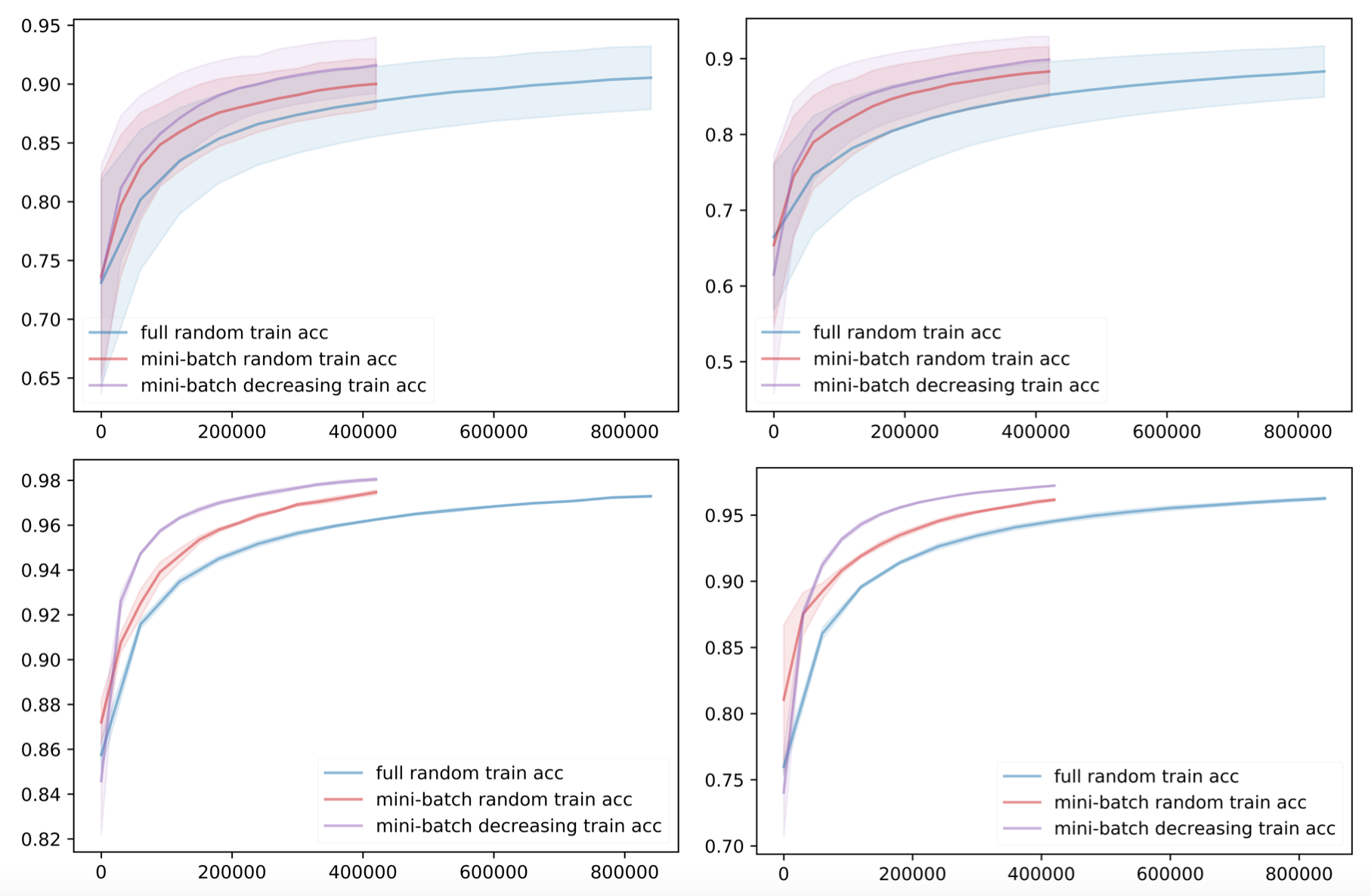}
    \caption{Train Accuracy for MNIST, sort within}
    \label{fig:mnist_train_acc_comp}
\end{figure}
\begin{figure}[!htb] \centering
    \includegraphics[width=0.8\textwidth]{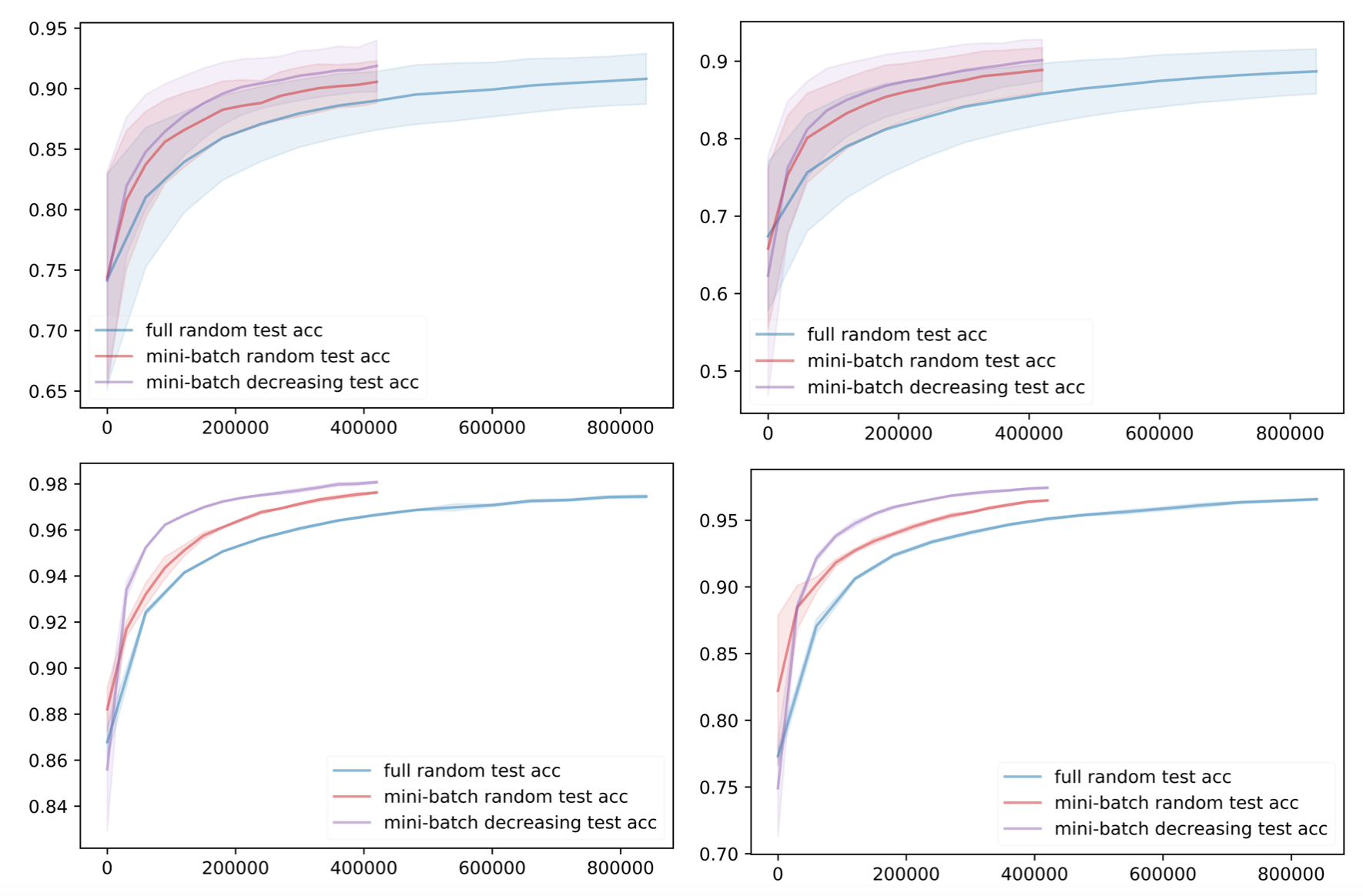}
    \caption{Test Accuracy for MNIST, sort within}
    \label{fig:mnist_test_acc_comp}
\end{figure}

\begin{figure}[!htb] \centering
    \includegraphics[width=0.8\textwidth]{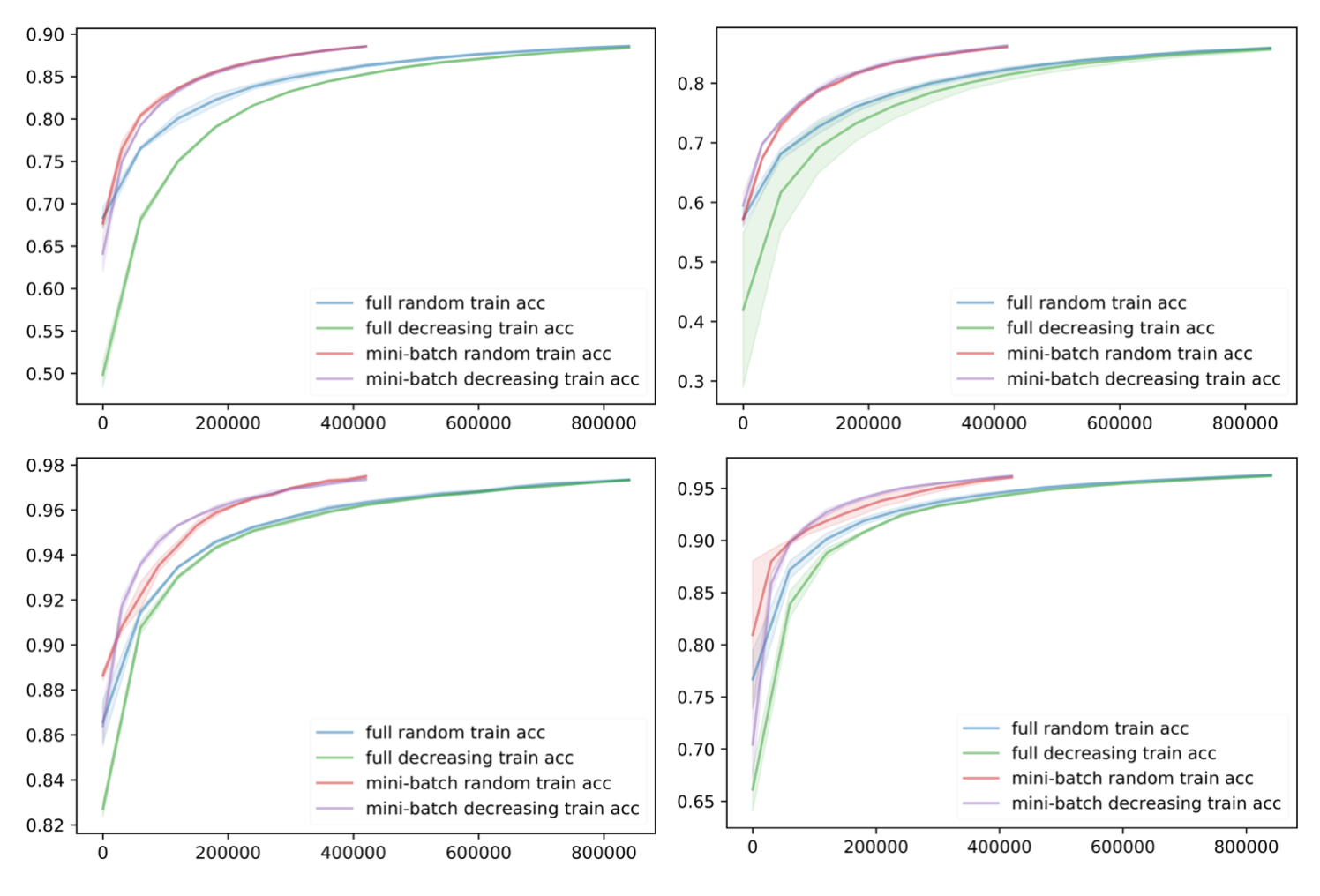}
    \caption{Train Accuracy for MNIST, sort first}
    \label{fig:mnist_bf_train_acc}
\end{figure}
\begin{figure}[!htb] \centering
    \includegraphics[width=0.8\textwidth]{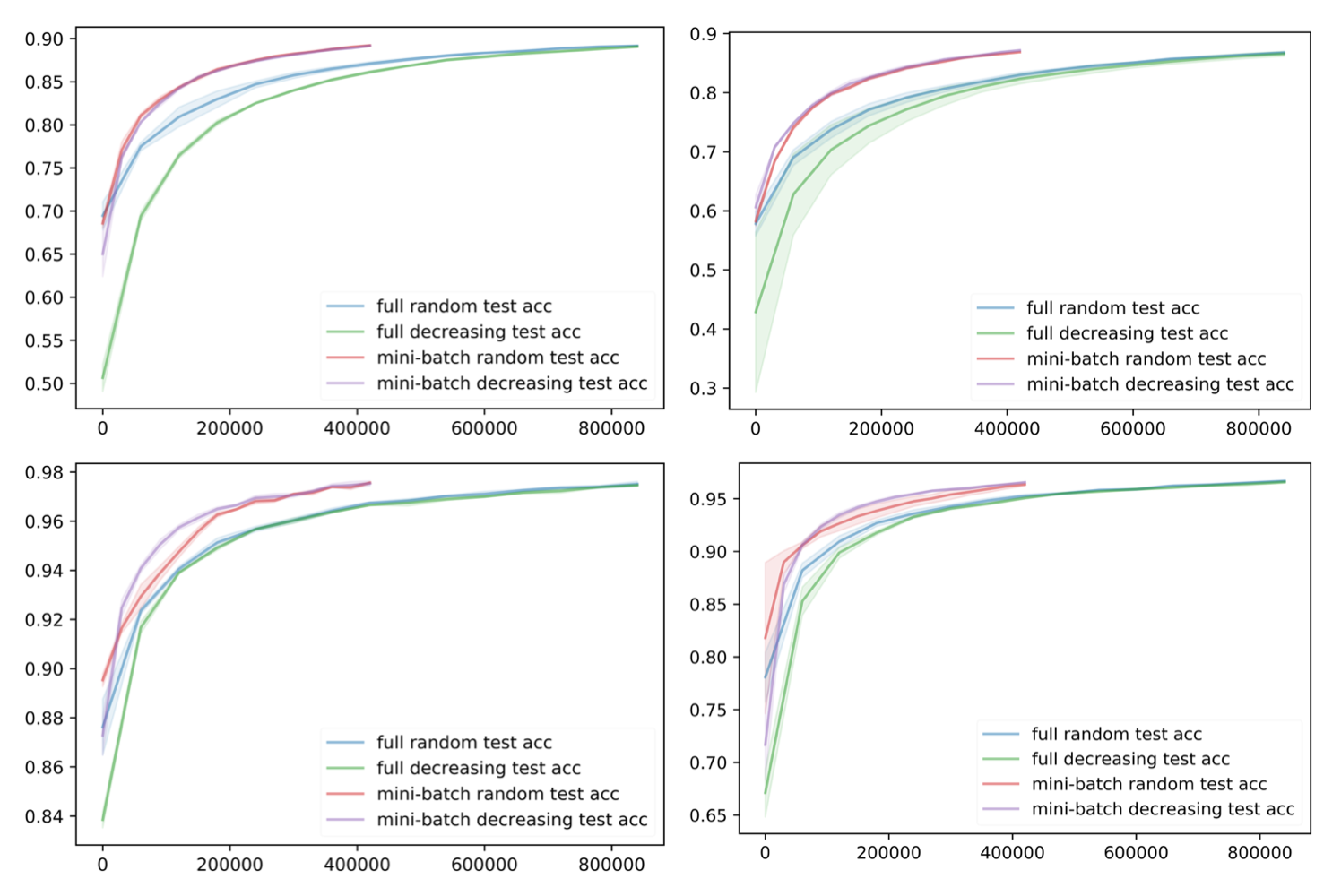}
    \caption{Test Accuracy for MNIST, sort first}
    \label{fig:mnist_bf_test_acc}
\end{figure}

The following results are from experiments for Fashion MNIST Data Set:\\

\begin{figure}[!htb] \centering
    \includegraphics[width=0.8\textwidth]{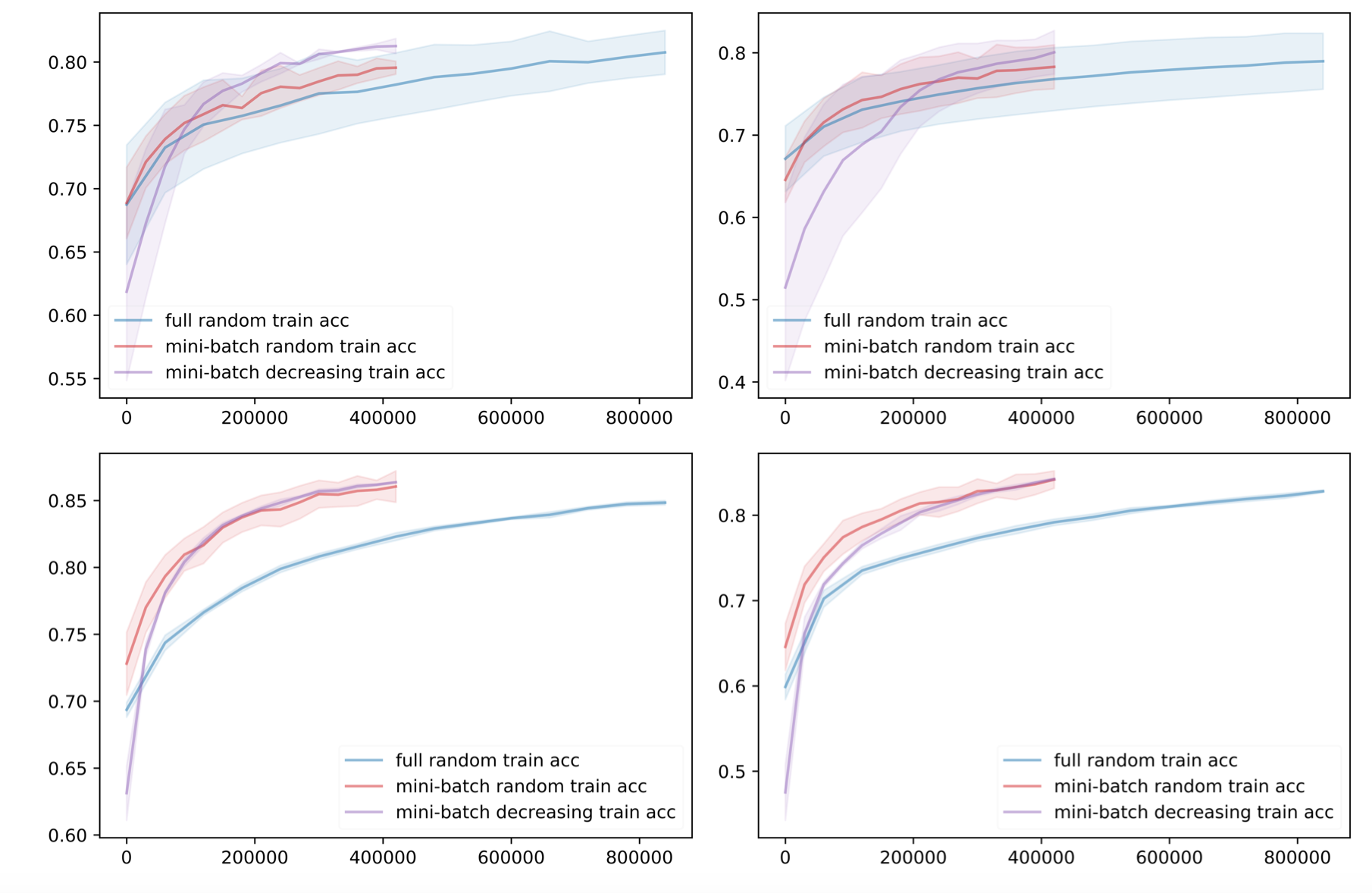}
    \caption{Train Accuracy for Fashion MNIST, sort within}
    \label{fig:fmnist_train_acc_comp}
\end{figure}
\begin{figure}[!htb] \centering
    \includegraphics[width=0.8\textwidth]{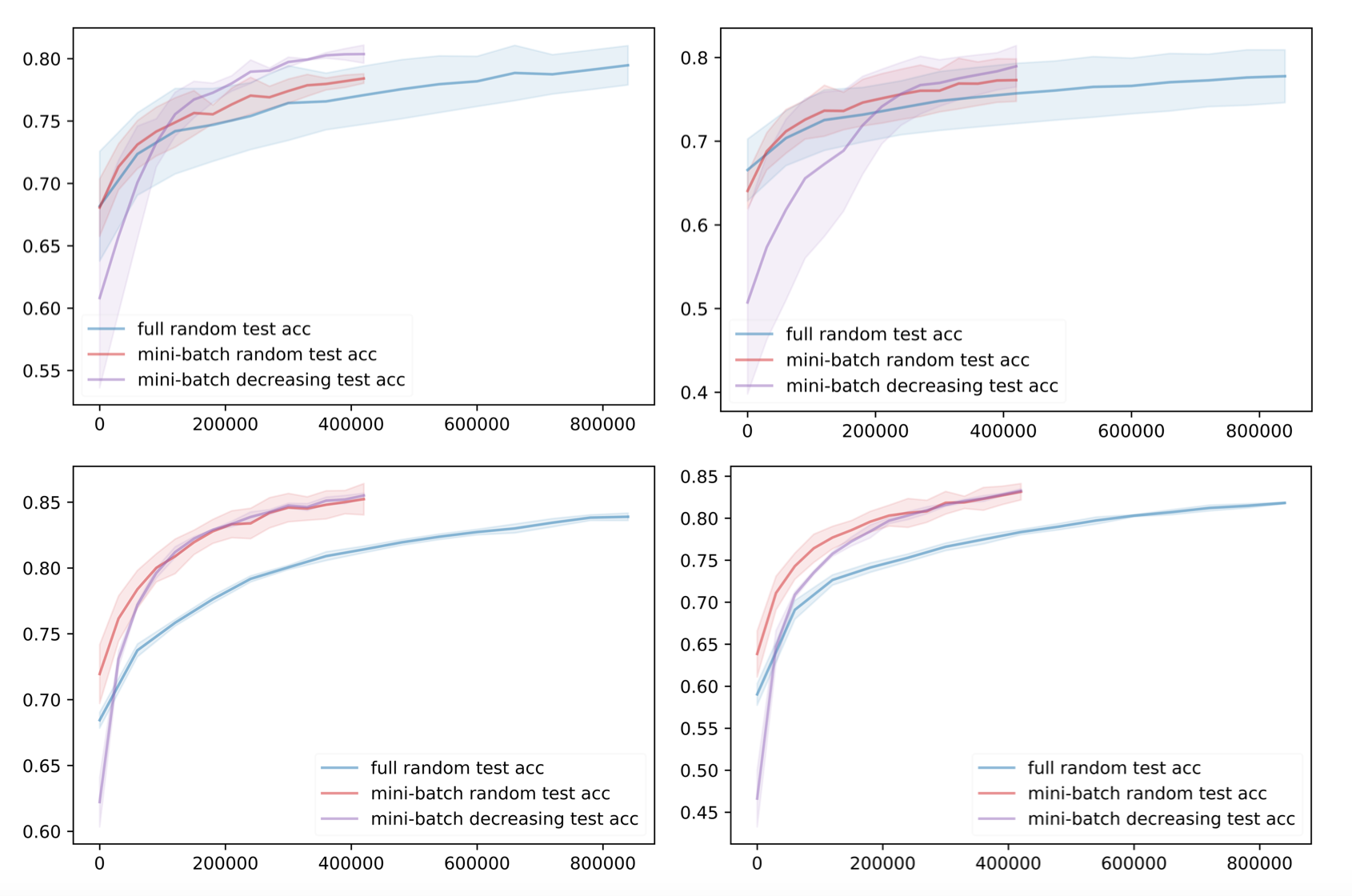}
    \caption{Test Accuracy for Fashion MNIST, sort within}
    \label{fig:fmnist_test_acc_comp}
\end{figure}

\begin{figure}[!htb] \centering
    \includegraphics[width=0.8\textwidth]{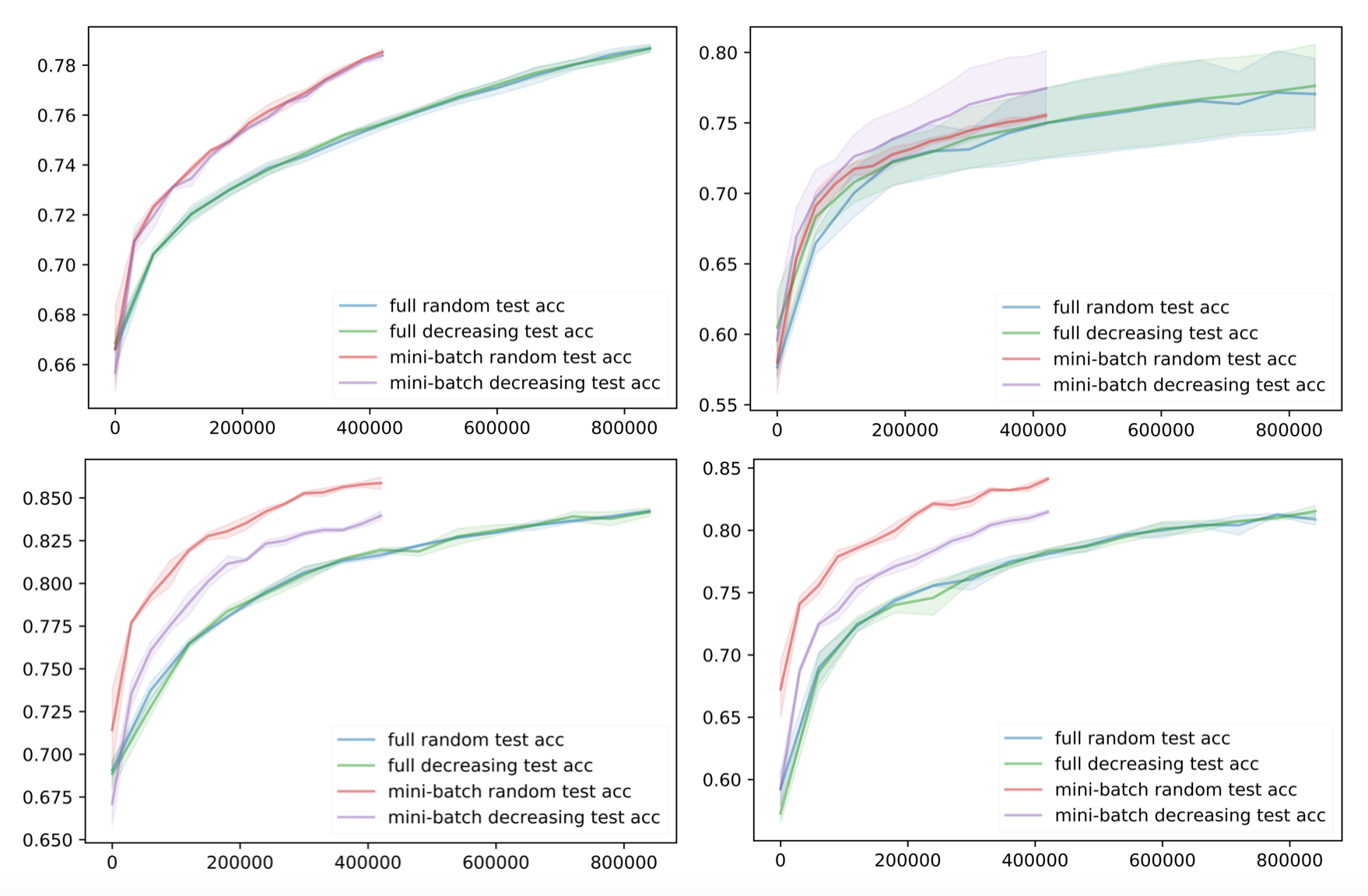}
    \caption{Train Accuracy for Fashion MNIST, sort first}
    \label{fig:fmnist_bf_train_acc}
\end{figure}
\begin{figure}[!htb] \centering
    \includegraphics[width=0.8\textwidth]{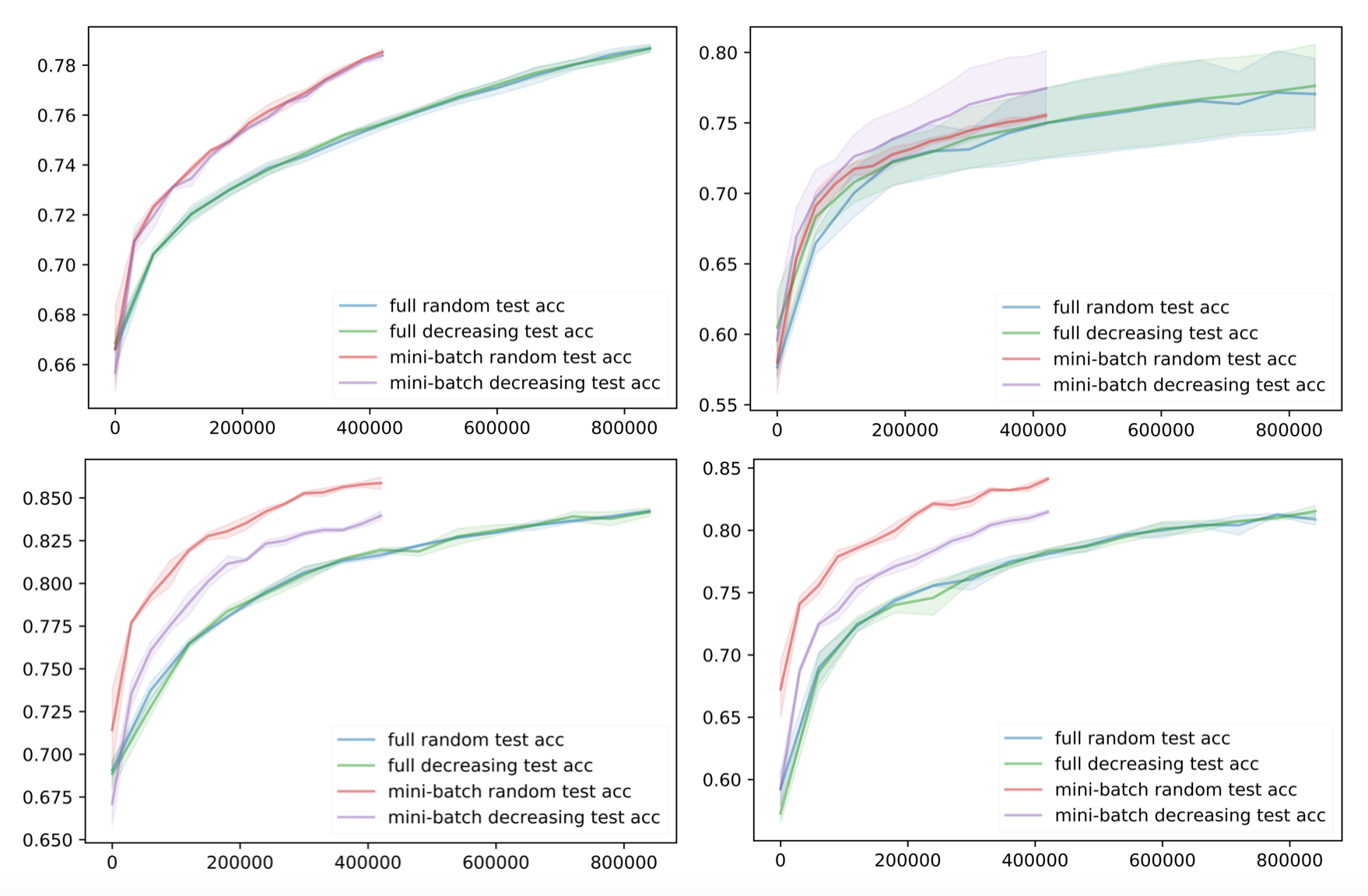}
    \caption{Test Accuracy for Fashion MNIST, sort first}
    \label{fig:fmnist_bf_test_acc}
\end{figure}

\clearpage
\subsection{Full Results for Experiments on Neural Networks with CIFAR Data Sets} \label{exp:nn_cifar}
\begin{figure}[!htb] \centering
    \includegraphics[width=0.8\textwidth]{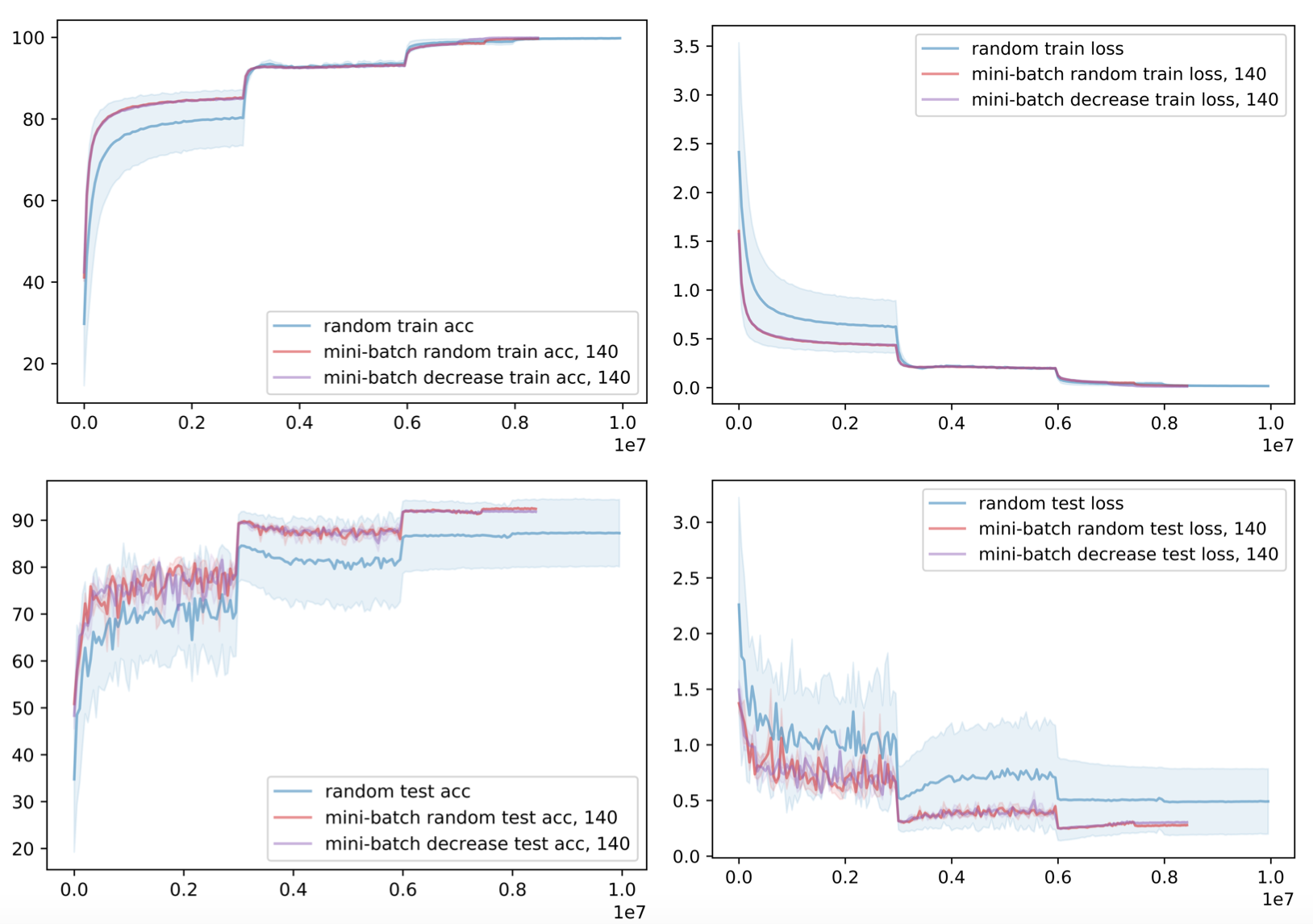}
    \caption{CIFAR-10 RESNET20 Full Plot, sort first}
    \label{fig:cifar10_resnet_20_full}
\end{figure}
\begin{figure}[!htb] \centering
    \includegraphics[width=0.8\textwidth]{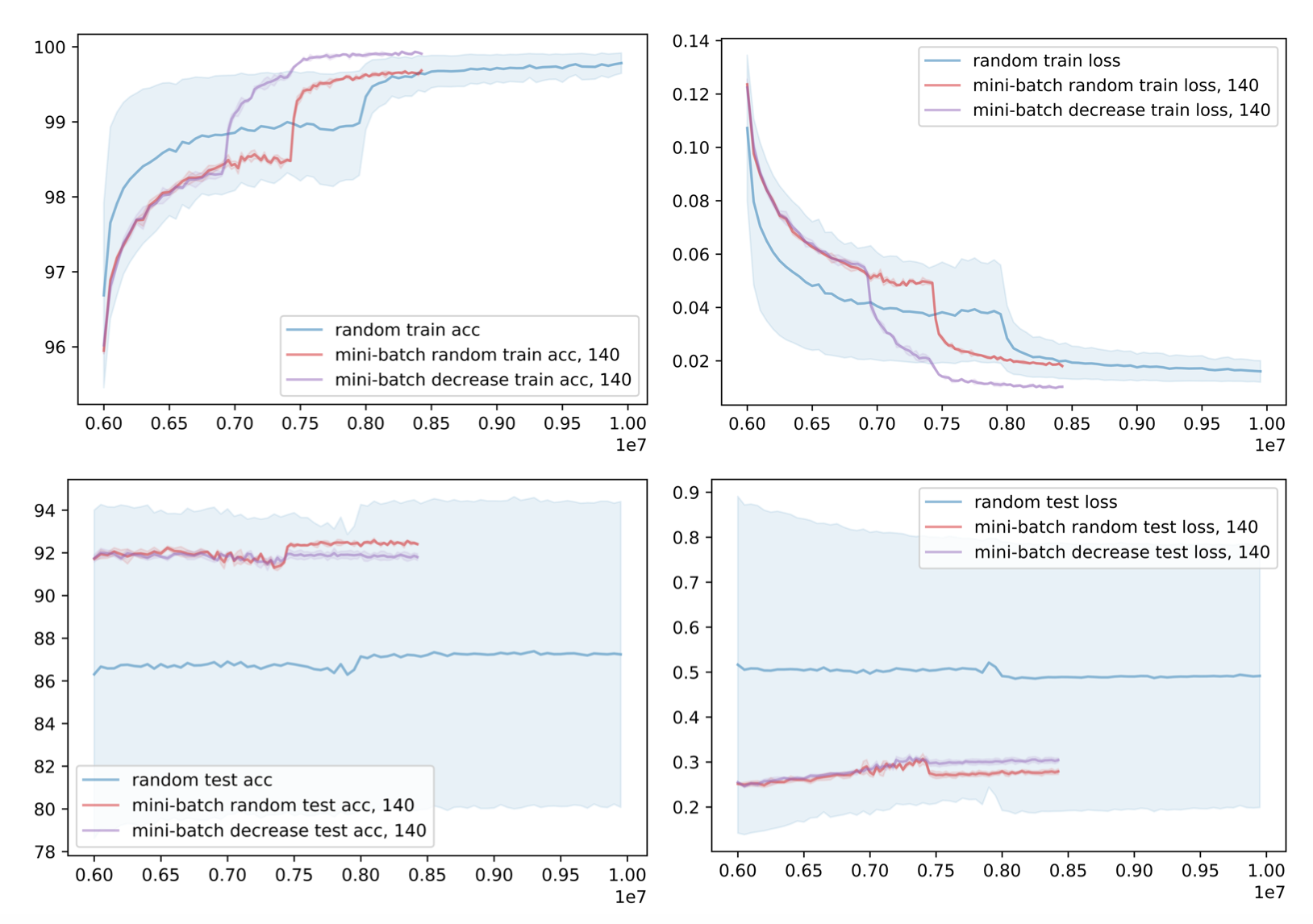}
    \caption{CIFAR-10 RESNET20 From Epoch 120, sort first}
    \label{fig:cifar10_resnet_20_0.6}
\end{figure}
\begin{figure}[!htb] \centering
    \includegraphics[width=0.8\textwidth]{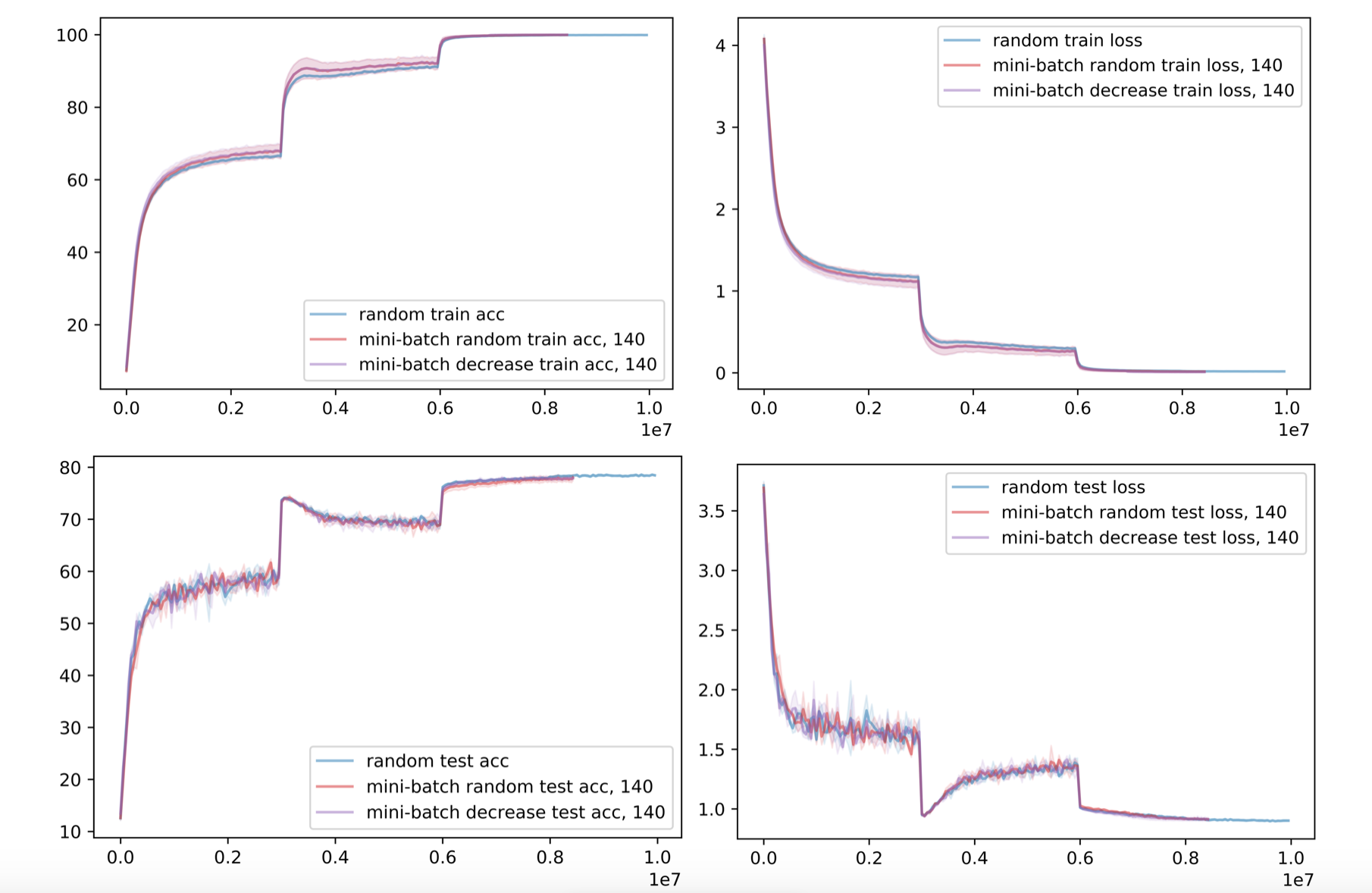}
    \caption{CIFAR-100 RESNET18 Full Plot, sort first}
    \label{fig:cifar100_resnet_18_full}
\end{figure}
\begin{figure}[!htb] \centering
    \includegraphics[width=0.8\textwidth]{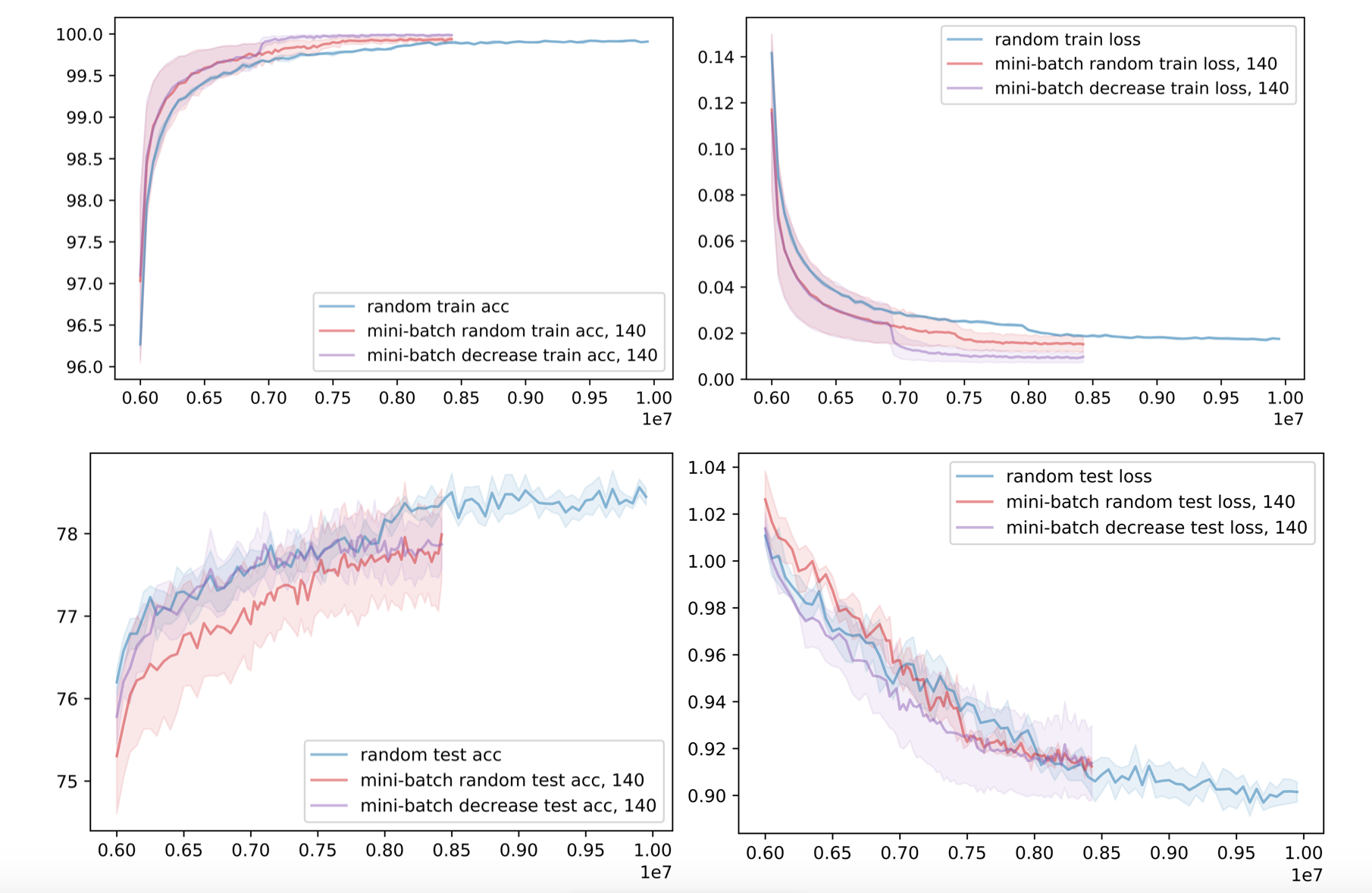}
    \caption{CIFAR-100 RESNET18 From Epoch 120, sort first}
    \label{fig:cifar100_resnet_18_0.6}
\end{figure}
\begin{figure}[!htb] \centering
    \includegraphics[width=0.8\textwidth]{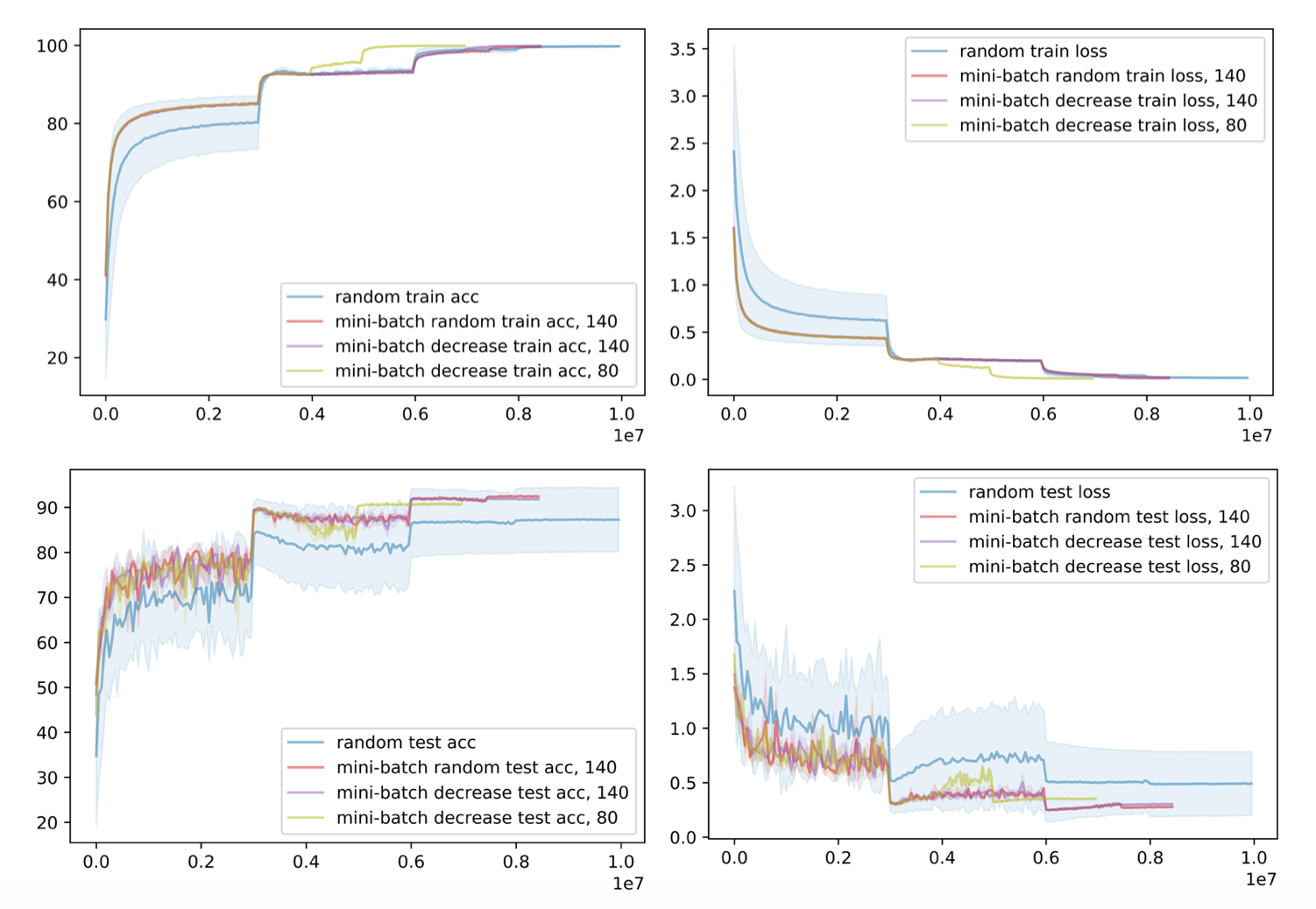}
    \caption{CIFAR-10 RESNET20: Start Ordering at Epoch 80, Full, sort first}
    \label{fig:cifar10_resnet_20_80_full}
\end{figure}
\begin{figure}[!htb] \centering
    \includegraphics[width=0.8\textwidth]{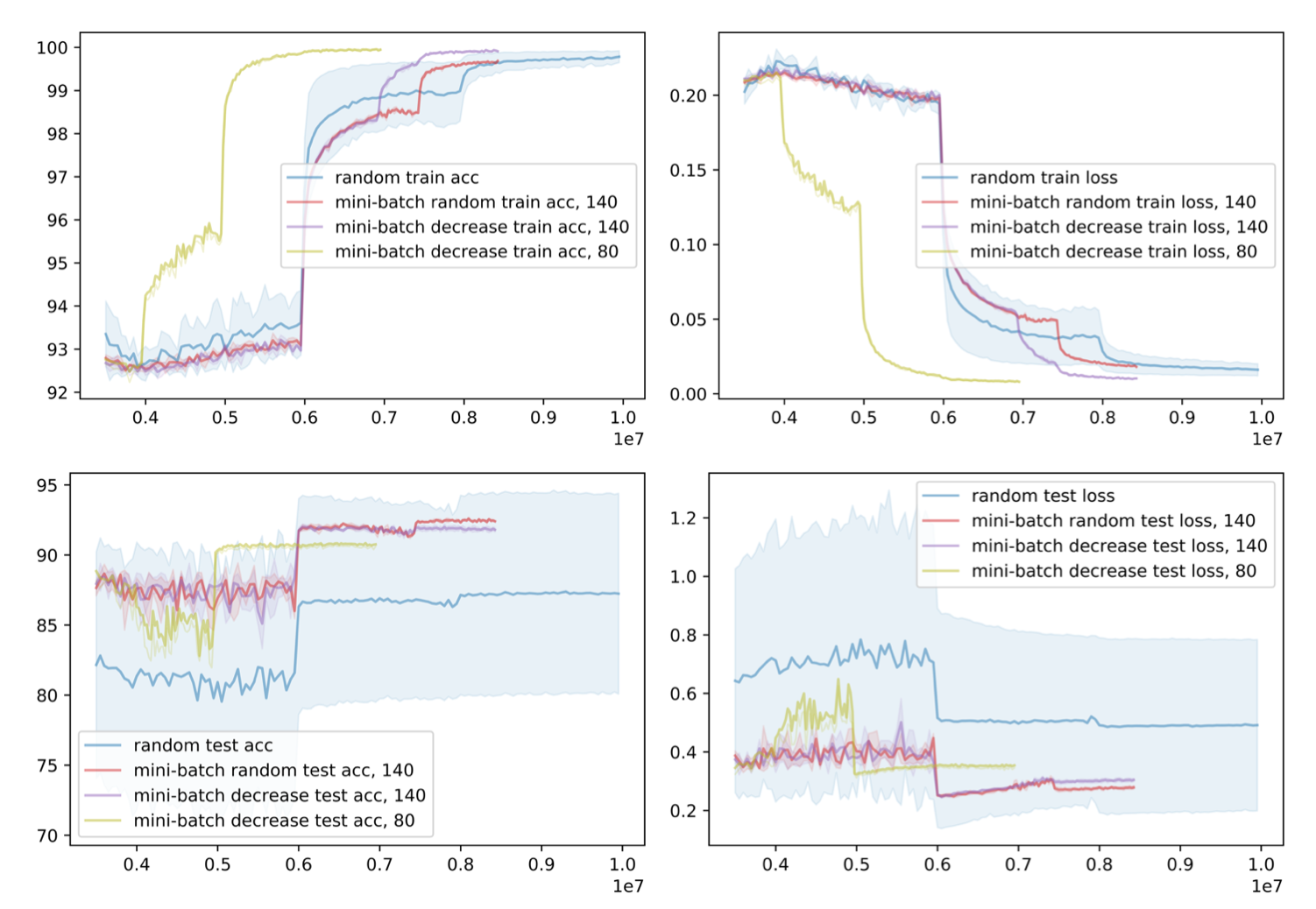}
    \caption{CIFAR-10 RESNET20: Start Ordering at Epoch 80, From Epoch 70, sort first}
    \label{fig:cifar10_resnet_20_80_0.6}
\end{figure}
\begin{figure}[!htb] \centering
    \includegraphics[width=0.8\textwidth]{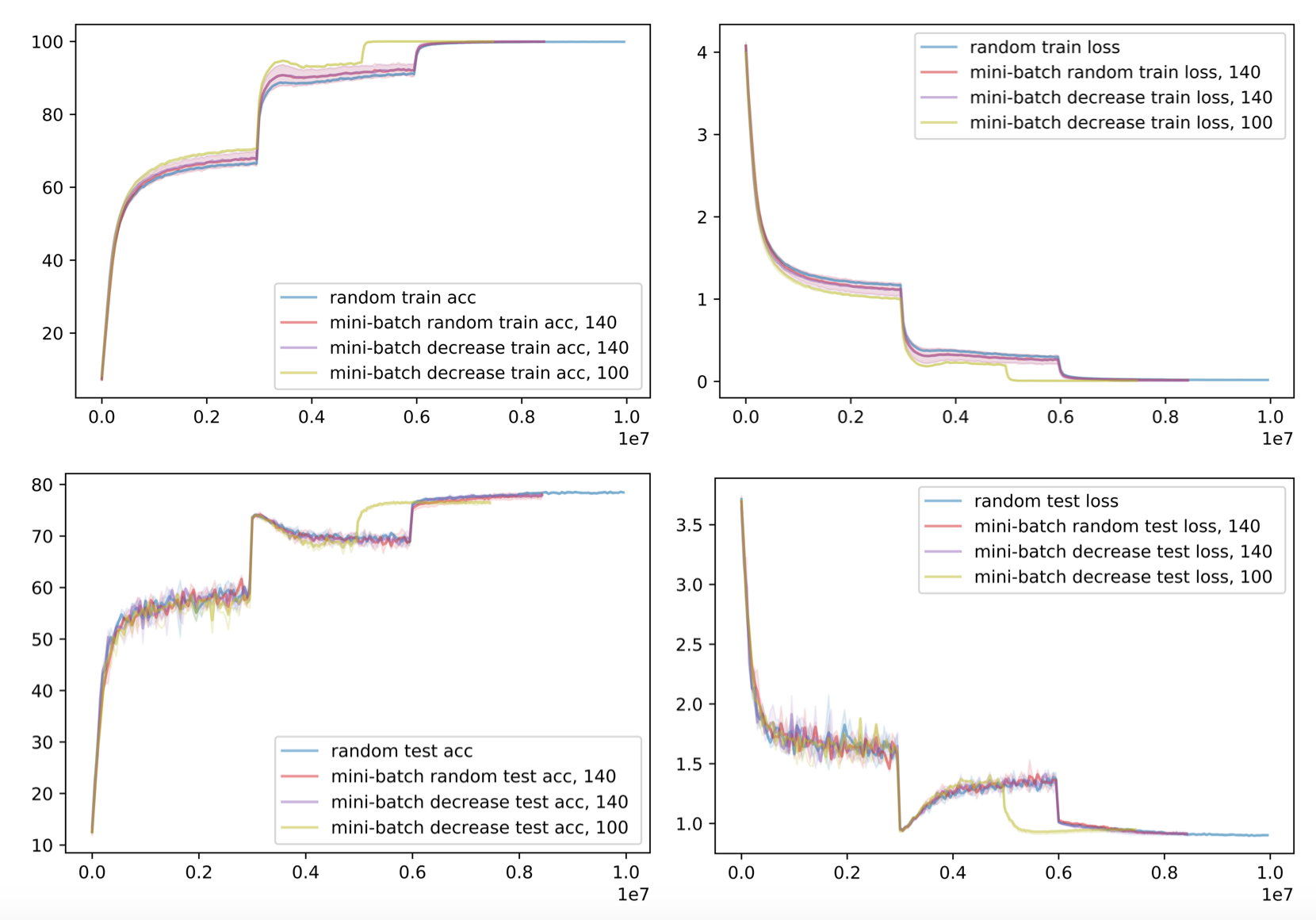}
    \caption{CIFAR-100 RESNET18: Start Ordering at Epoch 100, Full, sort first}
    \label{fig:cifar100_resnet_18_100_full}
\end{figure}
\begin{figure}[!htb] \centering
    \includegraphics[width=0.8\textwidth]{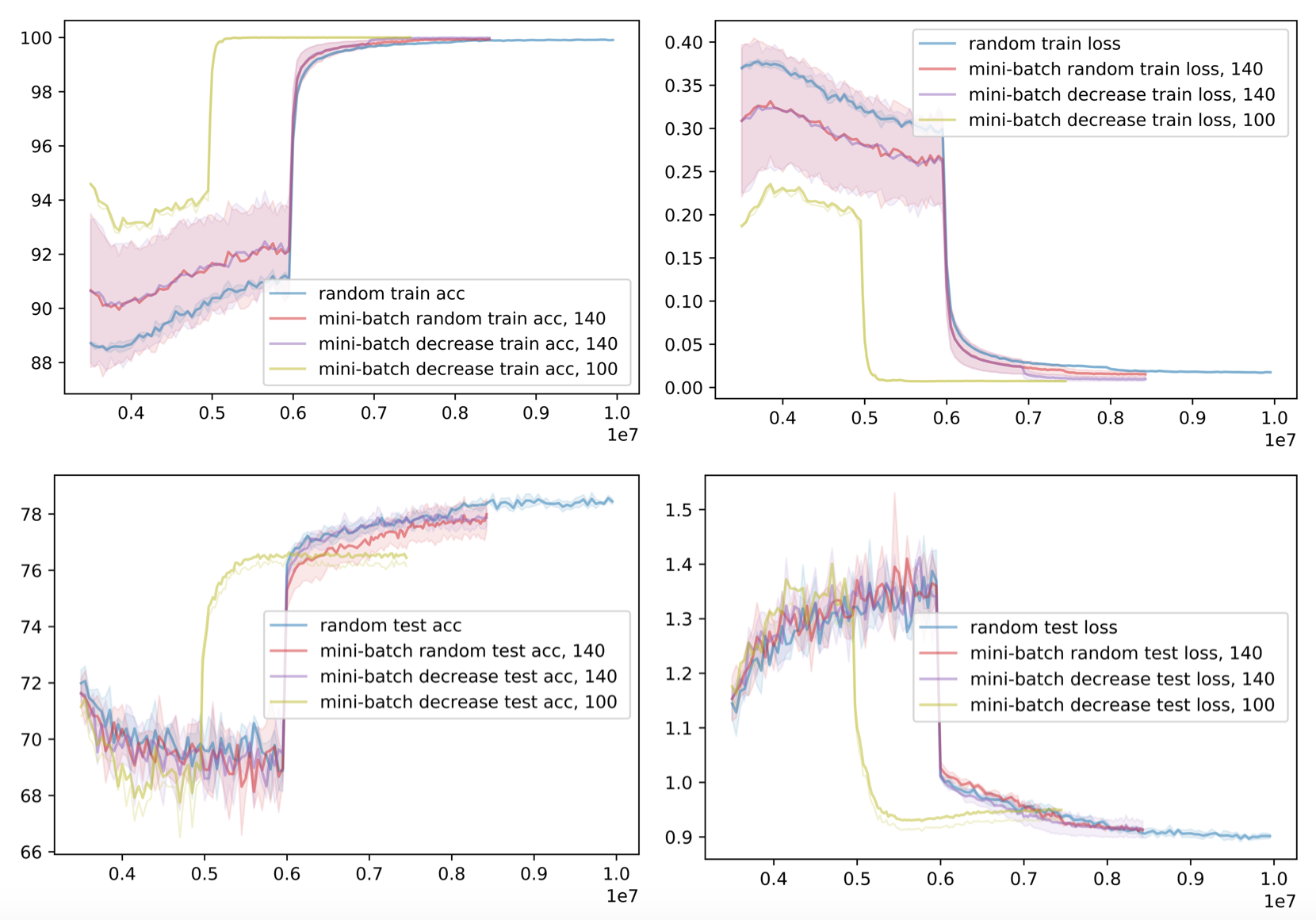}
    \caption{CIFAR-100 RESNET18: Start Ordering at Epoch 100, From Epoch 70, sort first}
    \label{fig:cifar100_resnet_18_100_0.6}
\end{figure}

\begin{figure}[!htb] \centering
    \includegraphics[width=0.8\textwidth]{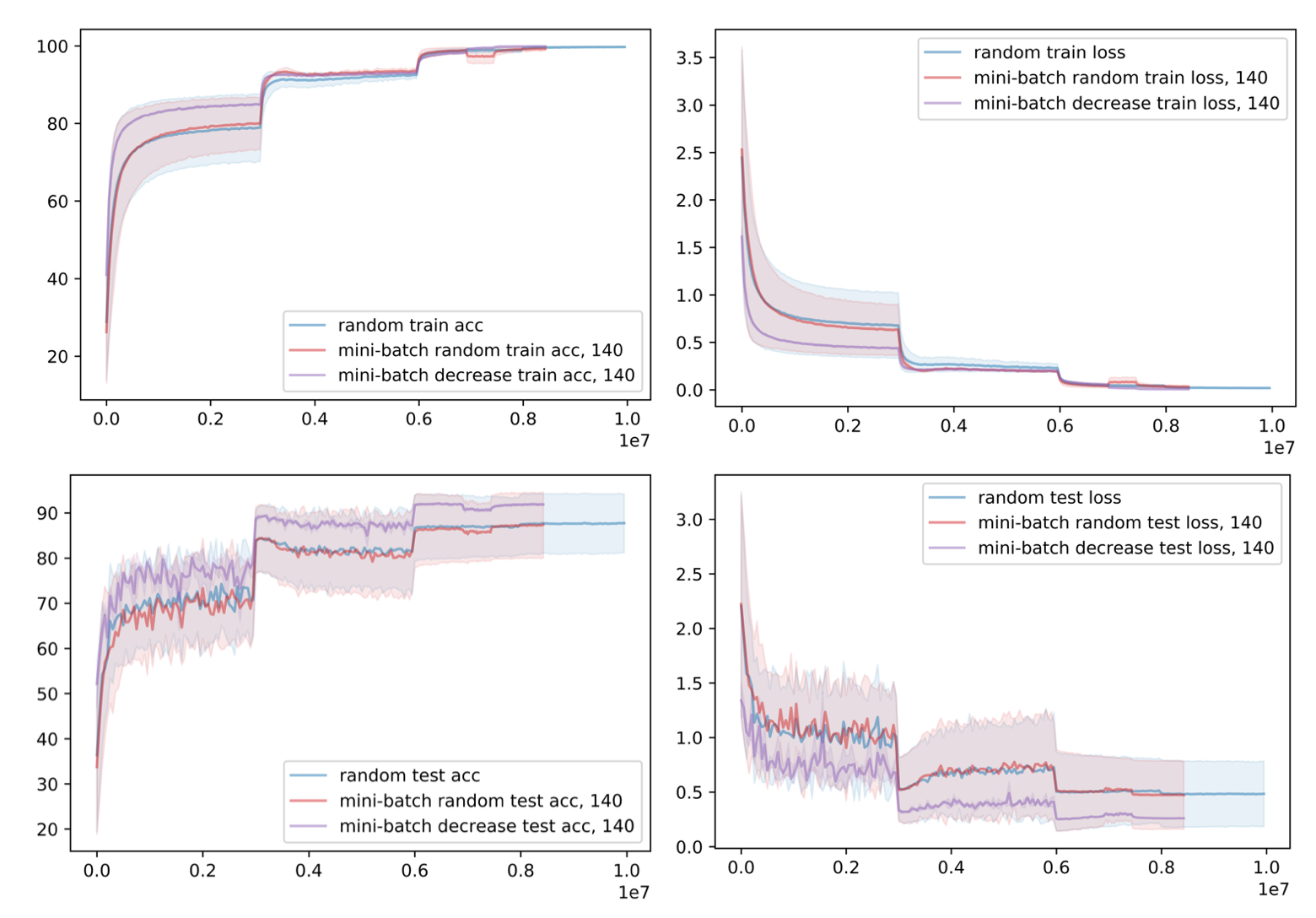}
    \caption{CIFAR-10 RESNET20 Full Plot, sort within}
    \label{fig:cifar10_resnet_20_af_full}
\end{figure}
\begin{figure}[!htb] \centering
    \includegraphics[width=0.8\textwidth]{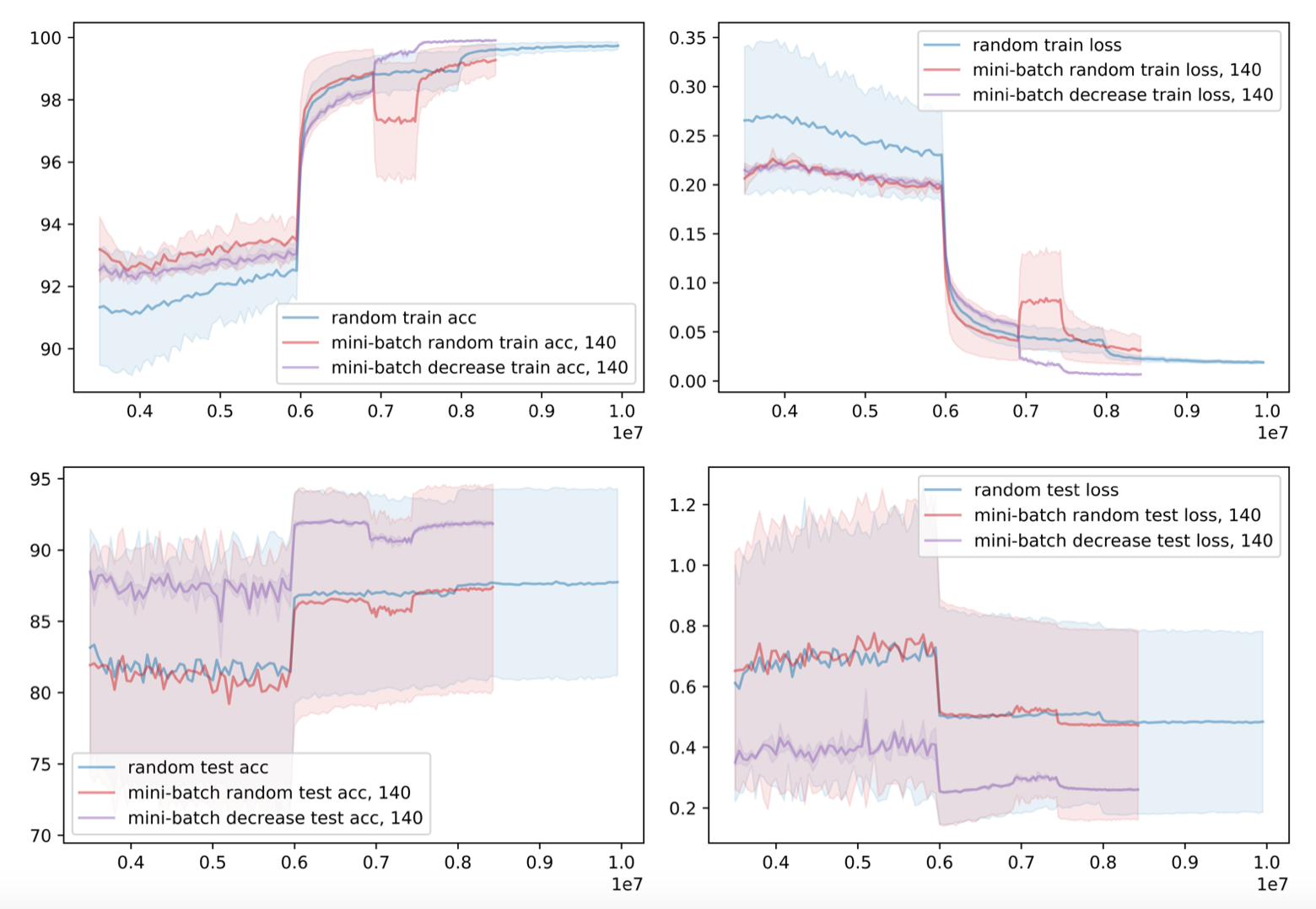}
    \caption{CIFAR-10 RESNET20 From Epoch 120, sort within}
    \label{fig:cifar10_resnet_20_af_120}
\end{figure}
\begin{figure}[!htb] \centering
    \includegraphics[width=0.8\textwidth]{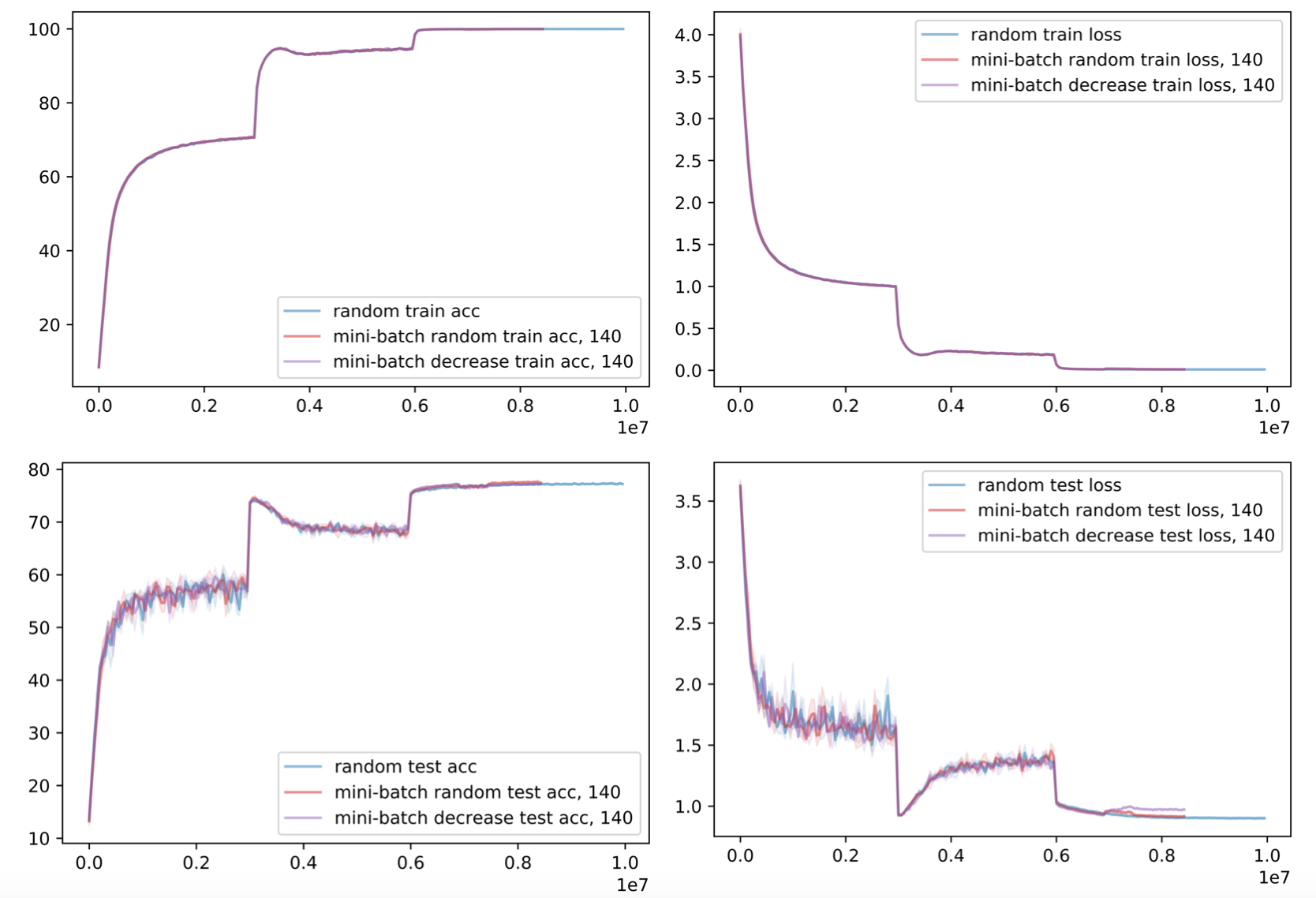}
    \caption{CIFAR-100 RESNET18 Full Plot, sort within}
    \label{fig:cifar100_resnet_18_af_full}
\end{figure}
\begin{figure}[!htb] \centering
    \includegraphics[width=0.8\textwidth]{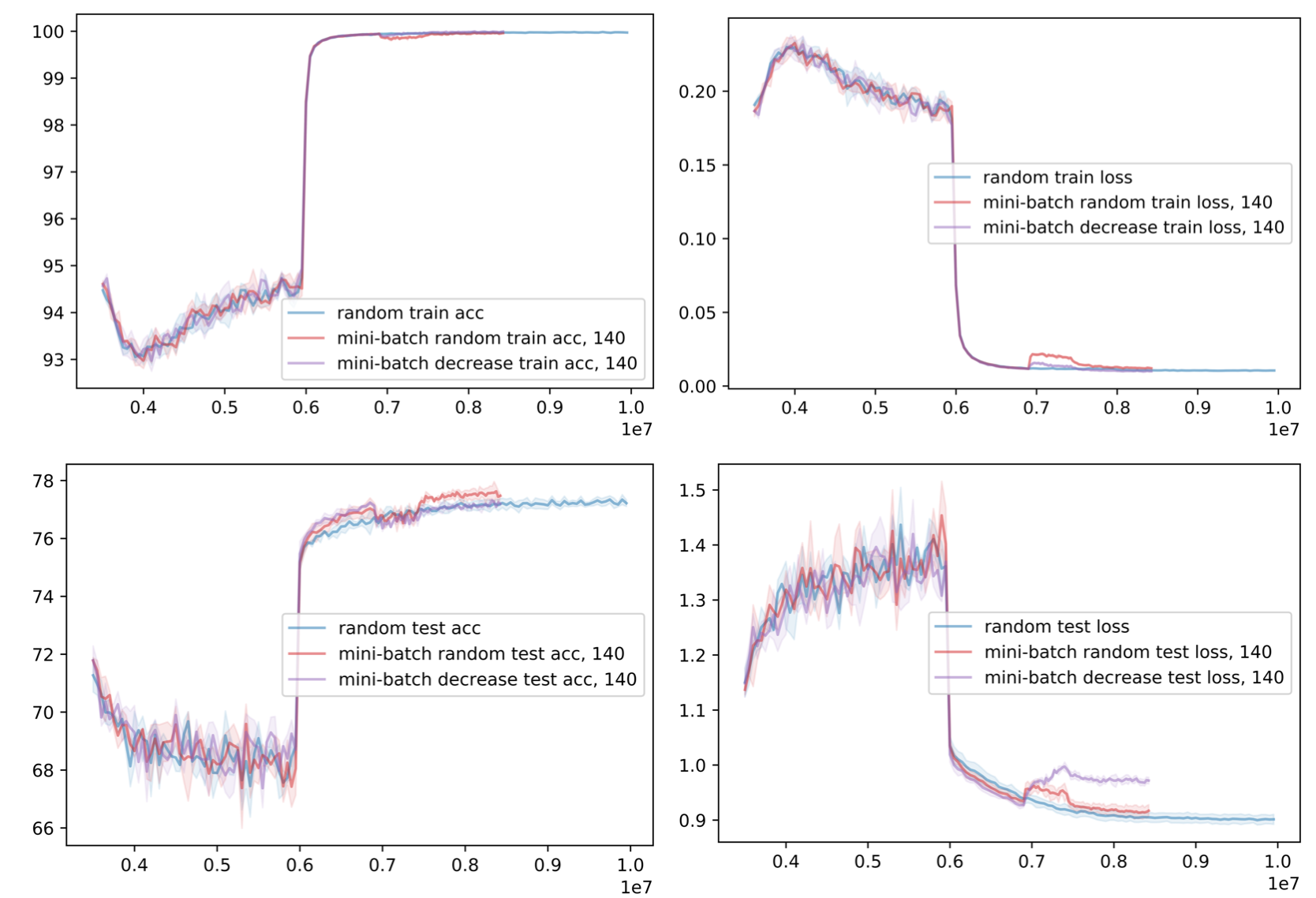}
    \caption{CIFAR-100 RESNET18 From Epoch 120, sort within}
    \label{fig:cifar100_resnet_18_af_120}
\end{figure}
\begin{figure}[!htb] \centering
    \includegraphics[width=0.8\textwidth]{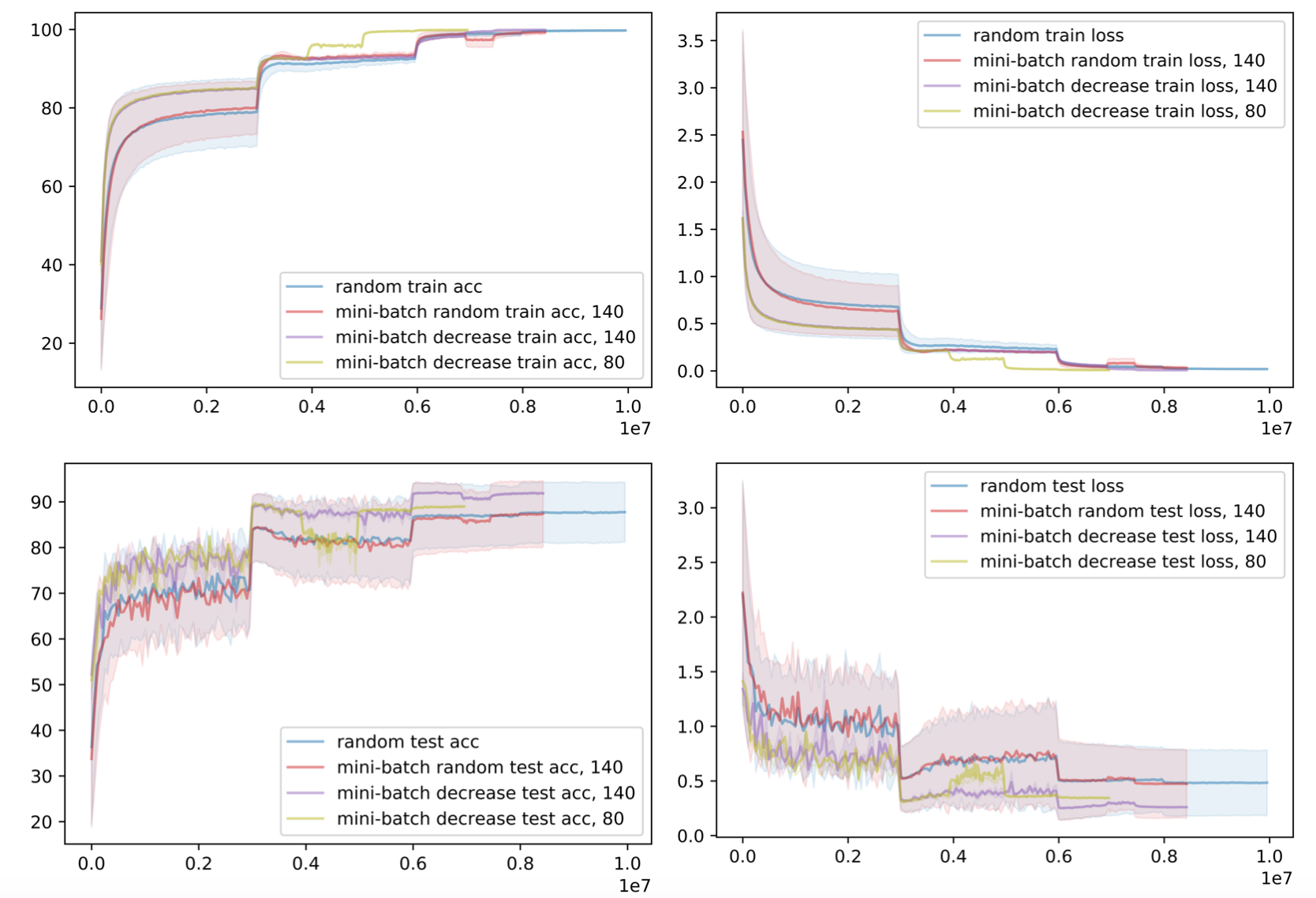}
    \caption{CIFAR-10 RESNET20: Start Ordering at Epoch 80, Full, sort within}
\label{fig:cifar10_resnet_20_af_80_full}
\end{figure}
\begin{figure}[!htb] \centering
    \includegraphics[width=0.8\textwidth]{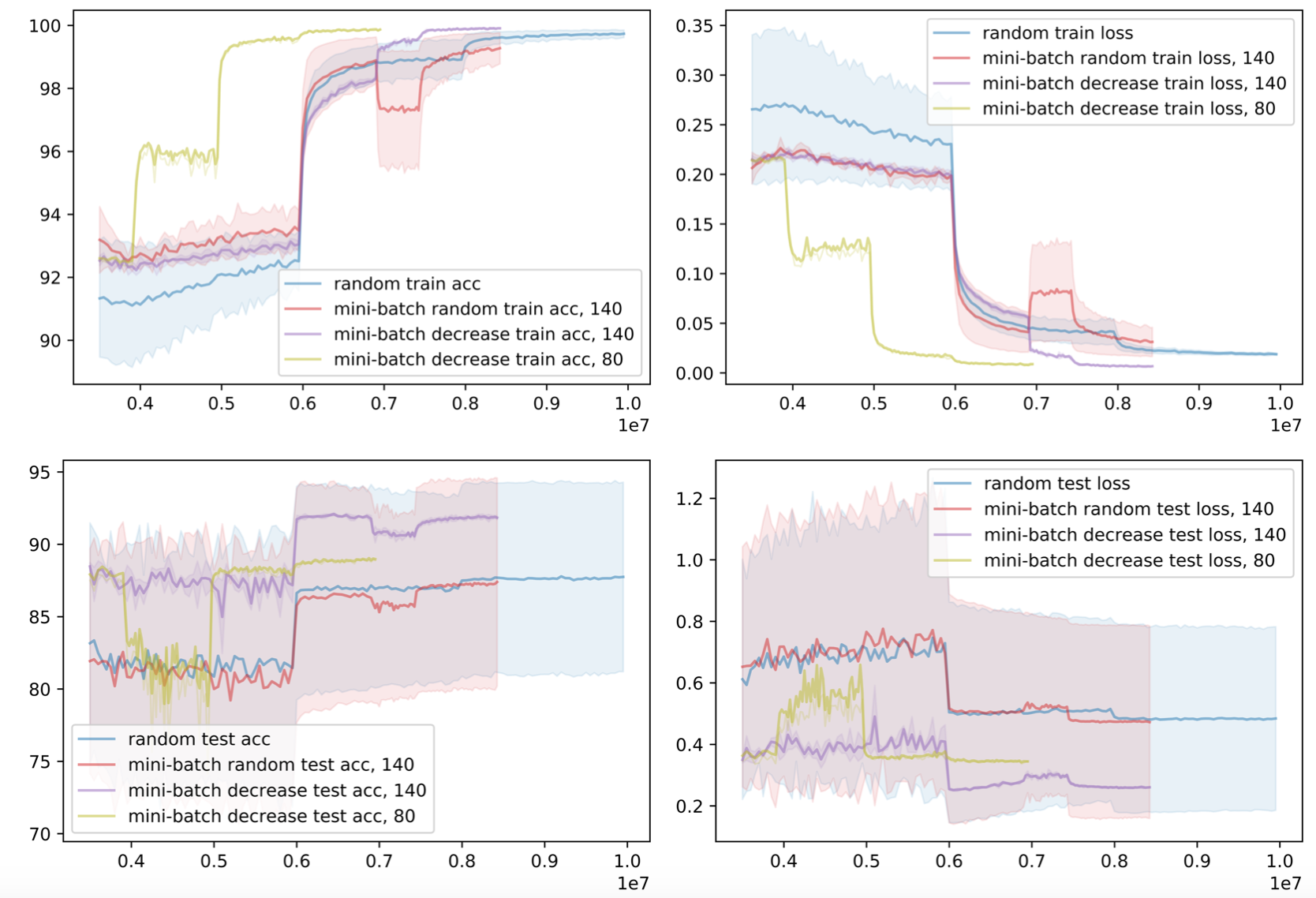}
    \caption{CIFAR-10 RESNET20: Start Ordering at Epoch 80, From Epoch 70, sort within}
    \label{fig:cifar10_resnet_20_af_80_70}
\end{figure}
\begin{figure}[!htb] \centering
    \includegraphics[width=0.8\textwidth]{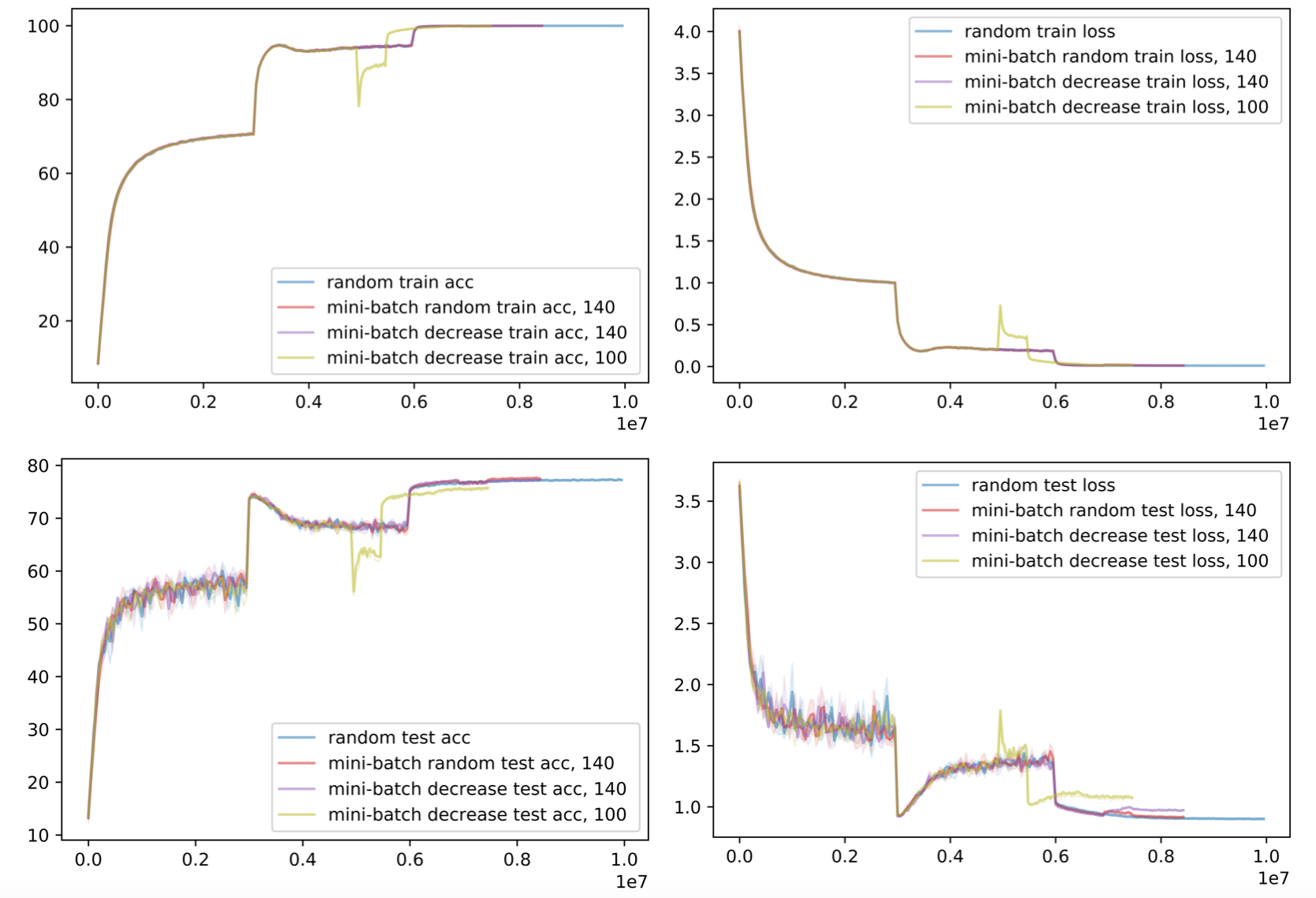}
    \caption{CIFAR-100 RESNET18: Start Ordering at Epoch 100, Full, sort within}    \label{fig:cifar100_resnet18_af_100_full}
\end{figure}
\begin{figure}[!htb] \centering
    \includegraphics[width=0.8\textwidth]{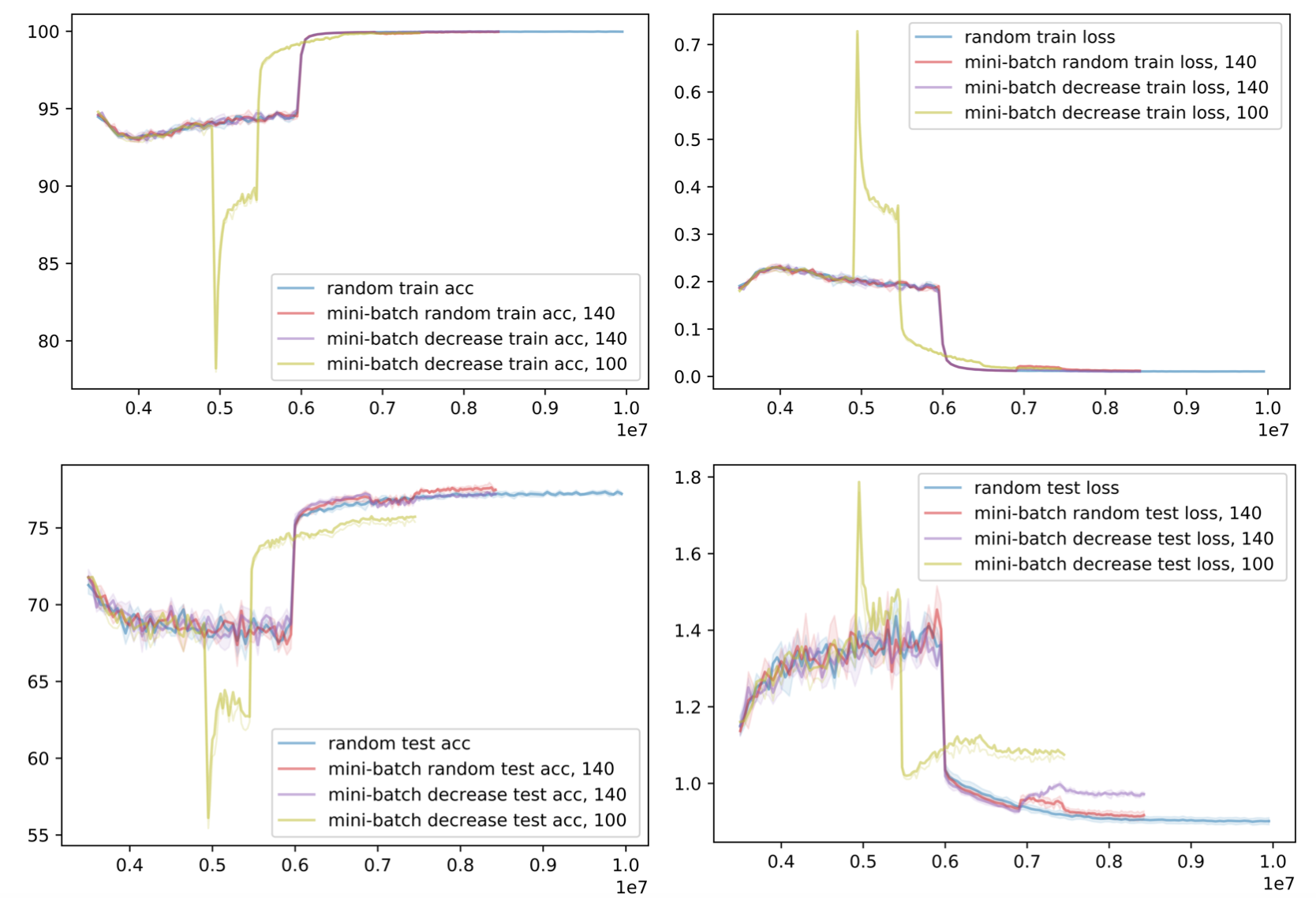}
    \caption{CIFAR-100 RESNET18: Start Ordering at Epoch 100, From Epoch 70, sort within}
    \label{fig:cifar100_resnet_18_af_100_70}
\end{figure}

\end{document}